\documentclass[10pt]{article} 
\usepackage[preprint]{tmlr}
\usepackage{hyperref}
\usepackage{url}
\usepackage{tikz}
\usetikzlibrary{mindmap,trees,positioning,shapes.geometric,arrows.meta}

\usepackage{graphicx}
\usepackage{adjustbox}
\usepackage{booktabs}
\usepackage{tabularx}
\usepackage{float}
\usepackage{caption}
\usepackage{accessibility}
\usepackage{xcolor} 

\usepackage{pgfplots}
\usepgfplotslibrary{colormaps}

\definecolor{lb}{HTML}{d0ebff}
\definecolor{lbb}{HTML}{74c0fc}

\usepackage{amsmath,amsfonts,bm}

\def\eqref#1{equation~\ref{#1}}

\def\1{\bm{1}}

\newcommand{\test}{\mathcal{D_{\mathrm{test}}}}

\DeclareMathAlphabet{\mathsfit}{\encodingdefault}{\sfdefault}{m}{sl}
\SetMathAlphabet{\mathsfit}{bold}{\encodingdefault}{\sfdefault}{bx}{n}

\begin{document}
\title{The Periodic Table of LLM Reasoning: \\A Structured Survey of Reasoning Paradigms, Methods, and Failure Modes}

\author{\name Avinash Anand \email Avinash.Anand@singaporetech.edu.sg \\
 \addr SIT $\times$ Nvidia AI Center (SNAIC) 
 \AND
\name Mahisha Ramesh \email mahisha23121@iiitd.ac.in \\
 \addr MIDAS Lab, IIIT Delhi
\AND
\name Avni Mittal \email avni.mittal2002@gmail.com\\
 \addr MIDAS Lab, IIT Mandi 
\AND
\name Ashutosh Kumar \email ak1825@rit.edu \\
\addr Owl Autonomous Imaging, Inc. \thanks{Work done outside of Owl AI}
\AND
\name Rishitej Reddy Vyalla \email rishitej23439@iiitd.ac.in \\
\addr MIDAS Lab, IIIT Delhi
\AND
\name Erik Cambria \email cambria@ntu.edu.sg\\
\addr Professor, College of Computing \& Data Science, NTU Singapore
\AND
\name Zhengkui Wang \email zhengkui.wang@singaporetech.edu.sg\\
\addr Associate Professor, Director, SIT $\times$ Nvidia AI Center (SNAIC)
\AND
\name Timothy Liu \email timothyl@nvidia.com\\
\addr NVIDIA AI Technology Centre, Singapore
\AND
\name Aik Beng Ng \email aikbengn@nvidia.com\\
\addr NVIDIA AI Technology Centre, Singapore
\AND
\name Simon See \email ssee@nvidia.com\\
\addr NVIDIA AI Technology Centre, Singapore
\AND
\name Rajiv Ratn Shah \email rajivratn@iiitd.ac.in\\
\addr Associate Professor, Department of Computer Science and Engineering, IIT Kanpur
}

\newcommand{\fix}{\marginpar{FIX}}
\newcommand{\new}{\marginpar{NEW}}

\def\month{MM} 
\def\year{YYYY} 
\def\openreview{\url{https://openreview.net/forum?id=XXXX}} 

\maketitle

\begin{abstract}

Reasoning is now a primary focus of how we evaluate, improve, and interpret Large Language Models (LLMs), covering Chain-of-Thought (CoT) inference, mathematical problem-solving, multi-hop question answering, code generation, retrieval-augmented reasoning, tool use, and multimodal decision-making. In this survey, we present the Periodic Table of LLM Reasoning, a structured framework to categorize more than 300 recent papers along reasoning paradigms, methodological mechanisms, evaluation settings, and common failure modes. This paper classifies LLM reasoning into nine major paradigms: Chain-of-Thought reasoning, Multi-Hop reasoning, Mathematical reasoning, Commonsense reasoning, Visual and Temporal reasoning, Code and Algorithmic reasoning, Retrieval-Augmented reasoning, Tool-Augmented or Agentic reasoning, and Reinforcement reasoning on learning. We discuss the most prominent methods for each paradigm, including prompting strategies, architectural interventions, supervised fine-tuning, verifier-guided inference, reward modeling, retrieval mechanisms, tool interfaces, agentic workflows, and benchmark design. The paper argues that LLM reasoning is not a single emergent capability but a family of scaffolded behaviors that are shaped by the interaction between a model scale, task structure, external memory, supervision signals, and evaluation protocols. We also synthesize recurring failure modes across paradigms, including reasoning hallucinations, brittle multi-step inference, spurious rationales, weak causal grounding, poor out-of-distribution generalization, benchmark contamination, and unreliable self-verification. Although many recent methods have shown improvements in specific reasoning settings, we observe that it is often hard to compare progress across paradigms, as gains might come from prompting, retrieval, verifier design, or benchmark-specific structure rather than from more general reasoning ability. The survey connects reasoning methods with their assumptions, strengths, and ways of failing, providing a reference map of the current field and a diagnostic framework for assessing future work. More generally, we argue that robust LLM reasoning will require systems that go beyond isolated task performance to support meta-reasoning, multimodal and temporal grounding, adaptive tool use, and principled evaluation of reasoning under distribution shift.

\end{abstract}

\section{Introduction}

\begin{figure}[]
 \centering
 \includegraphics[width=\textwidth]{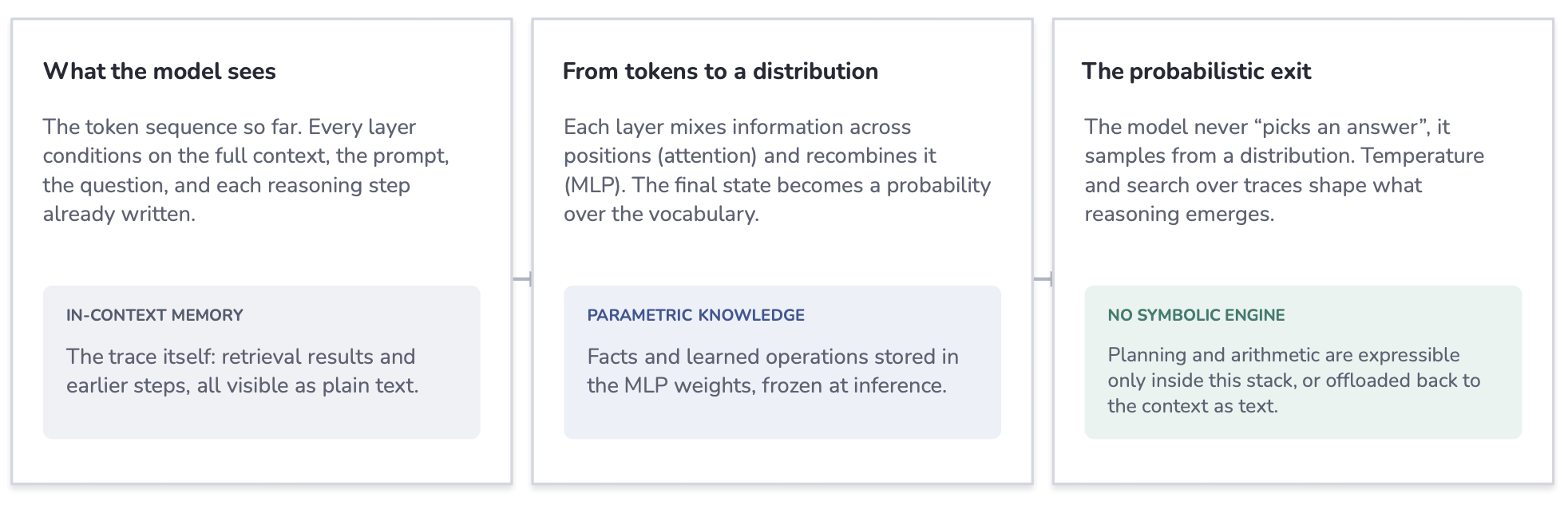}
 \caption{The foundational mechanism diagram. Shows one forward pass: context stream $\rightarrow$ layers $\rightarrow$ distribution $\rightarrow$ sampled token. Annotated to show where in-context info lives vs parametric knowledge vs attention, so students see that "reasoning" is repeated next-token prediction.
 }
 \label{fig:taxonomy_reasoning}
\end{figure}

LLMs have shown strong empirical performance on diverse reasoning tasks~\citep{livebench}; however, the depth, generality, and reliability of their reasoning abilities remain contested. This survey synthesizes existing work to examine the reasoning capabilities of LLMs, their limitations, and the open challenges in evaluating and improving reasoning capabilities (Figure~\ref{fig:taxonomy_reasoning}). The core question motivating this work: \textbf{can reasoning abilities emerge in LLMs without task-specific reasoning supervision?} While earlier neural language models relied largely on surface-level pattern matching and statistical regularities~\citep{geirhos2020shortcut, mccoy2019right, tenney2019bert}, recent large-scale transformer LLMs have demonstrated increasingly complex reasoning behaviors, including multi-step inference, compositional reasoning, temporal consistency, and limited causal or counterfactual reasoning~\citep{wei2022emergent, wei2022chain, chowdhery2023palm}. Nevertheless, these capabilities remain fragile and highly sensitive to prompting strategies, task formulation, distribution shift, and evaluation methodology~\citep{gendron2023large, berglund2024reversal}.

Prompting plays a central role in eliciting LLM capabilities. Techniques such as zero-shot prompting~\citep{xian2018zero,kojima2022large}, few-shot prompting~\citep{brown2020language}, and chain-of-thought (CoT) prompting~\citep{wei2022chain} demonstrate that the same model can exhibit qualitatively different behaviors depending on how a task is framed. This sensitivity has motivated extensive study of the interactions among model architecture, scale, and prompting strategies as mechanisms for reasoning. Although larger models generally achieve stronger reasoning performance, scale alone does not guarantee robust reasoning ability. Many reasoning behaviors emerge only when sufficient scale is paired with appropriate instruction tuning, prompting structure, and contextual guidance.

Transparency and interpretability present additional challenges for reasoning in LLMs. Although techniques such as chain-of-thought prompting~\citep{wang2023towards} expose intermediate reasoning traces~\citep{wei2022chain}, it remains doubtful whether these traces faithfully reflect the underlying computational processes~\citep{lyu2023faithful} or merely plausible post hoc rationalizations. This raises broader questions about whether LLM reasoning aligns with human-interpretable logical structure or instead emerges from statistical patterns learned during training~\citep{kadavath2022language}. As LLMs are increasingly deployed in high-stakes and real-world settings, understanding how models arrive at conclusions becomes as important as evaluating the correctness of their outputs. 

Despite recent progress, LLMs continue to exhibit important limitations, particularly in handling ambiguity and contradiction, distinguishing correlation from causation, transferring reasoning across domains, and grounding knowledge in physical or experiential reality~\citep{teolar}. These shortcomings have motivated the development of retrieval-augmented~\citep{lewis2020retrieval}, tool-augmented~\citep{schick2023toolformer}, and modular reasoning systems that combine language models with external memory, search, symbolic tools, or execution environments~\citep{yao2022react}. While such hybrid approaches often improve reliability and factual consistency, they also raise foundational questions about whether reasoning should be understood as a monolithic capability or as a composition of separable cognitive sub-processes~\citep{budagam2024hierarchical}.

This survey also examines reasoning in LLMs across multilingual and multimodal settings~\citep{achiam2023gpt, zhang2023multimodal}, mathematical and symbolic problem solving~\citep{hendrycks2021measuring}, and aspects of social and commonsense cognition. We compare these behaviors with human reasoning patterns while considering the extent to which current architectures genuinely perform structured reasoning versus simulating its observable outputs. Finally, we discuss meta-reasoning and self-reflective inference, where models evaluate and revise their own intermediate reasoning processes, as a potentially important direction for improving robustness, calibration, and generalization in future systems~\citep{shinn2023reflexion, madaan2023self}.

\begin{figure}[]
 \centering
 \includegraphics[width=\textwidth]{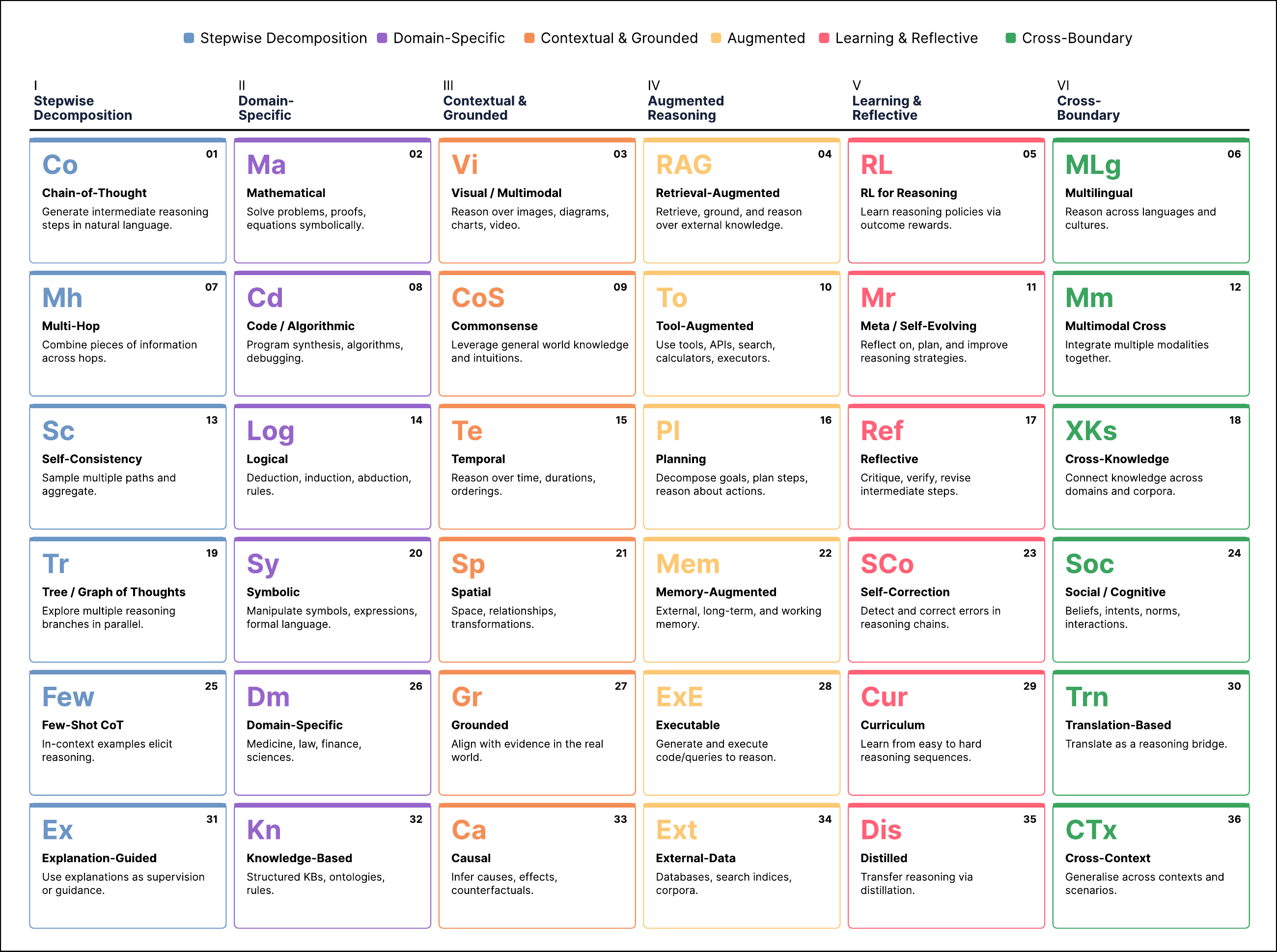}
 \caption{A taxonomy of LLM reasoning paradigms. Thirty-six families of methods are arranged into a $6\times6$ grid: columns group paradigms by how reasoning is composed (stepwise decomposition, domain-specific, contextual $\&$ grounded, augmented, learning $\&$ reflective, and cross-boundary), while rows situate them along a cognitive spectrum from training foundations up to high-level cognition.
 }
 \label{fig:taxonomy_reasoning2}
\end{figure}

The primary contributions of this survey are as follows:
\begin{itemize}

\item We develop a structured taxonomy of reasoning paradigms (Figure~\ref{fig:taxonomy_reasoning2}) in LLMs, covering approaches such as Chain-of-Thought, Multi-Hop, Mathematical, Commonsense, Multimodal, Retrieval-Augmented, Tool-Augmented, and Reinforcement Learning-based reasoning.

\item We examine methodological trends across the literature, including prompting strategies, architectural design choices, training paradigms, and evaluation benchmarks to assess reasoning capabilities.

\item We analyze recurring limitations and failure modes in LLM reasoning, including hallucinated reasoning traces, brittle multi-step inference, poor causal generalization, and sensitivity to prompting and task formulation.

\item We synthesize emerging research directions, including meta-reasoning, self-reflective and self-improving reasoning frameworks, multimodal reasoning, and socially grounded reasoning systems.

\item We discuss open research challenges and highlight directions toward more robust, interpretable, and generalizable reasoning systems.

\end{itemize}

\section{Methodology}
\label{sec:method}
This survey takes the canonical inspirations from a few established systematic review practices~\citep{moher2009preferred, keele2007guidelines, jesson2011doing}and follows a structured review methodology. This involves creating search parameters, selecting databases, applying inclusion and exclusion filters, and assessing the quality of each paper based on empirical rigor, relevance, reproducibility, and overall research impact.

\subsection{Search Process}
\label{method:I}
We began by identifying relevant keywords that represent the types of reasoning in LLMs. The broader terms include ``large language models'', ``reasoning'', ``chain of thought reasoning'', ``multi-hop reasoning'', and ``commonsense reasoning''.

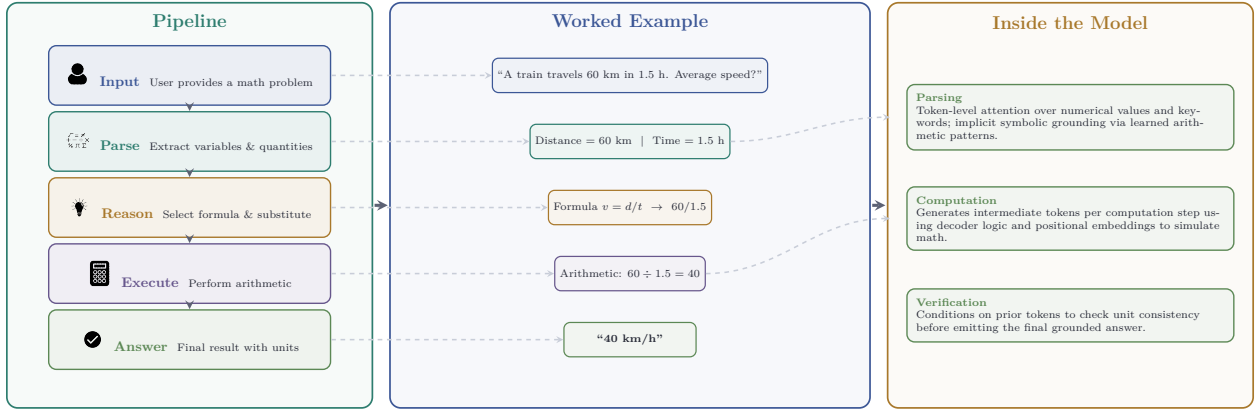
\begin{figure*}[htbp]
\centering
\definecolor{ink}{RGB}{38,41,54}
\definecolor{inksoft}{RGB}{92,97,115}
\definecolor{hair}{RGB}{200,205,216}
\definecolor{cBlue}{RGB}{60,87,150}
\definecolor{cTeal}{RGB}{46,125,111}
\definecolor{cAmber}{RGB}{168,120,43}
\definecolor{cViolet}{RGB}{103,87,140}
\definecolor{cGreen}{RGB}{92,136,85}
\resizebox{\textwidth}{!}{%
\begin{tikzpicture}[>=stealth,
  panel/.style={rounded corners=7pt, line width=1.1pt},
  stage/.style={draw=#1, line width=0.9pt, rounded corners=5pt, fill=#1!10,
    minimum width=2.5cm, minimum height=1.35cm, align=center, font=\small, text=ink},
  chip/.style={draw=#1, line width=0.8pt, rounded corners=4pt, fill=#1!8,
    minimum width=3.0cm, minimum height=0.78cm, align=center, font=\scriptsize, text=ink},
  det/.style={draw=cGreen, line width=0.9pt, rounded corners=5pt, fill=cGreen!8,
    align=left, font=\scriptsize, text=ink, inner sep=6pt},
  flow/.style={->, line width=1.5pt, draw=inksoft},
  hop/.style={->, line width=1pt, draw=hair, dashed},
  hdr/.style={font=\large\bfseries},
]
\draw[panel, draw=cTeal,   fill=cTeal!4]   (0,0)     rectangle (8.3,9.2);
\draw[panel, draw=cBlue,   fill=cBlue!4]   (8.7,0)   rectangle (19.6,9.2);
\draw[panel, draw=cAmber,  fill=cAmber!4]  (20.0,0)  rectangle (28.3,9.2);
\node[hdr, text=cTeal]  at (4.15,8.75) {Pipeline};
\node[hdr, text=cBlue]  at (14.15,8.75){Worked Example};
\node[hdr, text=cAmber] at (24.15,8.75){Inside the Model};

\node[stage=cBlue,   minimum width=6.4cm] (s1) at (4.15,7.55)
  {\includegraphics[width=0.55cm]{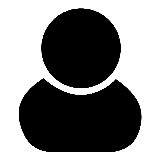}~~\textcolor{cBlue}{\textbf{Input}}~~\scriptsize User provides a math problem};
\node[stage=cTeal,   minimum width=6.4cm] (s2) at (4.15,6.05)
  {\includegraphics[width=0.55cm]{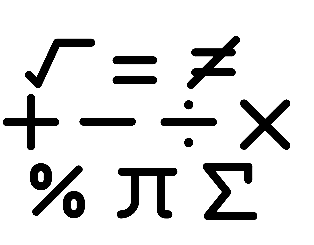}~~\textcolor{cTeal}{\textbf{Parse}}~~\scriptsize Extract variables \& quantities};
\node[stage=cAmber,  minimum width=6.4cm] (s3) at (4.15,4.55)
  {\includegraphics[width=0.55cm]{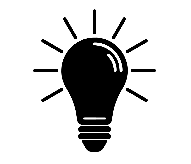}~~\textcolor{cAmber}{\textbf{Reason}}~~\scriptsize Select formula \& substitute};
\node[stage=cViolet, minimum width=6.4cm] (s4) at (4.15,3.05)
  {\includegraphics[width=0.55cm]{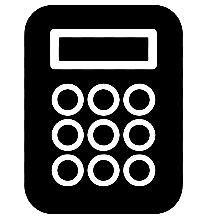}~~\textcolor{cViolet}{\textbf{Execute}}~~\scriptsize Perform arithmetic};
\node[stage=cGreen,  minimum width=6.4cm] (s5) at (4.15,1.55)
  {\includegraphics[width=0.55cm]{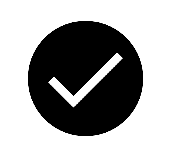}~~\textcolor{cGreen}{\textbf{Answer}}~~\scriptsize Final result with units};
\foreach \a/\b in {s1/s2,s2/s3,s3/s4,s4/s5}{\draw[flow] (\a) -- (\b);}

\node[chip=cBlue]   (e1) at (14.15,7.55) {``A train travels 60 km in 1.5 h. Average speed?''};
\node[chip=cTeal]   (e2) at (14.15,6.05) {Distance $=60$ km~~$\mid$~~Time $=1.5$ h};
\node[chip=cAmber]  (e3) at (14.15,4.55) {Formula $v=d/t$~~$\rightarrow$~~$60/1.5$};
\node[chip=cViolet] (e4) at (14.15,3.05) {Arithmetic: $60 \div 1.5 = 40$};
\node[chip=cGreen]  (e5) at (14.15,1.55) {\textbf{``40 km/h''}};
\foreach \s/\e in {s1/e1,s2/e2,s3/e3,s4/e4,s5/e5}{\draw[hop] (\s.east) -- (\e.west);}

\node[det, text width=7.0cm] at (24.15,6.6)
  {\textcolor{cGreen}{\textbf{Parsing}}\\[0pt]Token-level attention over numerical values and keywords; implicit symbolic grounding via learned arithmetic patterns.};
\node[det, text width=7.0cm] at (24.15,4.3)
  {\textcolor{cGreen}{\textbf{Computation}}\\[0pt]Generates intermediate tokens per computation step using decoder logic and positional embeddings to simulate math.};
\node[det, text width=7.0cm] at (24.15,2.1)
  {\textcolor{cGreen}{\textbf{Verification}}\\[0pt]Conditions on prior tokens to check unit consistency before emitting the final grounded answer.};
\draw[hop] (e2.east) to[out=0,in=180] (20.05,6.6);
\draw[hop] (e4.east) to[out=0,in=180] (20.05,4.3);

\draw[flow, line width=2pt] (8.35,4.6) -- (8.65,4.6);
\draw[flow, line width=2pt] (19.65,4.6) -- (19.95,4.6);
\end{tikzpicture}}
\caption{Mathematical reasoning flow. Left: the generic LLM reasoning pipeline. Middle: a math-specific worked example aligned stage-by-stage. Right: what happens inside the model at parsing, computation, and verification.}
\label{fig:math-reasoning-flow}
\end{figure*}

\noindent A list of more specific keywords was added for each reasoning type, such as for mathematical reasoning: ``mathematical reasoning'', ``equation solving by LLMs'', and ``symbolic reasoning''; for visual reasoning: ``image-text reasoning'' and ``visual commonsense reasoning''; for RAG-based reasoning: ``RAG models'' and ``information retrieval for reasoning''; for code reasoning: ``algorithmic reasoning'' and ``program synthesis with LLMs''; for tool-augmented reasoning: ``agentic reasoning'' and ``external tools.'' Then, we performed an expanded search using the combinations of Boolean operators and keyword variants to maximize coverage across reasoning paradigms and terminology differences across multiple venues. Conclusively, the keyword-driven approach covered both theoretical and applied work on LLM reasoning. The duplicate entries across databases were removed with manual supervision, and papers were screened based on title, abstract, introduction, and full-text relevance where necessary.

\subsection{Databases}
\label{method:II}

We conducted a systematic search across several academic databases and research repositories to identify peer-reviewed publications and high-impact preprints related to language modeling. The sources included:

\begin{itemize}
\item \textbf{IEEE Xplore}, for peer-reviewed conference proceedings, journal articles, and technical standards in computer science and artificial intelligence.

\item \textbf{Google Scholar}, for broad interdisciplinary coverage and citation-based discovery of influential journal articles, conference papers, and preprints.

\item \textbf{ACM Digital Library}, for peer-reviewed journals and conference proceedings in computing, artificial intelligence, and related fields.

\item \textbf{arXiv}, for recent preprints that provide early access to emerging research prior to formal peer review.

\item \textbf{SpringerLink}, for journal articles, conference proceedings, and scholarly books offering theoretical and methodological perspectives.

\item \textbf{Papers with Code}, for identifying research papers accompanied by publicly available implementations, datasets, and benchmark results, thereby supporting reproducibility and empirical comparison.

\end{itemize}

\subsection{Inclusion and Exclusion Criteria}
\label{method:III}
Given the deliberately broad scope of the search, the initial set of retrieved records contained both studies outside the survey’s focus and duplicate entries across databases. Thus, to improve the relevance and consistency of the final corpus, we applied predefined inclusion and exclusion criteria during the screening process as shown in Table~\ref{tab:inclusion-exclusion}.

\begin{table}[htbp]
\centering
\small
\begin{tabularx}{\textwidth}{ >{\raggedright\arraybackslash}p{0.16\textwidth} >{\raggedright\arraybackslash}X >{\raggedright\arraybackslash}X }
\toprule
\textbf{Dimension} & \textbf{Include} & \textbf{Exclude} \\
\midrule
Topic relevance & Addresses one or more listed reasoning types (e.g., CoT, Multi-Hop) & Does not primarily address a listed reasoning type \\
\addlinespace
Venue \& review status & Peer-reviewed journal or conference; preprints only if widely cited, adopted in benchmarks, later accepted, or needed for very recent developments & Non-peer-reviewed source (blog, white paper, preprint) without strong citation impact \\
\addlinespace
Recency & Published within the last 5 years, unless a seminal paper & Older than 5 years with no historical significance \\
\addlinespace
Methodological clarity & Methods described clearly (training schemes, models, experiments) & Lacks methodological detail or sufficient experimental validation \\
\addlinespace
Novelty & Presents novel techniques, theories, or applications & Substantially duplicates another included paper \\
\addlinespace
Empirical evidence & Reports results with performance metrics (accuracy, precision, recall) when applicable & Lacks technical depth or reports no empirical results \\
\bottomrule
\end{tabularx}
\caption{Inclusion and exclusion criteria applied to candidate papers.}
\label{tab:inclusion-exclusion}
\end{table}

\subsection{Quality Assessment}
\label{method:IV}

We developed eight Quality Assessment Criteria (QACs) to evaluate the relevance, rigor, clarity, and impact of each candidate study. Each criterion was rated qualitatively using a four-point scale: \textit{poor}, \textit{fair}, \textit{good}, or \textit{excellent}. The assessment supported consistent comparison across studies and informed their selection and prioritization for inclusion in the survey.

\begin{itemize}
\item \textbf{QAC1 (Contribution):} To what extent does the study advance the understanding or capability of the reasoning paradigm it investigates?

\item \textbf{QAC2 (Clarity):} Are the study's motivation, methodology, and findings presented clearly enough to be understood by readers outside the immediate subfield?

\item \textbf{QAC3 (Theoretical and practical relevance):} Does the study address both the theoretical foundations and practical implications of the proposed approach?

\item \textbf{QAC4 (Experimental rigor):} Are the experiments sufficiently rigorous in terms of dataset size, benchmark coverage, baseline selection, evaluation metrics, ablation studies, and experimental controls?

\item \textbf{QAC5 (Scholarly impact):} Has the study demonstrated substantial research influence, considering its citation count relative to its publication age?

\item \textbf{QAC6 (Research novelty):} Does the study investigate an under-explored problem, reasoning capability, methodology, or application area?

\item \textbf{QAC7 (Positioning within the literature):} Does the study adequately contextualize its contributions and distinguish them from related work?

\item \textbf{QAC8 (Reproducibility):} Are the experimental configuration, datasets, evaluation procedures, implementation details, and availability of code or data documented sufficiently to support replication?

\end{itemize}

\subsection{Data Collection and Analysis}
\label{method:V}
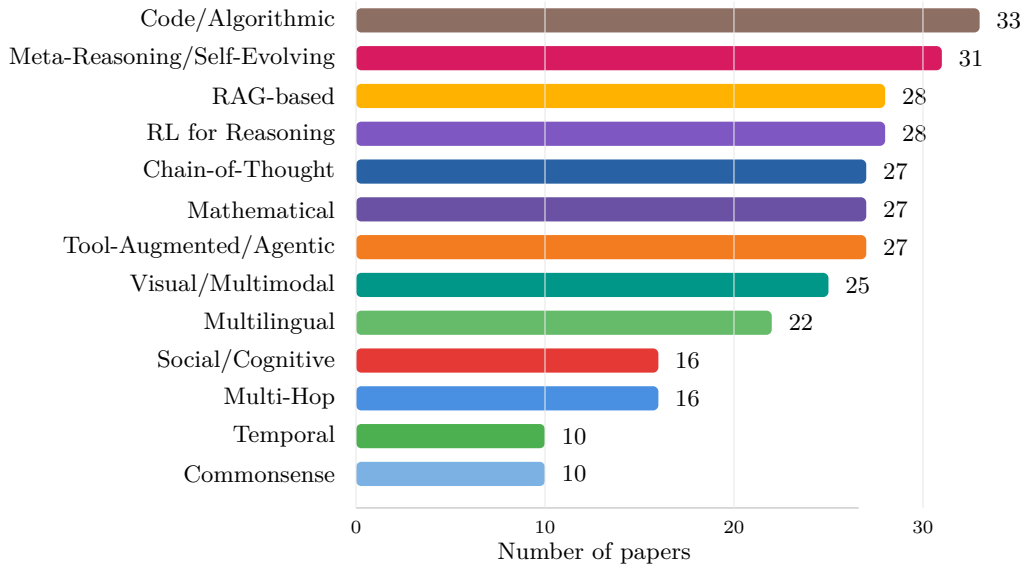
\begin{figure}[hbt]
\centering

\begin{tikzpicture}

\def\barh{0.32}
\def\gap{0.18}
\def\scale{0.25}


\definecolor{cotblue}{RGB}{40,98,165}
\definecolor{multihopblue}{RGB}{74,144,226}
\definecolor{mathviolet}{RGB}{106,81,163}
\definecolor{commonblue}{RGB}{123,178,227}

\definecolor{visualteal}{RGB}{0,150,136}
\definecolor{temporalgreen}{RGB}{76,175,80}
\definecolor{multilingualgreen}{RGB}{102,187,106}

\definecolor{ragamber}{RGB}{255,179,0}
\definecolor{toolorange}{RGB}{243,124,32}
\definecolor{codebrown}{RGB}{141,110,99}

\definecolor{rlpurple}{RGB}{126,87,194}
\definecolor{metamagenta}{RGB}{216,27,96}
\definecolor{socialred}{RGB}{229,57,53}

\newcommand{\barrow}[4]{
    \node[
        anchor=east,
        font=\small
    ] at (0,#3) {#1};

    \fill[
        #4,
        rounded corners=2pt
    ]
    (0.15,#3-\barh/2)
    rectangle
    ({0.15 + #2*\scale},#3+\barh/2);

    \node[
        anchor=west,
        font=\small
    ] at ({0.25 + #2*\scale},#3) {#2};
}


\barrow{Code/Algorithmic}{33}{6.0}{codebrown}

\barrow{Meta-Reasoning/Self-Evolving}{31}{5.5}{metamagenta}

\barrow{RAG-based}{28}{5.0}{ragamber}

\barrow{RL for Reasoning}{28}{4.5}{rlpurple}

\barrow{Chain-of-Thought}{27}{4.0}{cotblue}

\barrow{Mathematical}{27}{3.5}{mathviolet}

\barrow{Tool-Augmented/Agentic}{27}{3.0}{toolorange}

\barrow{Visual/Multimodal}{25}{2.5}{visualteal}

\barrow{Multilingual}{22}{2.0}{multilingualgreen}

\barrow{Social/Cognitive}{16}{1.5}{socialred}

\barrow{Multi-Hop}{16}{1.0}{multihopblue}

\barrow{Temporal}{10}{0.5}{temporalgreen}

\barrow{Commonsense}{10}{0.0}{commonblue}


\draw[gray!55]
(0.15,-0.45) -- (6.8,-0.45);

\foreach \x in {0,10,20,30}{
    
    \draw[gray!18]
    ({0.15+\x*\scale},-0.45)
    --
    ({0.15+\x*\scale},6.25);

    \node[
        font=\scriptsize
    ] at ({0.15+\x*\scale},-0.7) {\x};
}


\node[
    font=\small
] at (3.3,-1.05) {Number of papers};

\end{tikzpicture}
\caption{
Distribution of research papers by reasoning paradigm.
}
\label{fig:paper_distribution_bar}
\end{figure}

The collected papers were organized into a database annotated with metadata (title, authors, year, and primary focus) and categorized by reasoning type. Figure~\ref{fig:paper_distribution_bar} shows the distribution across reasoning types in LLMs, and Figure~\ref{fig:reasoning_heatmap} tracks publication volume per category over time. We analyzed the collection to identify recurring themes, methodological trends, and gaps in coverage.
\begin{figure*}[hbt]
\centering

\begin{tikzpicture}

\def\w{1.2}
\def\h{0.55}

\newcommand{\cell}[4]{
    \fill[blue!#4] (#1,#2) rectangle ++(\w,\h);
    \node at (#1+0.5*\w,#2+0.5*\h) {\scriptsize #3};
}

\node at (1.5,8.2) {\small $\leq$2022};
\node at (2.7,8.2) {\small 2023};
\node at (3.9,8.2) {\small 2024};
\node at (5.1,8.2) {\small 2025};

\node[anchor=east] at (0.9,7.7) {\scriptsize Chain-of-Thought};
\node[anchor=east] at (0.9,7.1) {\scriptsize Multi-Hop};
\node[anchor=east] at (0.9,6.5) {\scriptsize Mathematical};
\node[anchor=east] at (0.9,5.9) {\scriptsize Commonsense};
\node[anchor=east] at (0.9,5.3) {\scriptsize Visual/Multimodal};
\node[anchor=east] at (0.9,4.7) {\scriptsize Temporal};
\node[anchor=east] at (0.9,4.1) {\scriptsize Code/Algorithmic};
\node[anchor=east] at (0.9,3.5) {\scriptsize RAG-based};
\node[anchor=east] at (0.9,2.9) {\scriptsize Tool-Augmented};
\node[anchor=east] at (0.9,2.3) {\scriptsize RL for Reasoning};
\node[anchor=east] at (0.9,1.7) {\scriptsize Multilingual};
\node[anchor=east] at (0.9,1.1) {\scriptsize Meta-Reasoning};
\node[anchor=east] at (0.9,0.5) {\scriptsize Social/Cognitive};

\cell{1.0}{7.4}{0}{5}
\cell{2.2}{7.4}{0}{5}
\cell{3.4}{7.4}{11}{45}
\cell{4.6}{7.4}{16}{70}

\cell{1.0}{6.8}{0}{5}
\cell{2.2}{6.8}{0}{5}
\cell{3.4}{6.8}{15}{60}
\cell{4.6}{6.8}{1}{10}

\cell{1.0}{6.2}{0}{5}
\cell{2.2}{6.2}{0}{5}
\cell{3.4}{6.2}{8}{35}
\cell{4.6}{6.2}{19}{80}

\cell{1.0}{5.6}{1}{10}
\cell{2.2}{5.6}{4}{20}
\cell{3.4}{5.6}{3}{15}
\cell{4.6}{5.6}{2}{12}

\cell{1.0}{5.0}{1}{10}
\cell{2.2}{5.0}{0}{5}
\cell{3.4}{5.0}{6}{28}
\cell{4.6}{5.0}{18}{75}

\cell{1.0}{4.4}{6}{30}
\cell{2.2}{4.4}{2}{15}
\cell{3.4}{4.4}{9}{40}
\cell{4.6}{4.4}{5}{25}

\cell{1.0}{3.8}{1}{10}
\cell{2.2}{3.8}{5}{25}
\cell{3.4}{3.8}{11}{45}
\cell{4.6}{3.8}{16}{70}

\cell{1.0}{3.2}{0}{5}
\cell{2.2}{3.2}{5}{25}
\cell{3.4}{3.2}{11}{45}
\cell{4.6}{3.2}{12}{55}

\cell{1.0}{2.6}{0}{5}
\cell{2.2}{2.6}{2}{12}
\cell{3.4}{2.6}{11}{45}
\cell{4.6}{2.6}{14}{65}

\cell{1.0}{2.0}{0}{5}
\cell{2.2}{2.0}{0}{5}
\cell{3.4}{2.0}{8}{35}
\cell{4.6}{2.0}{20}{85}

\cell{1.0}{1.4}{6}{30}
\cell{2.2}{1.4}{2}{12}
\cell{3.4}{1.4}{9}{40}
\cell{4.6}{1.4}{5}{25}

\cell{1.0}{0.8}{0}{5}
\cell{2.2}{0.8}{1}{10}
\cell{3.4}{0.8}{15}{60}
\cell{4.6}{0.8}{15}{60}

\cell{1.0}{0.2}{2}{12}
\cell{2.2}{0.2}{3}{15}
\cell{3.4}{0.2}{4}{20}
\cell{4.6}{0.2}{7}{35}

\draw[thick] (1.0,0.2) rectangle (5.8,7.95);

\node[font=\bfseries] at (3.4,8.8)
{Publication Trends Across Reasoning Paradigms};

\end{tikzpicture}

\caption{Distribution of publications by reasoning paradigm and year.
Darker cells indicate higher publication volume.}
\label{fig:reasoning_heatmap}

\end{figure*}

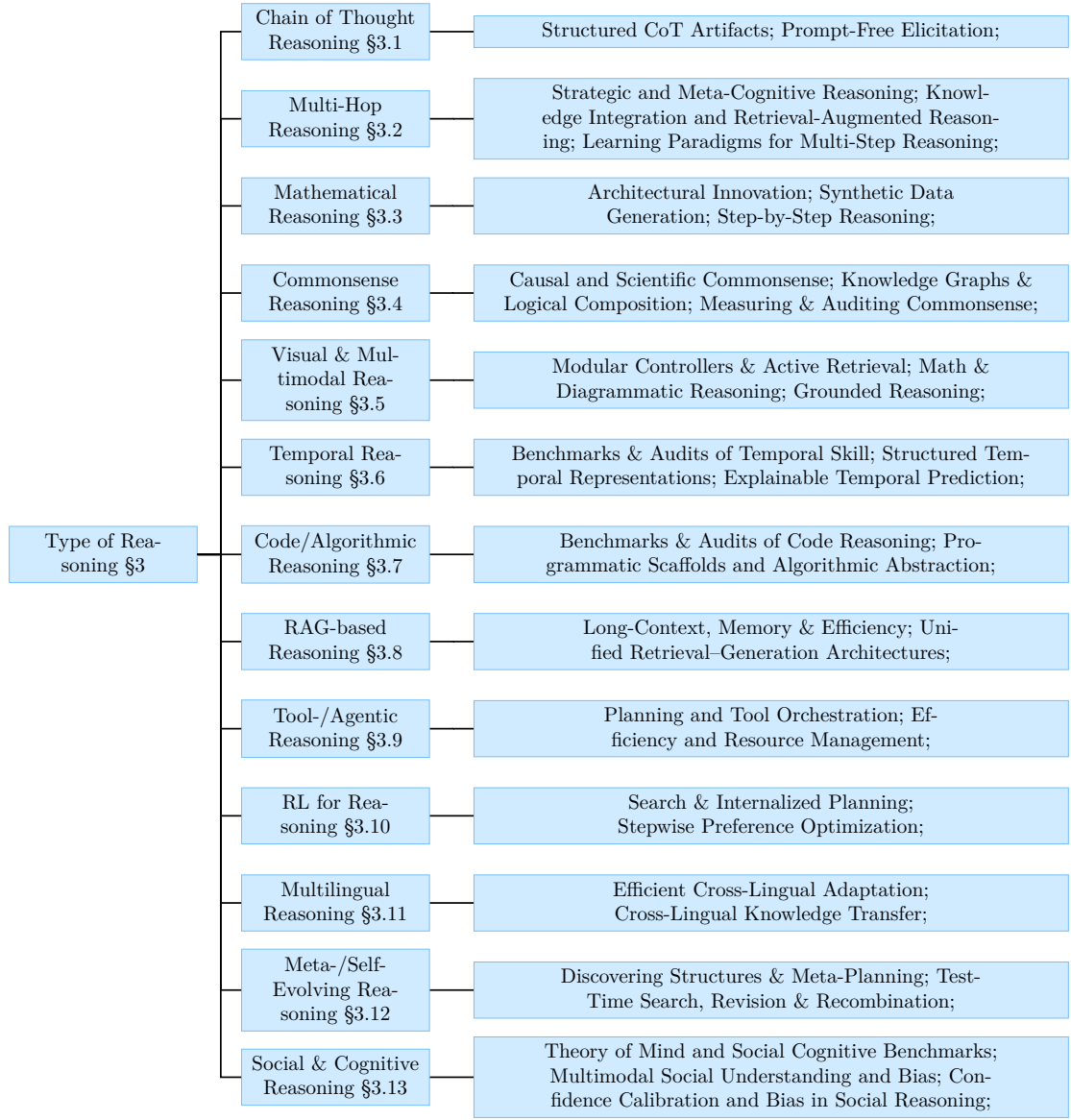
\begin{figure*}[h]
\centering
\scalebox{0.8}{
\begin{tikzpicture}[
 grow=right,
 level 1/.style={sibling distance=15mm, level distance=40mm},
 level 2/.style={sibling distance=10mm, level distance=38mm},
 every node/.style={
 align=center,
 font=\normalsize,
 rectangle, draw, thin, fill=lb,
 text width=3cm
 },
 edge from parent path={(\tikzparentnode.east) -- ++(4mm,0) |- (\tikzchildnode.west)},
 edge from parent/.style={draw, thick}
]

\node[fill=lb, draw=lbb, thin] {Type of Reasoning §\ref{sec:reason}}
 child {node[fill=lb, draw=lbb, thin] {Social \& Cognitive Reasoning §\ref{reason:XIII}}
 child[level distance=75mm] {
 node[fill=lb, draw=lbb, thin, text width=10cm] {Theory of Mind and Social Cognitive Benchmarks; Multimodal Social Understanding and Bias; Confidence Calibration and Bias in Social Reasoning;}
 }
 }
 child {node[fill=lb, draw=lbb, thin] {Meta-/Self-Evolving Reasoning §\ref{reason:XII}}
 child[level distance=75mm] {
 node[fill=lb, draw=lbb, thin, text width=10cm] {Discovering Structures \& Meta-Planning; Test-Time Search, Revision \& Recombination;}
 }
 }
 child {node[fill=lb, draw=lbb, thin] {Multilingual Reasoning §\ref{reason:XI}}
 child[level distance=75mm] {
 node[fill=lb, draw=lbb, thin, text width=10cm] {Efficient Cross-Lingual Adaptation; Cross-Lingual Knowledge Transfer;}
 }
 }
 child {node[fill=lb, draw=lbb, thin] {RL for Reasoning §\ref{reason:X}}
 child[level distance=75mm] {
 node[fill=lb, draw=lbb, thin, text width=10cm] {Search \& Internalized Planning; Stepwise Preference Optimization;}
 }
 }
 child {node[fill=lb, draw=lbb, thin] {Tool-/Agentic Reasoning §\ref{reason:IX}}
 child[level distance=75mm] {
 node[fill=lb, draw=lbb, thin, text width=10cm] {Planning and Tool Orchestration; Efficiency and Resource Management;}
 }
 }
 child {node[fill=lb, draw=lbb, thin] {RAG-based Reasoning §\ref{reason:VIII}}
 child[level distance=75mm] {
 node[fill=lb, draw=lbb, thin, text width=10cm] {Long-Context, Memory \& Efficiency; Unified Retrieval–Generation Architectures;}
 }
 }
 child {node[fill=lb, draw=lbb, thin] {Code/Algorithmic Reasoning §\ref{reason:VII}}
 child[level distance=75mm] {
 node[fill=lb, draw=lbb, thin, text width=10cm] {Benchmarks \& Audits of Code Reasoning; Programmatic Scaffolds and Algorithmic Abstraction;}
 }
 }
 child {node[fill=lb, draw=lbb, thin] {Temporal Reasoning §\ref{reason:VI}}
 child[level distance=75mm] {
 node[fill=lb, draw=lbb, thin, text width=10cm] {Benchmarks \& Audits of Temporal Skill; Structured Temporal Representations; Explainable Temporal Prediction;}
 }
 }
 child {node[fill=lb, draw=lbb, thin] {Visual \& Multimodal Reasoning §\ref{reason:V}}
 child[level distance=75mm] {
 node[fill=lb, draw=lbb, thin, text width=10cm] {Modular Controllers \& Active Retrieval; Math \& Diagrammatic Reasoning; Grounded Reasoning;}
 }
 }
 child {node[fill=lb, draw=lbb, thin] {Commonsense Reasoning §\ref{reason:IV}}
 child[level distance=75mm] {
 node[fill=lb, draw=lbb, thin, text width=10cm] {Causal and Scientific Commonsense; Knowledge Graphs \& Logical Composition; Measuring \& Auditing Commonsense;}
 }
 }
 child {node[fill=lb, draw=lbb, thin] {Mathematical Reasoning §\ref{reason:III}}
 child[level distance=75mm] {
 node[fill=lb, draw=lbb, thin, text width=10cm] {Architectural Innovation; Synthetic Data Generation; Step-by-Step Reasoning;}
 }
 }
 child {node[fill=lb, draw=lbb, thin] {Multi-Hop Reasoning §\ref{reason:II}}
 child[level distance=75mm] {
 node[fill=lb, draw=lbb, thin, text width=10cm] {Strategic and Meta-Cognitive Reasoning; Knowledge Integration and Retrieval-Augmented Reasoning; Learning Paradigms for Multi-Step Reasoning;}
 }
 }
 child {node[fill=lb, draw=lbb, thin] {Chain of Thought Reasoning §\ref{reason:I}}
 child[level distance=75mm] {
 node[fill=lb, draw=lbb, thin, text width=10cm] {Structured CoT Artifacts; Prompt-Free Elicitation;}
 }
 };
\end{tikzpicture}
}
\caption{Taxonomy of reasoning types covered in this survey, organized by category and subcategory.}
\label{fig:taxonomy}
\end{figure*}

Based on the collected literature, we developed a hierarchical taxonomy of reasoning paradigms and their associated sub-problems, as illustrated in Figure~\ref{fig:taxonomy}. This framework supports a large-scale synthesis of prior work across prompting strategies, model architectures, training objectives, and evaluation settings. The resulting taxonomy not only organizes the literature into interconnected reasoning categories but also reveals shared methodological patterns, recurring limitations, and emerging directions for future research.

\begin{figure}[]
 \centering
 \includegraphics[width=\textwidth]{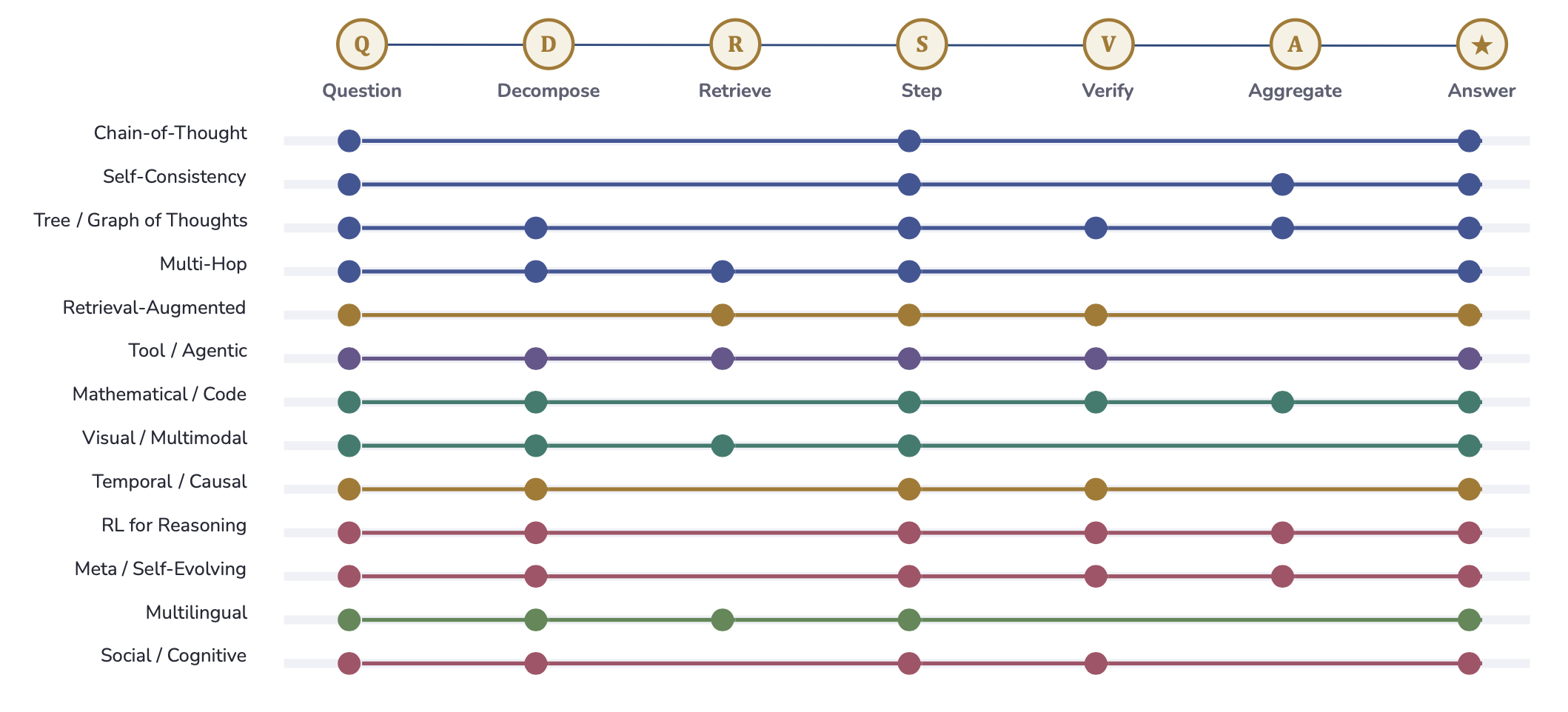}
 \caption{A shared skeleton (Question $\rightarrow$ Decompose $\rightarrow$ Retrieve $\rightarrow$ Step $\rightarrow$ Verify $\rightarrow$ Aggregate $\rightarrow$ Answer) with thirteen paradigm "lines" showing which stations each one stops at. Teaches that paradigms are commitments about which stages matter.
 }
 \label{fig:generation_pipeline}
\end{figure}

\section{Types of Reasoning}
\label{sec:reason}

Reasoning in LLMs is typically understood as the ability to perform structured transformations on a given information to derive conclusions, solve multi-step problems, generalize across contexts, or generate coherent intermediate inferences. Unlike classical symbolic AI systems, where reasoning is explicitly encoded through formal rules and logical operators, reasoning in LLMs emerges implicitly through large-scale language modeling and is observed primarily through model behavior during inference. It is often evaluated operationally, by examining whether models can sustain coherent chains of inference, integrate distributed information, perform abstraction, adapt across domains, or revise conclusions under new evidence.

Different reasoning paradigms emphasize different capabilities and computational structures. Some focus on explicit intermediate reasoning traces, such as Chain-of-Thought and Multi-Hop reasoning, while others emphasize domain-specific competence, including Mathematical and Code/Algorithmic reasoning. Additional paradigms address contextual and grounded understanding, such as Commonsense, Visual, Temporal, and Social reasoning. Recent work further extends reasoning through external augmentation mechanisms, including RAG and Tool-Augmented systems, as well as through reflective and adaptive frameworks such as RL-based and Meta-Reasoning approaches. This section organizes these paradigms into a unified taxonomy of reasoning behaviors studied in modern LLMs.

\subsection{Chain-of-Thought Reasoning}
\label{reason:I}

LLMs are typically trained using an autoregressive next-token prediction objective, where the model learns to generate text by predicting subsequent tokens from the previous tokens as context~\citep{brown2020language, radford2022language}. Early prompting strategies often encouraged models to produce direct answers without explicitly revealing intermediate reasoning steps. Chain-of-Thought (CoT) prompting~\citep{wei2022chain} introduced a simple but influential shift in inference behavior: instead of requesting only a final answer, the prompt encourages the model to generate intermediate natural-language reasoning steps, often through instructions such as ''think step-by-step''~\citep{kojima2022large}. 

By externalizing intermediate reasoning traces, CoT prompting encourages models to decompose complex tasks into smaller sub-problems, maintain intermediate state information, and perform sequential inference over multiple reasoning steps. This process often improves performance on tasks requiring arithmetic, symbolic manipulation, logical deduction, and compositional reasoning~\citep{wei2022chain,kojima2022large}. Despite these advantages, CoT reasoning does not guarantee the correctness or faithfulness of the generated reasoning process. The model continues to operate through probabilistic next-token generation, and intermediate reasoning traces may contain inconsistencies, hallucinated justifications, or post hoc rationalizations~\citep{lyu2023faithful, turpin2023language}. Nevertheless, the stepwise structure improves interpretability and provides opportunities for local verification and error analysis compared to direct-answer prompting.

\begin{figure}[htbp]
 \centering
 \includegraphics[width=0.9\textwidth]{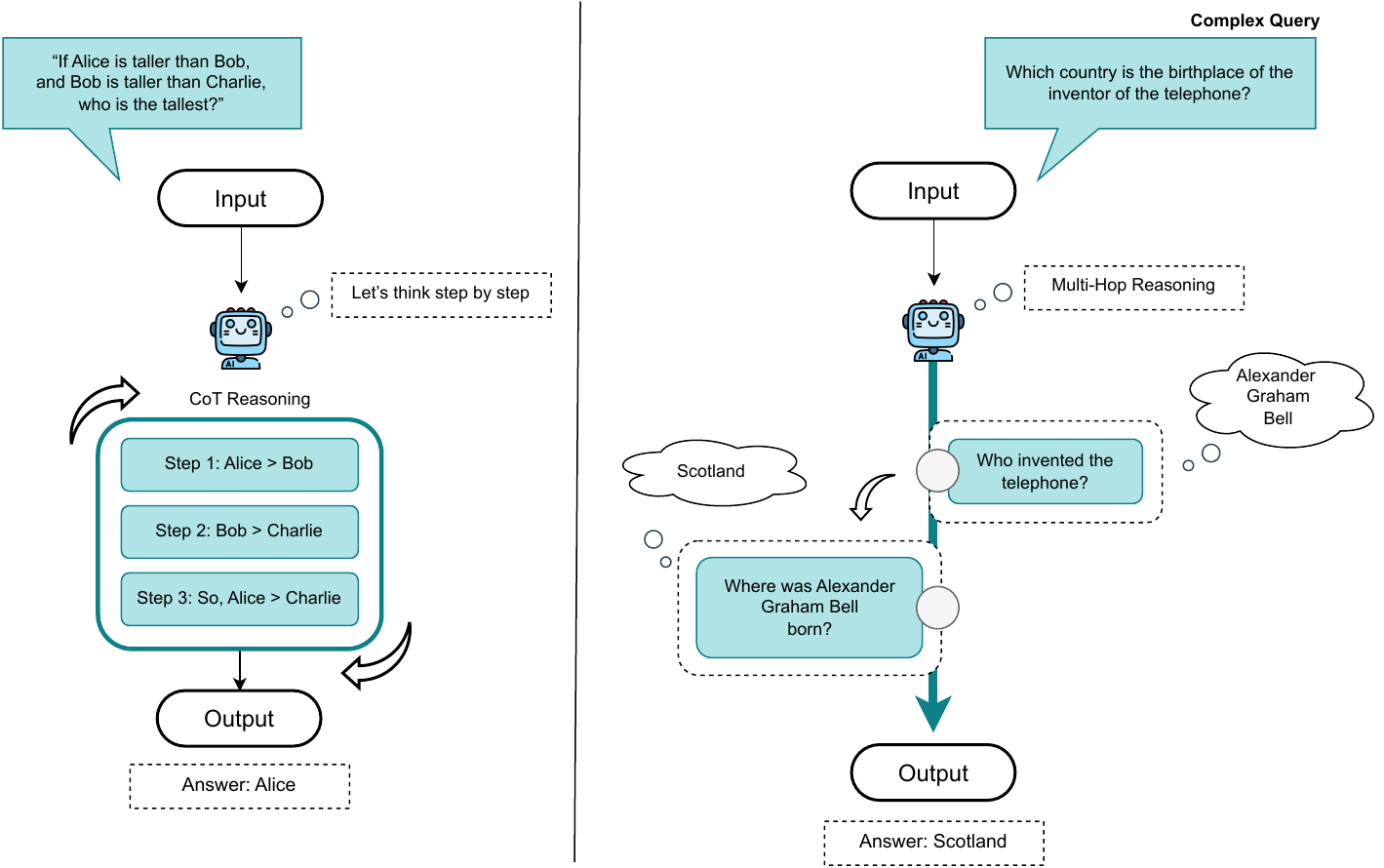}
 \caption{Chain-of-Thought and Multi-Hop reasoning interaction. Each reasoning hop produces an intermediate result that feeds into the next, linking stepwise CoT decomposition with the multi-hop structure required for complex queries.}
 \label{fig:cot_multihop_diagram}
\end{figure}

Figure~\ref{fig:cot_multihop_diagram} illustrates the relationship between Chain-of-Thought prompting and Multi-Hop reasoning, where successive inference steps build upon earlier intermediate conclusions. Tables~\ref{tab:cot-summary-part1} and~\ref{tab:cot-summary-part2} summarize representative experimental settings and benchmarks across the literature discussed in this subsection.

\subsubsection{Structured CoT Artifacts}

While standard Chain-of-Thought (CoT) prompting represents reasoning as free-form natural language, several recent approaches introduce explicit, structured artifacts that persist throughout the reasoning process. Instead of treating intermediate reasoning solely as text generation, these methods externalize state through tables, graphs, tagged evidence, or multimodal representations that the model can iteratively update and reference. Such structured reasoning spaces improve interpretability, grounding, and long-range consistency during multi-step inference.

\citep{wang2024chain} explores reasoning over dynamically updated tabular structures and proposes a framework in which the model incrementally edits and expands a table while reasoning, effectively transforming the table into a persistent workspace rather than a static input representation. Each modification corresponds to an intermediate reasoning step, enabling the model to maintain and revise structured state throughout inference. The approach demonstrates strong performance on tabular reasoning benchmarks, including WikiTQ~\citep{pasupat2015compositional} and TabFact~\citep{chen2019tabfact}.

Another line of work grounds CoT reasoning in structured knowledge representations. \citep{zhang2024chain} introduces Chain-of-Knowledge (CoK), which augments reasoning with explicit knowledge graph traversal and rule-guided inference. The framework constructs a reasoning dataset (KNOWREASON) by mining relational reasoning paths from knowledge graphs and employs a try-check-refine procedure in which the model proposes candidate reasoning paths, verifies them against graph structure, and iteratively improves them. By anchoring intermediate reasoning steps to entities and relations, CoK improves robustness on multi-hop and knowledge-intensive reasoning tasks. Other approaches focus on improving evidence attribution within reasoning traces. \citep{nguyen2025hot} proposes Highlighted Chain-of-Thought (HoT), where key facts from the input are explicitly tagged using lightweight XML-style markers and referenced throughout subsequent reasoning steps. These highlighted anchors help maintain factual grounding and improve traceability by exposing which intermediate conclusions depend on which evidence fragments.

Structured CoT has also been extended into multimodal reasoning settings. \citep{qin2025uni} propose Uni-CoT, a unified multimodal reasoning framework that combines textual and visual reasoning through hierarchical reasoning traces. The framework separates reasoning into macro-level planning and micro-level execution stages while jointly modeling evolving visual and textual states during inference. By integrating image understanding and generation within a unified reasoning loop, Uni-CoT demonstrates strong performance on multimodal reasoning benchmarks such as WISE~\citep{niu2025wise}, RISE~\citep{zhao2026envisioning}, and KRIS~\citep{wu2026kris}, although scalability and training complexity remain important challenges for large-scale deployment. 

\subsubsection{Prompt-Free Elicitation}
Models can also produce step-by-step reasoning without explicit CoT prompts. \citep{Xuezhi2024chain} shows that models can reveal step-by-step reasoning without special prompts. Instead of greedy decoding, they branch on the first token (top-k) and follow those alternatives, which often produce natural CoT traces. They then select the path using an answer-confidence margin (the gap between the top-1 and top-2 token probabilities over the final answer), which correlates with reliable CoT and lifts accuracy on math and commonsense tasks, no fine-tuning required. \citep{jin2024self} introduces ECHO, a simple way to turn auto-generated CoT demonstrations into a single, consistent style so models learn clearer patterns. The recipe is: cluster questions, draft initial rationales with zero-shot CoT, then iteratively regenerate each demo using the others as examples until the set ``harmonizes.'' This reduces inconsistency and cognitive load from overly diverse demos, and across arithmetic, commonsense, and symbolic tasks.

\subsubsection{Meta-Reasoning Engines}
Several systems now turn CoT into a controllable process: planners that decide when and what to think, template libraries that can be composed, and self-questioning loops that decompose problems. For the broader treatment of meta-reasoning including test-time search, capacity-matched distillation, process-aware training, and meta-judging, see Section~\ref{reason:XII}.

\begin{figure}[htbp]
\centering
\definecolor{ink}{RGB}{38,41,54}
\definecolor{inksoft}{RGB}{92,97,115}
\definecolor{hair}{RGB}{200,205,216}
\definecolor{cAmber}{RGB}{168,120,43}
\definecolor{cGreen}{RGB}{92,136,85}
\definecolor{cViolet}{RGB}{103,87,140}
\definecolor{cRose}{RGB}{169,79,103}
\begin{tikzpicture}[
  sysbox/.style={rectangle, draw=#1, line width=0.9pt, rounded corners=5pt, fill=#1!8,
    minimum height=1.2cm, align=center, font=\scriptsize, text width=2.3cm, text=ink, inner sep=4pt},
  axis/.style={-{Stealth[length=2.5mm]}, draw=inksoft, line width=1.1pt},
]
\draw[axis] (0,0) -- (13,0);
\node[font=\scriptsize\itshape, text=inksoft, anchor=north] at (0.2,-0.18) {Prompt-only};
\node[font=\scriptsize\itshape, text=inksoft, anchor=north] at (12.8,-0.18) {Trained search};
\node[sysbox=cGreen]  (sq)  at (1.5,1.45)  {\textcolor{cGreen}{\textbf{SQuARE}}\\[0pt]Sub-question prompting};
\node[sysbox=cAmber]  (bot) at (4.8,1.45)  {\textcolor{cAmber}{\textbf{Buffer of Thoughts}}\\[0pt]Pattern retrieval};
\node[sysbox=cViolet] (rf)  at (8.3,1.45)  {\textcolor{cViolet}{\textbf{ReasonFlux}}\\[0pt]Template planning via RL};
\node[sysbox=cRose]   (mc)  at (11.8,1.45) {\textcolor{cRose}{\textbf{Meta-CoT}}\\[0pt]Internalized MCTS / A*};
\foreach \n/\x/\c in {sq/1.5/cGreen, bot/4.8/cAmber, rf/8.3/cViolet, mc/11.8/cRose}{
  \draw[draw=hair, line width=0.7pt] (\n.south) -- (\x,0);
  \filldraw[\c] (\x,0) circle (2.4pt);
}
\end{tikzpicture}
\caption{Meta-reasoning engines arranged by training requirement. SQuARE operates through prompting alone. Buffer of Thoughts adds a reusable pattern library. ReasonFlux learns template plans via hierarchical RL. Meta-CoT internalizes the full search process through reinforcement learning on linearized traces.}
\label{fig:meta-reasoning-spectrum}
\end{figure}
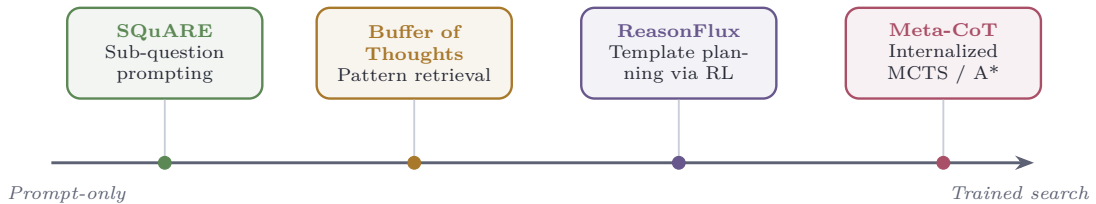

\noindent Table~\ref{tab:meta-reasoning-engines} compares mechanisms and reported strengths of the prominent meta-reasoning systems in the literature. Taken together, these systems (Figure~\ref{fig:meta-reasoning-spectrum}) turn CoT from a transcript into a procedure: choose a plan, reuse prior know-how, ask and check along the way. They add overhead but buy more reliable and inspectable reasoning that transfers across tasks.

\begin{table}[htbp]
\centering
\small
\begin{tabularx}{\textwidth}{ >{\raggedright\arraybackslash}p{0.18\textwidth} >{\raggedright\arraybackslash}p{0.22\textwidth} >{\raggedright\arraybackslash}X >{\raggedright\arraybackslash}p{0.22\textwidth} }
\toprule
\textbf{System} & \textbf{Core mechanism} & \textbf{Training / inference recipe} & \textbf{Reported strengths} \\
\midrule
SQuARE \citep{fleischer2025square} & Self-generated sub-questions before the final answer & Prompting only; drops into existing LLMs without fine-tuning & Beats standard CoT and rephrase-and-respond on TriviaQA~\citep{joshi-etal-2017-triviaqa}, HotpotQA~\citep{yang2018hotpotqa}, ASQA~\citep{stelmakh2022asqa} \\
\addlinespace
Buffer of Thoughts \citep{yang2024buffer} & Retrieves reusable reasoning patterns from a meta buffer & Manager component refines and grows the buffer as problems are solved & Higher accuracy, lower token and time cost on Game of 24~\citep{yao2023tree}, Geometric Shapes~\citep{suzgun2023challenging}, Checkmate~\citep{srivastava2023beyond} \\
\addlinespace
ReasonFlux \citep{yang2025reasonflux} & Plans with a library of high-level thought templates & Hierarchical RL over template sequences; retrieves and adapts templates at inference & Gains on challenging math benchmarks over strong baselines \\
\addlinespace
Meta-CoT \citep{xiang2025towards} & Learns the search process behind a chain (exploration, verification, backtracking) & Generate MCTS/A* traces, add process-level supervision, fine-tune with instruction tuning and RL on linearized traces & Handles problems where linear, step-by-step CoT breaks down \\
\bottomrule
\end{tabularx}
\caption{Meta-reasoning engines that turn CoT into a controllable procedure, ordered from lightweight prompting to full search internalization.}
\label{tab:meta-reasoning-engines}
\end{table}

\subsubsection{LongCoT Bootstrapping}
In this method, the pattern is to start with short chains and extend them. It combines in-context learning with supervised fine-tuning and online updates, adding checks so the model can maintain and correct extended reasoning. \citep{pang2025bolt} presents \textbf{BOLT}, a procedure to teach a regular instruction model to produce long chains of thought without copying a teacher model or paying for heavy labels. It follows three simple steps: (1) seed a few example solutions in context to make the model generate long reasoning traces, (2) fine-tune on this synthetic set, and (3) finally keep training online with a reward model that prefers good final answers. They scaled this recipe from 7B to 70B, and saw strong gains on Arena-Hard~\citep{li2024crowdsourced}, MT-Bench~\citep{bai2024mt}, WildBench~\citep{lin2025wildbench}, ZebraLogic~\citep{lin2025zebralogic}, and MATH500~\citep{lin2025step}. \citep{yeo2025demystifying} investigated how long reasoning traces actually arise and how to train them in practice. Through controlled supervised fine-tuning and reinforcement learning, they show that longer chains benefit from a good supervised start, careful reward design that stabilizes length, and verifiable signals even when mined from noisier web data. The takeaway is practical guidance for building traces that branch, check, and correct, which helps especially with math and other STEM-style tasks. The open problem is how to balance length, cost, and reliability as contexts and tasks grow.

\subsubsection{Efficient \& Robust CoT}
Four families of CoT methods target efficiency and robustness, each trading off different aspects of the chain:
\begin{itemize}
\item \textbf{Dense compression.} CCoT~\citep{cheng2024compressed} replaces text rationales with a small number of dense ``contemplation tokens,'' cutting tokens and latency while preserving reasoning content. LightThinker~\citep{zhang2025lightthinker} compresses on the fly, turning each intermediate thought into a few gist tokens and continuing from the summaries, with accuracy close to full CoT on GSM8K~\citep{cobbe2021training}, MMLU~\citep{hendrycks2021measuring}, GPQA~\citep{rein2023gpqa}, and BBH~\citep{suzgun2023challenging}.
\item \textbf{Concise drafting.} Chain of Draft~\citep{xu2025chain} has the model write tiny drafts at each step instead of full sentences, using as little as 7.6\% of standard CoT tokens with much lower latency. It keeps reasoning in natural language while making it compact and cheap.
\item \textbf{Per-question budgeting.} Token-Budget-Aware Reasoning~\citep{han2024token} treats chain length as a resource, estimating a sensible token budget per question. Across math tasks, this cuts CoT tokens by roughly two-thirds with only small accuracy drops. They note a token-elasticity effect: budgets that are too tight can backfire, so the method targets a sweet spot.
\item \textbf{Continuous-space reasoning.} Soft Thinking~\citep{zhang2025soft} replaces discrete tokens with probability-weighted mixtures of embeddings, enabling implicit exploration of multiple reasoning paths in a single step. It shows modest pass@1 gains on math and coding benchmarks with reduced token usage, though interpretability may be limited.
\end{itemize}

\begin{figure}[htbp]
\centering
\begin{tikzpicture}[
  box/.style={rectangle, draw, rounded corners=3pt, minimum height=0.7cm, align=center, font=\scriptsize, inner sep=4pt},
  arr/.style={-{Stealth[length=2mm]}, thick},
  zone/.style={rectangle, draw=gray!40, dashed, rounded corners=4pt, inner sep=8pt},
  annot/.style={font=\tiny, gray!60},
]
\draw[arr, gray] (0,0) -- (11.2,0) node[right, font=\small] {Lower cost (fewer tokens)};
\draw[arr, gray] (0,0) -- (0,6.5) node[above, font=\small] {Higher robustness};
\draw[zone, fill=green!4] (0.3,0.3) rectangle (10.5,3.4);
\node[annot, anchor=north east] at (10.3,3.25) {\textit{Efficiency methods}};
\node[box, fill=red!10] (full) at (1.2,1.2) {Full CoT};
\node[box, fill=yellow!15] (ccot) at (3.8,1.2) {CCoT};
\node[box, fill=orange!12] (st) at (6.4,1.2) {Soft Thinking};
\node[box, fill=green!12] (cod) at (9.2,1.2) {Chain of Draft};
\node[box, fill=yellow!15] (lt) at (3.8,2.6) {LightThinker};
\node[box, fill=green!12] (tb) at (7.0,2.6) {Token Budget};
\draw[->, thick, dashed, gray!40] (full) -- (ccot);
\draw[->, thick, dashed, gray!40] (ccot) -- (st);
\draw[->, thick, dashed, gray!40] (st) -- (cod);
\draw[zone, fill=blue!4] (0.3,3.8) rectangle (6.5,5.8);
\node[annot, anchor=north east] at (6.3,5.65) {\textit{Robustness methods}};
\node[box, fill=blue!12] (db) at (2.0,4.8) {DeltaBench\\(audit)};
\node[box, fill=blue!12] (cd) at (5.0,4.8) {CD-CoT\\(denoise)};
\node[box, fill=purple!10, line width=0.8pt] (goal) at (9.2,5.8) {\textbf{Goal:}\\cheap + audited};
\draw[->, thick, dashed, green!40!black] (cod) .. controls (10.5,3.2) .. (goal);
\draw[->, thick, dashed, blue!40!black] (cd) .. controls (7.5,5.6) .. (goal);
\end{tikzpicture}
\caption{Two dimensions of CoT improvement. The \emph{efficiency zone} (bottom) maps methods that reduce token cost: CCoT and LightThinker compress rationales into dense tokens, Soft Thinking replaces discrete tokens with embedding mixtures, Token Budget caps per-question length, and Chain of Draft writes minimal drafts. The \emph{robustness zone} (upper left) maps methods that audit or clean chains: DeltaBench benchmarks error detection in long traces, and CD-CoT denoises corrupted demonstrations. Neither dimension alone suffices; the goal (top right) is chains that are both cheap and self-auditing.}
\label{fig:cot-efficiency-robustness}
\end{figure}
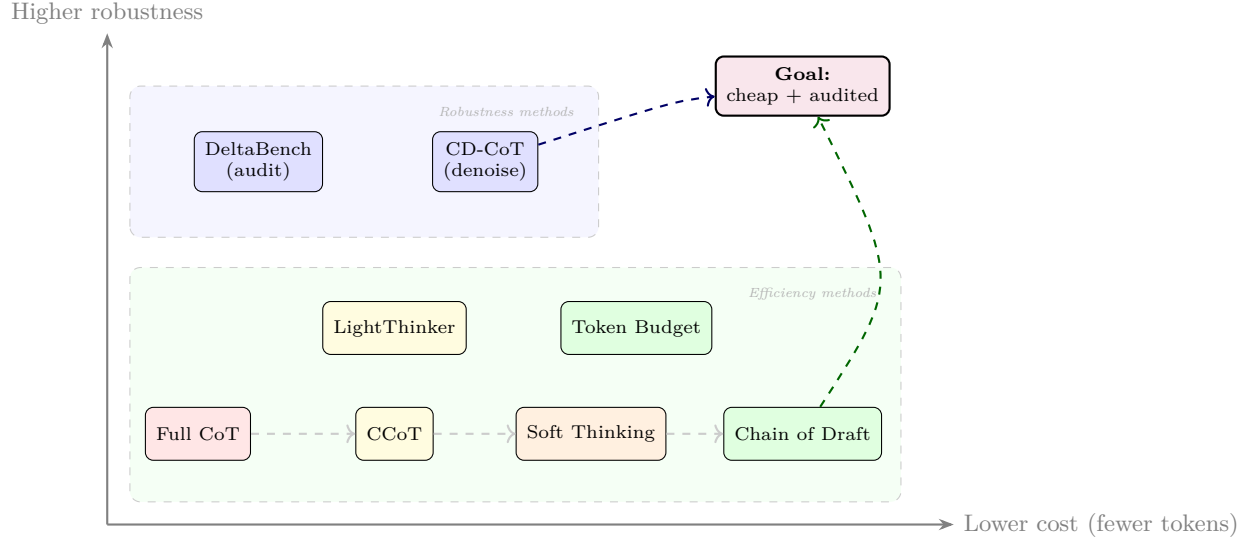

\noindent With cheaper and faster chains in hand, a second line of work asks whether these chains are trustworthy. \citep{he2025can} introduces DeltaBench, a benchmark that tests whether models can find and explain errors inside long chains of thought. It gathers traces from o1-style models across math, programming, science, and general reasoning, splits each trace into sections, and labels each for usefulness, correctness, and reflection. Both critic LLMs and process reward models reach only modest F1 at spotting bad sections; fundamental mistakes and redundancy are common, while effective reflection is rare. Similarly, \citep{zhou2024can} builds NoRa, a testbed where demonstrations contain irrelevant or inaccurate thoughts. Common models and robustness tricks drop sharply under this noise, and self-consistency or self-correction helps only a little. Their CD-CoT, a contrastive denoising recipe that uses one clean example to rephrase and filter noisy rationales, produces a consistent lift across math, symbolic, and commonsense settings with average gains around 17.8\% over the base model. Figure~\ref{fig:cot-efficiency-robustness} maps these two dimensions.

\subsubsection{CoT Evaluation \& Adaptive Data Generation}

Recent work on Chain-of-Thought (CoT) reasoning increasingly couples evaluation and data generation into a unified feedback process, where evaluation identifies systematic reasoning failures and targeted data generation is used to address those weaknesses. This shift reflects a broader trend toward treating reasoning not only as an inference-time behavior, but also as a measurable and optimizable capability.

For CoT evaluation, \citep{bai2024longbench} introduces LongBench v2, a benchmark consisting of 503 multiple-choice tasks spanning six long-context reasoning settings, including single- and multi-document question answering, long-context in-context learning, dialogue history understanding, repository-level code reasoning, and structured data analysis. Context lengths range from 8k to 2M words, enabling systematic evaluation of long-range reasoning behavior. Their results show that explicit inference-time reasoning substantially improves performance in long-context settings, suggesting that CoT-style decomposition becomes increasingly important as contextual complexity grows. Focusing on visualizing and diagnosing reasoning trajectories, \citep{zhou2025landscape} proposes Landscape of Thoughts, a framework that projects intermediate reasoning steps into a two-dimensional reasoning landscape for multiple-choice tasks. Each intermediate state is represented as a point in the reasoning trajectory, making error propagation and reasoning collapse visually interpretable. The same trajectory features are further used to construct lightweight verification mechanisms that select more reliable reasoning paths and improve accuracy without modifying model parameters. Reasoning evaluation has also been formalized through capability-oriented frameworks. \citep{chen2024unlocking} introduces the Reasoning Boundary Framework (RBF), which models CoT's reasoning ability as a measurable capability frontier. The framework estimates the maximum difficulty level a model can solve reliably, studies how reasoning boundaries interact across different skill domains, and categorizes tasks into solvable, borderline, and out-of-capacity regions. This perspective reframes reasoning evaluation as a continuous capability estimation problem rather than a binary success/failure metric. 

To investigate adaptive reasoning data generation, \citep{yu2025rethinking} proposes a pipeline in which the target model first estimates problem difficulty, after which a stronger teacher model generates concise and verified reasoning traces for selected examples. The resulting dataset is balanced across difficulty levels and populated with validated Chain-of-Thought solutions generated using DeepSeek R1. By constructing ability-aligned reasoning corpora containing approximately 2k examples per domain, the approach reduces annotation cost while improving fine-tuning efficiency on mathematical and code reasoning tasks.

\begin{figure}[htbp]
\centering
\definecolor{ink}{RGB}{38,41,54}
\definecolor{inksoft}{RGB}{92,97,115}
\definecolor{cTeal}{RGB}{46,125,111}
\definecolor{cAmber}{RGB}{168,120,43}
\definecolor{cGreen}{RGB}{92,136,85}
\definecolor{cRose}{RGB}{169,79,103}
\begin{tikzpicture}[
  node distance=1.2cm,
  decision/.style={diamond, draw=cAmber, line width=0.9pt, fill=cAmber!10, aspect=1.6,
    text width=2.6cm, align=center, inner sep=2pt, font=\small, text=ink},
  block/.style={rectangle, draw=cGreen, line width=0.9pt, rounded corners=5pt, fill=cGreen!8,
    text width=3cm, align=center, minimum height=0.95cm, font=\small, text=ink},
  blockt/.style={rectangle, draw=cTeal, line width=0.9pt, rounded corners=5pt, fill=cTeal!8,
    text width=3cm, align=center, minimum height=0.95cm, font=\small, text=ink},
  blockr/.style={rectangle, draw=cRose, line width=0.9pt, rounded corners=5pt, fill=cRose!8,
    text width=3cm, align=center, minimum height=0.95cm, font=\small, text=ink},
  arr/.style={-{Stealth[length=2.5mm]}, draw=inksoft, line width=1pt},
  elab/.style={font=\scriptsize, text=inksoft, fill=white, inner sep=1.5pt},
]
\node[decision] (q1) {Task has symbolic or multi-step structure?};
\node[decision, right=2.8cm of q1] (q2) {Verbal reasoning fits the task?};
\node[block,  below=1.3cm of q1] (yes1) {\textcolor{cGreen}{\textbf{Use CoT}}\\[0pt]{\scriptsize strong gains}};
\node[blockr, below=1.3cm of q2] (no2)  {\textcolor{cRose}{\textbf{Skip CoT}}\\[0pt]{\scriptsize use tools or direct answering}};
\node[blockt, right=2.8cm of q2] (yes2) {\textcolor{cTeal}{\textbf{Use CoT}}\\[0pt]{\scriptsize with scaffolded steps}};
\draw[arr] (q1) -- node[elab, left] {Yes} (yes1);
\draw[arr] (q1) -- node[elab, above] {No} (q2);
\draw[arr] (q2) -- node[elab, left] {No} (no2);
\draw[arr] (q2) -- node[elab, above] {Yes} (yes2);
\end{tikzpicture}
\caption{When to apply Chain-of-Thought. CoT produces strong gains on symbolic and multi-step tasks. For non-symbolic tasks, it helps only when verbal reasoning fits; otherwise, tools or direct answering may perform equally well or better~\citep{sprague2024cot,liu2024mind}.}
\label{fig:cot-decision}
\end{figure}
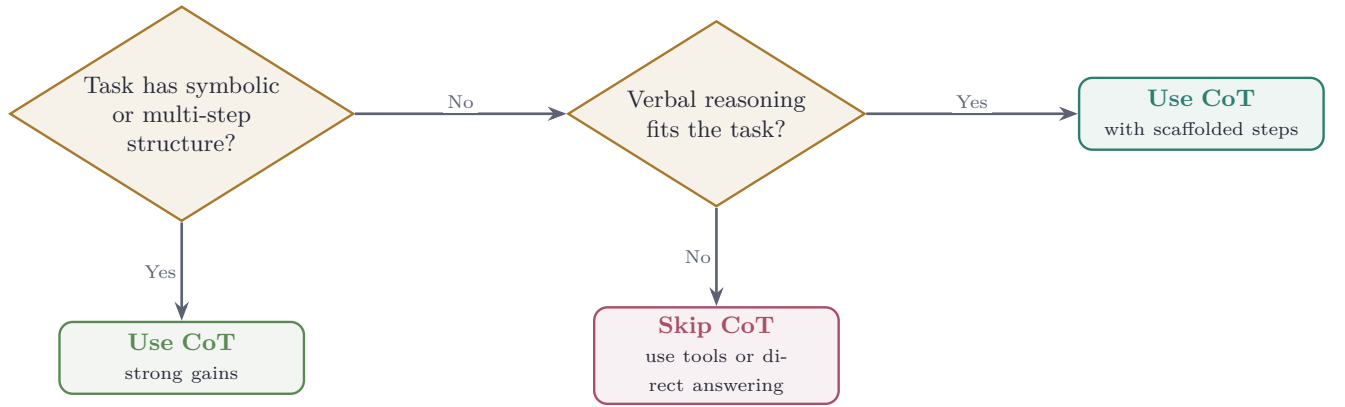

\subsubsection{Limitations of CoT}

Although Chain-of-Thought (CoT) prompting often improves reasoning performance, recent work shows that its benefits are highly task-dependent and that longer reasoning traces do not necessarily imply better reasoning quality.

A large-scale meta-analysis by \citep{sprague2024cot} examines over 100 papers and 20 datasets to evaluate the practical effectiveness of CoT prompting across diverse reasoning tasks. Their results show consistent and substantial improvements on mathematical and symbolic reasoning benchmarks, while gains on many non-symbolic tasks are limited or negligible. In several settings, direct-answer prompting performs comparably to CoT prompting, suggesting that explicit reasoning traces are most beneficial when tasks contain strong compositional or symbolic structure. The study further argues that external tools or symbolic solvers may be preferable when problems require precise formal manipulation beyond natural-language reasoning alone.

Other work examines whether explicit verbalization itself can interfere with reasoning. Drawing inspiration from findings in cognitive psychology, \citep{liu2024mind} identifies task settings in which ``thinking out loud'' reduces performance. They show that CoT prompting can negatively affect tasks involving implicit grammar induction, face description, and rule learning with exceptions, while producing mixed outcomes on other cognitively implicit tasks. These results suggest that some forms of reasoning are poorly aligned with explicit natural-language decomposition and that indiscriminate use of CoT prompting may introduce unnecessary cognitive constraints during inference.

The structure and length of reasoning traces also play an important role. \citep{jin2024impact} investigates how the number of intermediate reasoning steps affects downstream performance and finds that increasing reasoning depth often improves accuracy, even when intermediate steps contain imperfections or partial inconsistencies. The effect becomes more pronounced on difficult reasoning tasks, indicating that the organizational structure of reasoning traces may matter as much as the factual correctness of individual steps.

Related findings are reported by \citep{li2025llms}, who show that models can acquire long-form reflective reasoning behaviors using relatively modest datasets and parameter-efficient fine-tuning methods. Their experiments suggest that models learn primarily from structural properties of reasoning traces, including ordering, coherence, revision behavior, and hierarchical organization, rather than from the precise semantic content of every intermediate step. Disrupting the structural consistency of reasoning chains produces substantially larger performance degradation than perturbing local reasoning details.

Taken together, these studies suggest several practical guidelines for CoT reasoning:
\begin{itemize}
\item CoT is most effective for tasks with strong symbolic, compositional, or multi-step structure.
\item Explicit verbal reasoning may be counterproductive for tasks that rely on implicit cognition.
\item The organization and coherence of reasoning traces often matter more than the literal correctness of every intermediate step.
\item Longer reasoning chains can improve performance, but increased verbosity alone does not guarantee faithful or correct reasoning.
\end{itemize}

\begin{table}[htbp]
\centering
\small
\begin{tabularx}{\textwidth}{ >{\raggedright\arraybackslash}p{0.25\textwidth} >{\raggedright\arraybackslash\hsize=0.8\hsize\linewidth=\hsize}X >{\raggedright\arraybackslash\hsize=0.8\hsize\linewidth=\hsize}X >{\raggedright\arraybackslash\hsize=1.4\hsize\linewidth=\hsize}X }
\toprule
\textbf{Author} & \textbf{Datasets} & \textbf{Models} & \textbf{Evaluation Setup} \\
\midrule

\multicolumn{4}{l}{\textbf{Structured CoT Artifacts}} \\

Wang et al.~\citep{wang2024chain} & WikiTQ, FeTaQA, TabFact & PaLM 2-S1, GPT-3.5, LLaMA 2 & Denotation Acc., BLEU, ROUGE, Binary Classification Acc. \\

Zhang et al.~\citep{zhang2024chain} & KNOWREASON (Wikidata5M) & Mistral-7B-instruct, LLaMA3-8B-instruct & Exact Match Accuracy (ID/OOD), rule length splits \\

Nguyen et al.~\citep{nguyen2025hot} & 20 tasks (GSM8K, AQUA, StrategyQA, etc.) & GPT-4o, Gemini 1.5, LLaMA 3.1 & Accuracy, hallucination score via SelfCheckGPT \\

Qin et al.~\citep{qin2025uni} & Uni-CoT Multi-modal CoT Dataset & Uni-CoT (fine-tuned on Bagel) & Image generation on GenEval and WISE; image understanding on MME, MMMU, MathVista, MMBench, and Jigsaw-R1 \\

\midrule
\multicolumn{4}{l}{\textbf{Prompt-Free Elicitation}} \\

Wang et al.~\citep{Xuezhi2024chain} & GSM8K, MultiArith, Year Parity, BBH & PaLM-2 (X/S/M/L), Mistral-7B, Gemma-7B & Top-k+greedy decoding, accuracy across QA tasks \\

Jin et al.~\citep{jin2024self} & SingleEq, AddSub, GSM8K, SVAMP, etc. & GPT-3.5-Turbo, Mixtral-8x7B-Instruct & Accuracy across arithmetic, symbolic, commonsense tasks \\

\midrule
\multicolumn{4}{l}{\textbf{Meta-Reasoning Engines}} \\

Yang et al.~\citep{yang2025reasonflux} & AIME24, MATH, AMC, OlympiadBench & ReasonFlux, Qwen2.5, GPT-4o, LLaMA3.1 & Pass@1, consistency, scaling behavior \\

Yang et al.~\citep{yang2024buffer} & Game of 24, BBH, BIG-Bench, MGSM & GPT-4, LLaMA3 (8B/70B) & Accuracy, inference time, success rate \\

Fleischer et al.~\citep{fleischer2025square} & TriviaQA, HotpotQA, ASQA & LLaMA-3.1/3.2, GPT-4o & SubEM, Recall-EM, aggregation ablations \\

\midrule
\multicolumn{4}{l}{\textbf{LongCoT Bootstrapping}} \\

Pang et al.~\citep{pang2025bolt} & Arena-Hard, MT-Bench, WildBench, MATH500 & BOLT-trained LLaMA-3.1 (7B/8B/70B) & Elo rating, reasoning diversity, multi-turn score \\

Yeo et al.~\citep{yeo2025demystifying} & MATH, AIME24, MMLU-Pro, WebInstruct & LLaMA-3.1, Qwen2.5, QwQ-32B & Accuracy, SFT+RL improvements, CoT type comparison \\

\bottomrule
\end{tabularx}
\caption{Experimental setup overview of CoT reasoning papers (Part~1).}
\label{tab:cot-summary-part1}
\end{table}
\begin{table}[htbp]
\centering
\small
\begin{tabularx}{\textwidth}{ >{\raggedright\arraybackslash}p{0.25\textwidth} >{\raggedright\arraybackslash\hsize=0.8\hsize\linewidth=\hsize}X >{\raggedright\arraybackslash\hsize=0.8\hsize\linewidth=\hsize}X >{\raggedright\arraybackslash\hsize=1.4\hsize\linewidth=\hsize}X }

\toprule
\textbf{Author} & \textbf{Datasets} & \textbf{Models} & \textbf{Evaluation Setup} \\
\midrule

\multicolumn{4}{l}{\textbf{Efficient \& Robust CoT}} \\

Cheng et al.~\citep{cheng2024compressed} & GSM8K & CCOT, PAUSE, Transformer & EM Accuracy, Decode Time, Compression vs Performance \\

Zhang et al.~\citep{zhang2025lightthinker} & GSM8K, MMLU, GPQA, BBH & Qwen2.5-7B, LLaMA3.1-8B, DeepSeek-R1-Distill & Accuracy, inference time, token compression \\

Xu et al.~\citep{xu2025chain} & GSM8K, BIG-bench, coin flip tasks & GPT-4o, Claude 3.5, Qwen2.5, LLaMA-3.2 & Accuracy, token count, CoD vs CoT tradeoffs \\

Han et al.~\citep{han2024token} & GSM8K, MathBench, GSM8K-Zero & GPT-4o, o3-mini, LLaMA3.1 & Accuracy vs token budget, redundancy tradeoff \\

He et al.~\citep{he2025can} & DeltaBench (Math, Code, General Reasoning) & QwQ, DeepSeek-R1, PRMs, critics & Error detection, critic effectiveness per step \\

Zhou et al.~\citep{zhou2024can} & NoRa & GPT-3.5, GPT-4, Mistral, LLaMA-2/3 & Accuracy drop under noisy CoT, robustness from CD-CoT \\

Zhang et al.~\citep{zhang2025soft} & Math500, AIME 2024, GSM8K, GPQA-Diamond (mathematics); HumanEval, MBPP, LiveCodeBench (programming) & QwQ-32B, DeepSeek-R1-Distill-Qwen-32B, DeepSeek-R1-Distill-Llama-70B & Soft Thinking compared with Standard CoT Thinking and Greedy CoT \\

\midrule
\multicolumn{4}{l}{\textbf{CoT Evaluation \& Adaptive Data Generation}} \\

Bai et al.~\citep{bai2024longbench} & LongBench v2 (6 task types) & GPT-4o, Claude 3.5, Qwen2.5, LLaMA 3.1, Mistral & Accuracy, CoT gain, solve time, human baseline \\

Zhou et al.~\citep{zhou2025landscape} & GSM8K, MathBench, GSM8K-Zero & GPT-4o, Yi-lightning, LLaMA3.1, o3-mini & Output token count, accuracy, interpretability \\

Chen et al.~\citep{chen2024unlocking} & BIGGSM, GSM8K, MATH, MGSM & GPT-3.5, GPT-4, LLaMA, Code-LLaMA & Accuracy, reasoning boundary categorization \\

Yu et al.~\citep{yu2025rethinking} & NuminaMath, AIME, OlympicArena, CodeForces & DS-distill, Zmath, Zcode, Sky-T1, phi4 & CoT quality, token efficiency, difficulty grading \\

\midrule
\multicolumn{4}{l}{\textbf{Limitations of CoT}} \\

Sprague et al.~\citep{sprague2024cot} & MATH, GSM8K, MuSiQue, StrategyQA, etc. & GPT-4, Claude 2, LLaMA 3.1, Mixtral, Zeus & MCQ accuracy, GPT-4o judged free-text equivalence \\

Liu et al.~\citep{liu2024mind} & FSG Grammar, MNLI, SNLI, Faces, Vehicles & GPT-4o, Claude 3.5, Gemini 1.5, LLaMA 3.1 & CoT vs Zero-shot performance drops, significance tests \\

Jin et al.~\citep{jin2024impact} & GSM8K, MultiArith, AQuA, SVAMP, etc. & GPT-4, GPT-3.5, text-davinci-002 & Step length ablation, compressed reasoning trends \\

Li et al.~\citep{li2025llms} & AMC, AIME, Math500, OlympiadBench & Qwen2.5-32B, DeepSeek traces & Accuracy, demo structure vs content impact \\

\bottomrule
\end{tabularx}
\caption{Experimental setup overview of CoT reasoning papers (Part~2).}
\label{tab:cot-summary-part2}
\end{table}

\subsection{Multi-Hop Reasoning}
\label{reason:II}
\begin{figure}[htbp]
\centering
\definecolor{ink}{RGB}{38,41,54}
\definecolor{inksoft}{RGB}{92,97,115}
\definecolor{hair}{RGB}{200,205,216}
\definecolor{cBlue}{RGB}{60,87,150}
\definecolor{cAmber}{RGB}{168,120,43}
\definecolor{cGreen}{RGB}{92,136,85}
\definecolor{cViolet}{RGB}{103,87,140}
\definecolor{cRose}{RGB}{169,79,103}
\definecolor{cSlate}{RGB}{85,92,112}
\scalebox{0.85}{
\begin{tikzpicture}[
 node distance=0.6cm,
 box/.style={rectangle, draw=#1, line width=0.9pt, rounded corners=5pt, fill=#1!8,
   minimum height=1.1cm, align=center, font=\small, text width=2cm, text=ink},
 chal/.style={rectangle, draw=cRose, line width=0.7pt, dashed, rounded corners=4pt, fill=cRose!7,
   align=center, font=\scriptsize, text width=1.8cm, text=ink, inner sep=4pt},
 arr/.style={-{Stealth[length=3mm]}, draw=inksoft, line width=1pt},
 link/.style={-{Stealth[length=1.8mm]}, draw=hair, line width=0.7pt, dashed},
]
\node[box=cSlate] (q) {\textcolor{cSlate}{\textbf{Question}}\\[0pt]{\scriptsize (multi-hop)}};
\node[box=cGreen, right=0.9cm of q] (h1) {\textcolor{cGreen}{\textbf{Hop 1}}\\[0pt]{\scriptsize Entity Retrieval}};
\node[circle, draw=cAmber, line width=0.9pt, fill=cAmber!10, minimum size=1.1cm, right=0.7cm of h1, font=\scriptsize, align=center, text=ink] (br) {Bridge\\[0pt]Entity};
\node[box=cRose, right=0.7cm of br] (h2) {\textcolor{cRose}{\textbf{Hop 2}}\\[0pt]{\scriptsize Knowledge Utilization}};
\node[box=cViolet, right=0.9cm of h2] (a) {\textcolor{cViolet}{\textbf{Answer}}};
\draw[arr] (q) -- (h1);
\draw[arr] (h1) -- (br);
\draw[-{Stealth[length=3mm]}, draw=cRose, line width=1.1pt] (br) -- node[above, font=\scriptsize\bfseries, text=cRose] {Gap} (h2);
\draw[arr] (h2) -- (a);
\node[chal, above=0.7cm of q]  (c1) {Evidence Distance};
\node[chal, above=0.7cm of h1] (c2) {Premise-Order Sensitivity};
\node[chal, above=0.7cm of br] (c3) {Data Contamination};
\node[chal, below=0.7cm of h2] (c4) {Lost in the Middle};
\draw[link] (c1) -- (q);
\draw[link] (c2) -- (h1);
\draw[link] (c3) -- (br);
\draw[link] (c4) -- (h2);
\end{tikzpicture}
}
\caption{Multi-hop reasoning failure modes. Hop~1 (entity retrieval) typically succeeds, but Hop~2 (knowledge utilization) often fails due to the utilization gap~\citep{yang2024large}. Structural challenges like evidence distance, premise order sensitivity, data contamination, and lost-in-the-middle effects, compound this fragility.}
\label{fig:multihop-challenges}
\end{figure}
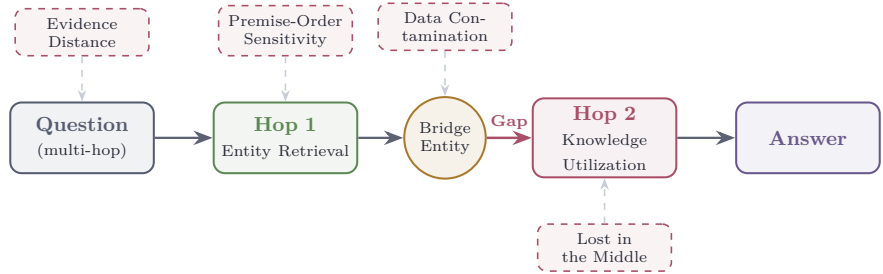

Multi-hop reasoning represents one of the most sophisticated cognitive capabilities in LLMs, requiring the synthesis of information across multiple sources, steps, or logical connections to reach valid conclusions. Unlike single-step inference, multi-hop reasoning demands that models maintain coherent chains of thought while navigating complex information landscapes. This complexity makes it an ideal stress test for probing the depth, robustness, and adaptability of reasoning in LLMs, and Figure~\ref{fig:multihop-challenges} visualizes the key failure modes that arise along the reasoning chain.
This section examines the diverse approaches researchers have developed to enhance and evaluate multi-hop reasoning in them. Recent research has approached multi-hop reasoning from multiple perspectives: some focus on strategic or meta-cognitive processes that guide reasoning steps, others emphasize how knowledge can be retrieved and integrated, while others propose new training paradigms, diagnostic analyses, or domain-specific benchmarks to better evaluate performance. In what follows, we categorize key works into five areas, each highlighting distinct challenges and innovations that shape the progress of multi-hop reasoning research. Table~\ref{tab:multihop-summary} summarizes the experimental setups.

\subsubsection{Strategic and Meta-Cognitive Reasoning}
One central line of work treats multi-hop reasoning as a process of strategic decision-making and self-regulation. These approaches frame reasoning not as a static sequence but as a dynamic process where models anticipate, self-correct, or adapt their intermediate steps.
 Zhang et al.~\citep{zhang2024k} introduces \textit{K-Level Reasoning}, a framework inspired by game theory that formalizes recursive anticipation of other agents’ beliefs. By embedding this kind of ''theory of mind'' capacity into LLMs, the model develops deeper strategic foresight, a crucial skill when multi-hop reasoning involves anticipating how intermediate knowledge will be used downstream. In a complementary direction, Huang et al.~\citep{huang2024queryagent} proposes \textit{QueryAgent}, which does not increase reasoning depth per se, but instead regulates it through selective self-correction. Their environmental feedback mechanism called ERASER, which allows the model to intervene only when necessary, preventing over-correction and wasted computation. This introduces an element of efficiency into the reasoning process rather than expanding reasoning indiscriminately, the model learns to know when to stop. Building on this theme of reasoning as self-managed dialogue,~\citep{radha2024iteration} presents the \textit{Iteration of Thought (IoT)} framework, which casts reasoning as a dynamic conversation between an Inner Dialogue Agent and the model itself. Unlike static prompting methods such as Chain-of-Thought or Tree-of-Thought, IoT adapts its trajectory with each iteration, guided by evolving context. The result is a reasoning process that is both autonomous and flexible, capable of refining responses in real time without discarding valuable intermediate explorations. These approaches illuminate a common shift in multi-hop reasoning research: from viewing reasoning as a rigid sequence of steps to treating it as a strategically managed process. They aim to introduce foresight and adaptive restraint to provide dynamic self-dialogue. Complex reasoning does not emerge solely from raw model capacity, but from mechanisms that enable models to strategically navigate, monitor, and refine their own cognitive pathways.

\subsubsection{Knowledge Integration and Retrieval-Augmented Reasoning}
Another challenge in multi-hop reasoning lies in how LLMs integrate external knowledge as retrieval must not only surface relevant information, but also structure it in ways that enable logical synthesis across steps. Recent approaches therefore move beyond naïve retrieve-and-read pipelines, embedding richer reasoning frameworks directly into the retrieval process.~\citep{kharbanda2024rcone} address this by introducing \textit{RConE}, which fuses symbolic logic with multi-modal knowledge graphs. Instead of treating retrieval as direct lookup, RConE encodes entities and logical operations in a shared space, enabling models to directly reason over conjunctions and negations of facts. This establishes a blueprint for retrieval that is inherently logical and compositional, rather than surface-level.~\citep{chu2024beamaggr} extend this structural orientation with \textit{BeamAggR}, which decomposes questions into reasoning trees and explores multiple knowledge paths in parallel through beam search. By aggregating candidate answers bottom-up, BeamAggR emphasizes diversity of reasoning routes and robustness to partial errors, capturing the intuition that multi-hop inference often requires triangulating across imperfect sources.~\citep{gan2024similarity} take the integration problem further with \textit{MetRAG}, which critiques the over-reliance on similarity-based retrieval in RAG. Their framework introduces ``multi-layered thoughts'': similarity-oriented retrieval is complemented with utility judgments (how useful a passage actually is for answering) and compactness via task-adaptive summarization. In doing so, MetRAG reframes retrieval not as a single-step filter, but as a layered reasoning process that pre-structures knowledge before generation. The recognition that multi-hop reasoning requires not just more documents, but smarter integration and retrieval that anticipates the reasoning demands of the task, rather than passively supplying context. 

\subsubsection{Learning Paradigms for Multi-Step Reasoning}
Beyond frameworks and retrieval, recent work focuses on training LLMs for stronger multi-hop performance. These studies experiment with reinforcement learning, data augmentation, and human feedback as ways to refine multi-step reasoning abilities. 
The nature of the learning signal itself is a key differentiator in these paradigms. Some methods rely on a signal that is external, explicit, and diagnostic. The work by~\citep{joshi2024improving} exemplifies this by using rich human feedback as the primary learning source. Here, the signal is not just corrective but also educational, as human annotators provide natural language explanations and categorize errors. This highly explicit signal proves effective for improving not only accuracy but also the model's own ability to judge the correctness of its outputs. In contrast, the OREO framework in ~\citep{wang2024offline} develops a signal that is internalized and predictive. It learns from an offline dataset of trajectories with sparse terminal rewards. The core innovation is training a value function that internalizes the task structure, learning to predict the likelihood of future success from any intermediate step. This learned value function becomes an internal critic, generating a dense, step-by-step training signal from an initially sparse external one. The quality of this signal is demonstrated not only by improved policy performance but also by its utility in guiding test-time beam search to find better solutions. ~\citep{abramov2025grokking} in sources its learning signal from the structural and implicit properties of the data distribution itself. The ``signal'' here is the statistical density of the knowledge graph; by synthetically augmenting the data to raise the ratio of inferred to atomic facts, they create a learning environment where generalization becomes a more efficient strategy than memorization. Collectively, these paradigms illustrate a clear trajectory in the field to move from direct, labor-intensive supervision of reasoning steps toward more scalable and autonomous methods and the choice ultimately depends on the desired trade-offs between sample efficiency, scalability, and the explicitness of the reasoning process being taught, with hybrid approaches representing a promising avenue for future research.

\subsubsection{Evaluating and Benchmarking Multi-Hop Capabilities}
Multi-hop reasoning is difficult to quantify, the field has invested in designing robust benchmarks. These range from augmented datasets to open-source agentic frameworks that stress-test reasoning across tools and domains.
A foundational challenge in this area is data contamination. Early evaluations on benchmarks like HotpotQA were potentially confounded by the test data's presence in the LLMs' vast pre-training corpora. To address this, ~\citep{wu2024mrke} developed MRKE, a benchmark created by systematically editing knowledge in the HotpotQA dataset~\citep{yang2018hotpotqa} to ensure it is novel. This work revealed a significant performance drop when models could no longer rely on memorization, establishing the need for ``clean'' evaluation sets. Furthermore, MRKE was among the first to formalize the evaluation of the intermediate reasoning chain, not just the final answer, showing that models often arrive at correct conclusions through flawed logic. Building on the need to test reasoning on unseen information, subsequent benchmarks have focused on naturally occurring knowledge gaps such as the MINTQA benchmark~\citep{he2024mintqa} that specifically evaluates reasoning over new and long-tail (out-of-distribution) knowledge, pushing models beyond the well-represented facts of their training data. Similarly,~\citep{khodadad2025evaluating} created a benchmark for the highly specialized domain of chemistry, using an automated pipeline to construct a knowledge graph from recent scientific literature. Their work highlights that even with perfect retrieval, state-of-the-art models still struggle with the compositional reasoning required to connect facts.
Most recently, the focus of evaluation has expanded from static question-answering to assessing dynamic, tool-using agents. The HUSKY framework~\citep{kim2024husky} provides a unified, open-source agent designed to solve problems requiring a mix of tools like search, code execution, and mathematical calculation. They also introduced the HuskyQA dataset, which explicitly requires this mixed-tool use (e.g., finding a value via search and then using it in a calculation). This represents a shift towards evaluating the entire agentic process, providing a more holistic and practical measure of a model's multi-hop reasoning capabilities in real-world scenarios. The consistent finding across these benchmarks is that while LLMs show promise, their reasoning chains are often brittle, they struggle to generalize beyond common knowledge, and the ability to reliably compose information remains a significant bottleneck, even when all necessary facts are readily available.

\subsubsection{Structural and Contextual Challenges}
Finally, a growing body of work highlights the fragility of multi-hop reasoning when faced with contextual or structural challenges. LLMs often struggle with issues such as long-context attention failures, sensitivity to premise ordering, or just the implicit nature of multi-hop inference. Diagnostic studies in this category uncover these weaknesses and propose targeted solutions, helping to identify blind spots that must be addressed for more reliable reasoning systems. 
One of the most significant contextual challenges manifests at the macro level of long documents. .~\citep{baker2024lost} shows that the problem is particularly damaging for multi-hop reasoning, where multiple pieces of evidence must be synthesized. Performance is degraded not only by the absolute position of evidence but also by the relative distance between those pieces of evidence. As crucial facts are spaced further apart by distractor documents, the model’s ability to form a coherent reasoning chain diminishes, highlighting a critical failure in utilizing large context spaces effectively. This sensitivity to information placement persists even at the micro level of short, self-contained problems as the issue shifts from location to sequence. ~\citep{chen2024premise} finds that LLMs are highly brittle to premise order. Performance is optimal when facts and rules are presented in an order that aligns with the steps of a proof, but shuffling them causes a severe drop in accuracy. This suggests a strong bias towards linear reasoning and a fundamental weakness in synthesizing information from a disorganized set of premises.
Probing deeper, some work questions whether this synthesis happens reliably at the most fundamental, latent level. ~\citep{yang2024large} investigates the internal reasoning process itself and find a crucial asymmetry. While models are adept at the first hop of reasoning (identifying an intermediate entity from a description), they are far less successful at the second hop (using knowledge about that entity to find the final answer). This ``utilization gap'' (depicted in Figure~\ref{fig:multihop-challenges}), which notably does not improve with model scale, suggests that the connection between reasoning steps is a core point of failure, independent of context length or premise order.
These studies reveal that multi-hop reasoning in LLMs is a process severely affected by structural dependencies. LLMs consistently struggle when information is not presented in a simple, linear fashion. This suggests that true, robust reasoning will require models to move beyond their inherent sequential processing biases and develop more flexible mechanisms for accessing and connecting information across their entire context.

\begin{table}[htbp]
\centering
\small
\begin{tabularx}{\textwidth}{ >{\raggedright\arraybackslash}p{0.25\textwidth} >{\raggedright\arraybackslash\hsize=0.8\hsize\linewidth=\hsize}X >{\raggedright\arraybackslash\hsize=0.8\hsize\linewidth=\hsize}X >{\raggedright\arraybackslash\hsize=1.4\hsize\linewidth=\hsize}X }

\toprule
\textbf{Author} & \textbf{Datasets} & \textbf{Models} & \textbf{Evaluation Setup} \\
\midrule

\multicolumn{4}{l}{\textbf{Strategic and Meta-Cognitive Reasoning}} \\

Zhang et al.~\citep{zhang2024k} & G0.8A, SAG, NEG, SOTOPIA & GPT-4, GPT-3.5, LLaMA2-7B & Win rate, survival rounds, utility maximization, GPT-4 eval on BEL, REL, KNO \\

Huang et al.~\citep{huang2024queryagent} & GrailQA, GraphQ, WebQSP, MetaQA & GPT-3.5, GPT-4 & One-shot QA, F1 score, ERASER ablation, feedback-based correction \\

Radha et al.~\citep{radha2024iteration} & GPQA Diamond & GPT-4o Mini & Accuracy, AIoT vs GIoT vs CoT, dialog-based reasoning \\

\midrule
\multicolumn{4}{l}{\textbf{Knowledge Integration and Retrieval-Augmented Reasoning}} \\

Kharbanda et al.~\citep{kharbanda2024rcone} & FB15k, YAGO15k, DB15k & RConE (w/wo Transformer) & MRR, HITS@K, 14 query types, scene-to-KG linking \\

Chu et al.~\citep{chu2024beamaggr} & HotpotQA, MuSiQue, 2WikiMQA, Bamboogle & GPT-3.5, Mistral-7B & F1 score, BeamAggR vs CoT, IRCoT, open-domain retrieval \\

Gan et al.~\citep{gan2024similarity} & NQ, TriviaQA, HotpotQA, PopQA & ChatGLM2, LLaMA2, Baichuan, Qwen & EM, F1 score, RAG vs RECOMP, METRAG, SELF-RAG \\

\midrule
\multicolumn{4}{l}{\textbf{Learning Paradigms for Multi-Step Reasoning}} \\

Joshi et al.~\citep{joshi2024improving} & StrategyQA, Sports QA & LLaMA2, GPT-J, Flan-T5 & Accuracy, multitask error detection and correction \\

Wang et al.~\citep{wang2024offline} & GSM8K, MATH, ALFWorld & Qwen2.5-Math, DeepSeekMath, MiniCPM & Accuracy, ALFWorld success rate, OREO loss \\

Abramov et al.~\citep{abramov2025grokking} & 2WikiMultiHopQA (augmented) & GPT-2, GPT-4o, o1-mini & ID/OOD accuracy, phi-triggered grokking, structured gain \\

\midrule
\multicolumn{4}{l}{\textbf{Evaluating and Benchmarking Multi-Hop Capabilities}} \\

Wu et al.~\citep{wu2024mrke} & MRKE Benchmark, HotpotQA & GPT-4, GPT-3.5, Gemini & EM, F1, joint chain score, hop-wise sub-question evaluation \\

He et al.~\citep{he2024mintqa} & MINTQA-POP, MINTQA-TI & GPT-4o, LLaMA3, Qwen2.5 & Accuracy, 1–4 hop QA, tool retrieval effectiveness \\

Khodadad et al.~\citep{khodadad2025evaluating} & Chem-MH QA, Hotpot-Chem & GPT-4o, Claude, LLaMA3, DeepSeek, Qwen & Correctness, latency, token usage, context comparison \\

Kim et al.~\citep{kim2024husky} & GSM8K, TAT-QA, HUSKYQA, HotpotQA & HUSKY (LLMs + Tools) & Zero-shot tool-aligned modules, reasoning accuracy \\

\midrule
\multicolumn{4}{l}{\textbf{Structural and Contextual Challenges}} \\

Baker et al.~\citep{baker2024lost} & HotpotQA, 2WikiMHQA, MuSiQue & GPT-3.5, LLaMA2-LongLoRA, MPT & Best-subspan accuracy, Lost-in-the-Middle analysis \\

Chen et al.~\citep{chen2024premise} & Logical Tasks, R-GSM & GPT-4, Gemini, PaLM-2 & Accuracy, hallucination types, premise ordering effects \\

Yang et al.~\citep{yang2024large} & TWOHOPFACT & LLaMA2 (7B, 13B, 70B) & ENTREC (bridge recall), CNSTSCORE (2-hop consistency) \\

\bottomrule
\end{tabularx}
\caption{Experimental setup overview of multi-hop reasoning papers.}
\label{tab:multihop-summary}
\end{table}

\subsection{Mathematical Reasoning}
\label{reason:III}
Mathematical reasoning represents one of the most stringent tests of genuine intelligence in large language models, requiring precise symbolic manipulation, multi-step logical inference, and abstract thinking that goes far beyond pattern matching. Unlike other language tasks where approximate solutions may suffice, mathematics demands absolute correctness where even minor errors can invalidate entire solution paths. Figure~\ref{fig:math-reasoning-flow} illustrates the end-to-end pipeline from problem input through symbol extraction, formula application, and arithmetic execution to a final verified answer. The pursuit of mathematical reasoning in LLMs has catalyzed innovations across multiple research dimensions, from specialized model architectures and training paradigms to sophisticated inference techniques and evaluation frameworks. These advances have not only improved mathematical performance but have also contributed foundational insights into reasoning capabilities more broadly. The following subsections examine how researchers have approached this challenge through distinct yet interconnected strategies, each addressing different aspects of the mathematical reasoning problem while collectively advancing our understanding of how to build truly reasoning-capable language models. Tables~\ref{tab:mathreasoning-summary-part1} and~\ref{tab:mathreasoning-summary-part2} provide the corresponding experimental details.

\subsubsection{Architectural Innovation and Model Development}
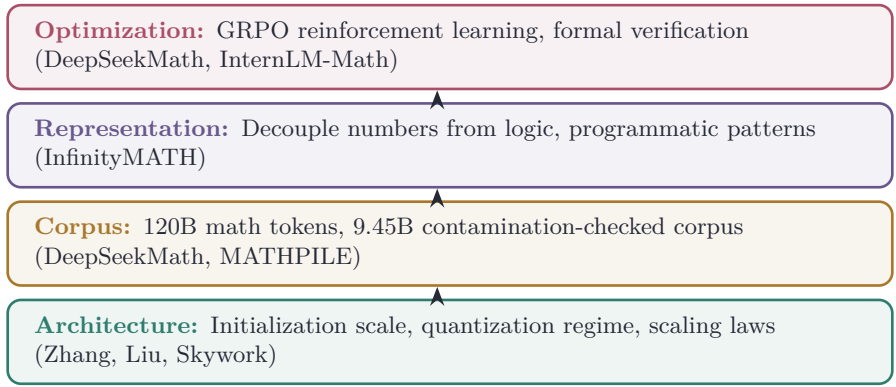
\begin{figure}[htbp]
\centering
\definecolor{ink}{RGB}{38,41,54}
\definecolor{cTeal}{RGB}{46,125,111}
\definecolor{cAmber}{RGB}{168,120,43}
\definecolor{cViolet}{RGB}{103,87,140}
\definecolor{cRose}{RGB}{169,79,103}
\begin{tikzpicture}[
  layer/.style={rectangle, draw=#1, line width=1pt, rounded corners=5pt, fill=#1!8,
    text width=11cm, minimum height=1.0cm, align=left, font=\small, text=ink,
    inner xsep=10pt, inner ysep=6pt, anchor=west},
  arr/.style={-{Stealth[length=2.5mm]}, draw=ink, line width=1.1pt},
]
\node[layer=cRose]   (l4) at (0,3.9) {\textcolor{cRose}{\textbf{Optimization:}} GRPO reinforcement learning, formal verification (DeepSeekMath, InternLM-Math)};
\node[layer=cViolet] (l3) at (0,2.6) {\textcolor{cViolet}{\textbf{Representation:}} Decouple numbers from logic, programmatic patterns (InfinityMATH)};
\node[layer=cAmber]  (l2) at (0,1.3) {\textcolor{cAmber}{\textbf{Corpus:}} 120B math tokens, 9.45B contamination-checked corpus (DeepSeekMath, MATHPILE)};
\node[layer=cTeal]   (l1) at (0,0.0) {\textcolor{cTeal}{\textbf{Architecture:}} Initialization scale, quantization regime, scaling laws (Zhang, Liu, Skywork)};
\draw[arr] (l1.north -| 5.7,0) -- (l2.south -| 5.7,0);
\draw[arr] (l2.north -| 5.7,0) -- (l3.south -| 5.7,0);
\draw[arr] (l3.north -| 5.7,0) -- (l4.south -| 5.7,0);
\end{tikzpicture}
\caption{Four layers of architectural innovation for mathematical reasoning. From bottom to top: low-level architecture choices (initialization, quantization, scaling laws), curated pretraining corpora, representation design, and domain-specific optimization. Each layer builds on the ones below it.}
\label{fig:math-arch-stack}
\end{figure}

Recent work re-engineers multiple layers of the LLM stack for mathematics, from pretraining corpora and optimization algorithms through data representation to low-level hyperparameters like initialization and quantization. These innovations cluster into three themes (Figure~\ref{fig:math-arch-stack}), and Table~\ref{tab:math-arch-innovations} summarizes each contribution.

\paragraph{Specialized pretraining and optimization.}
DeepSeekMath~\citep{shao2024deepseekmath} continues pretraining on 120B math-focused tokens and introduces Group Relative Policy Optimization (GRPO), an RL algorithm tailored to efficient reasoning. The resulting 7B model rivals closed-source systems without tools or ensembles. InternLM-Math~\citep{ying2024internlm} integrates chain-of-thought reasoning with formal verification tools (e.g., LEAN), positioning the model as a solver and prover-verifier hybrid that bridges informal reasoning with formal proof.

\paragraph{Scaling laws, representation, and corpus quality.}
Skywork-Math~\citep{zeng2024skywork} validates that mathematical ability follows systematic scaling behaviors: more structured, diverse pretraining data produces steady benchmark gains, providing practical scaling principles for future models. InfinityMATH~\citep{zhang2024infinitymath} innovates at the representation level by decoupling numbers from problem logic so models learn programmatic reasoning patterns independent of specific values, yielding large gains and demonstrating that program-level abstraction unlocks transferable reasoning. MATHPILE~\citep{wang2024mathpile} sets a new standard for corpus quality with its 9.45B-token, contamination-checked math corpus, showing that meticulous filtering and domain-specific curation produce a cleaner pretraining substrate than generic corpora.

\paragraph{Hidden architectural levers.}
~\citep{zhang2024initialization} reveal that initialization scale determines whether transformers reason or memorize when solving compositional functions: small initializations encourage inferential behavior, while large ones bias toward rote memorization.~\citep{liu2025quantization} conduct the first systematic study of quantization effects on reasoning, showing that W8A8 or W4A16 quantization can degrade or preserve reasoning depending on model size, architecture, and task difficulty.

\begin{table}[htbp]
\centering
\small
\begin{tabularx}{\textwidth}{ >{\raggedright\arraybackslash}p{0.19\textwidth} >{\raggedright\arraybackslash}p{0.16\textwidth} >{\raggedright\arraybackslash}X >{\raggedright\arraybackslash}p{0.22\textwidth} }
\toprule
\textbf{System} & \textbf{Innovation layer} & \textbf{Key contribution} & \textbf{Main finding} \\
\midrule
DeepSeekMath \citep{shao2024deepseekmath} & Pretraining + RL & 120B math tokens; GRPO algorithm & 7B model rivals closed-source systems \\
\addlinespace
InternLM-Math \citep{ying2024internlm} & Framework & CoT + formal verification (LEAN prover) & Solver and prover-verifier in one model \\
\addlinespace
Skywork-Math \citep{zeng2024skywork} & Scaling laws & Validates math-specific scaling behaviors & Structured data yields steady benchmark gains \\
\addlinespace
InfinityMATH \citep{zhang2024infinitymath} & Representation & Decouples numbers from problem logic & Program-level abstraction transfers across tasks \\
\addlinespace
MATHPILE \citep{wang2024mathpile} & Corpus curation & 9.45B-token contamination-checked corpus & Quality outweighs scale for pretraining \\
\addlinespace
Zhang et al. \citep{zhang2024initialization} & Initialization & Small vs. large init scale on compositional tasks & Small init $\to$ reasoning; large init $\to$ memorization \\
\addlinespace
Liu et al. \citep{liu2025quantization} & Quantization & W8A8 and W4A16 effects on reasoning & Impact varies by model size, arch, and task \\
\bottomrule
\end{tabularx}
\caption{Architectural innovations for mathematical reasoning, grouped by the layer of the model stack each targets.}
\label{tab:math-arch-innovations}
\end{table}

\subsubsection{Data Scaling and Synthetic Data Generation}
Scaling both the quantity and the quality of training data has proven to be one of the most reliable levers for advancing mathematical reasoning in LLMs. Yet, the frontier of progress lies not merely in generating more data but in curating it in ways that target reasoning weaknesses and amplify generalization. Across recent work, we see a convergence toward strategies that blend large-scale synthesis with difficulty awareness and structural augmentation, reshaping how models acquire problem-solving depth.
A recurring discovery is that even modestly sized models harbor latent mathematical abilities that can be surfaced with the right data scaling strategy. ~\citep{li2024common} show that common 7B models, without any specialized pretraining, already possess surprisingly strong math capabilities when exposed to large volumes of synthetic supervision. By expanding supervised fine-tuning data and leveraging synthetic augmentation, their results reveal how data scale alone can unlock dormant reasoning potential. Yet this also underscores the fragility of such gains, as raw scale may inflate accuracy but does not guarantee robustness across varied problem structures.
Building on this observation, ~\citep{tang2024mathscale} argue that the structure of scaling is as important as its size. Their \texttt{MathScale} framework introduces a concept-graph-driven method of dataset generation that gradually expands problem difficulty, resembling a curriculum tailored for LLMs. This approach ensures that as models scale, they are not just memorizing more examples but are progressively developing reasoning depth. Complementing this structured expansion, ~\citep{toshniwal2024openmathinstruct} also demonstrates in \texttt{OpenMathInstruct-1} that high-quality synthetic datasets can be produced without relying on closed-source models. By generating 1.8 million problem-solution pairs entirely with open-source tools, they provide a pathway for reproducible, transparent scaling that narrows the performance gap with GPT-distilled data.
Even with size and openness addressed, another critical gap remains of imbalance between easy and hard problems. ~\citep{tong2024dart} confront this with DART-Math, a difficulty-aware rejection tuning strategy that directs synthesis efforts toward harder questions, thereby producing compact yet impactful datasets. Their findings emphasize that quality lies not in volume alone but in targeting the reasoning bottlenecks where models struggle most. Taken together, these directions charts a trajectory for data scaling research that moves beyond brute force toward deliberate, fine-grained design. This evolution highlights that the future of mathematical reasoning progress will depend less on simply ``more data'' and more on ``better, smarter data."

\subsubsection{Process Supervision and Step-by-Step Reasoning}
Mathematical reasoning requires precise intermediate steps, making process-level supervision an increasingly important paradigm. Rather than judging only final answers, recent methods generate, align, and correct reasoning traces through progressively more automated approaches. Table~\ref{tab:process-supervision-spectrum} compares three representative systems along this automation spectrum.

~\citep{luo2024improve} introduce OmegaPRM, which uses Monte Carlo Tree Search to pinpoint errors within CoT traces automatically. It annotates reasoning steps without human input, yielding large accuracy gains while keeping annotation costs low. ~\citep{zhang2025booststep} take a different angle with BoostStep, which retrieves exemplars that match the current reasoning stage through step-aligned in-context learning. This fine-grained alignment reduces common slip-ups in multi-step solutions and benefits even powerful models like GPT-4o. ~\citep{chen2024alphamath} go further with AlphaMath Almost Zero, where process supervision emerges without any external step-level labels. By combining MCTS with a value model to explore and prioritize reasoning trajectories, it achieves state-of-the-art results with no reliance on human or GPT-generated annotations.

\begin{table}[htbp]
\centering
\small
\begin{tabularx}{\textwidth}{ >{\raggedright\arraybackslash}p{0.17\textwidth} >{\raggedright\arraybackslash}p{0.17\textwidth} >{\raggedright\arraybackslash}X >{\raggedright\arraybackslash}p{0.18\textwidth} >{\raggedright\arraybackslash}p{0.15\textwidth} }
\toprule
\textbf{System} & \textbf{Supervision source} & \textbf{Mechanism} & \textbf{External labels needed} & \textbf{Automation level} \\
\midrule
OmegaPRM \citep{luo2024improve} & Automated MCTS annotations & Searches CoT traces to localize errors at each step & None (replaces human annotators) & Medium \\
\addlinespace
BoostStep \citep{zhang2025booststep} & Step-aligned retrieval & Retrieves exemplars matching the current reasoning stage via in-context learning & Existing solved examples & Medium-High \\
\addlinespace
AlphaMath \citep{chen2024alphamath} & Self-supervised MCTS + value model & Explores and prioritizes trajectories; model generates and critiques its own paths & None at all & Full \\
\bottomrule
\end{tabularx}
\caption{Process supervision approaches for mathematical reasoning, ordered by increasing automation. All three shift focus from outcome checking to guiding the reasoning process itself.}
\label{tab:process-supervision-spectrum}
\end{table}

\subsubsection{Search, Self-Refinement, and Inference-Time Optimization}
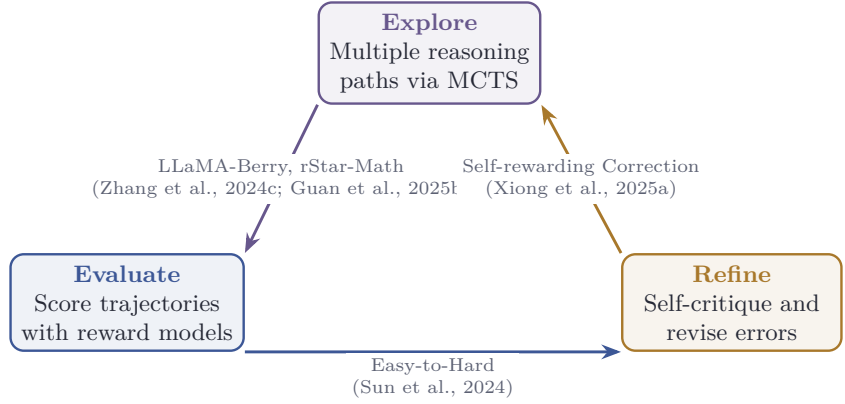
\begin{figure}[htbp]
\centering
\definecolor{ink}{RGB}{38,41,54}
\definecolor{inksoft}{RGB}{92,97,115}
\definecolor{cBlue}{RGB}{60,87,150}
\definecolor{cAmber}{RGB}{168,120,43}
\definecolor{cGreen}{RGB}{92,136,85}
\definecolor{cViolet}{RGB}{103,87,140}
\begin{tikzpicture}[
  phase/.style={rectangle, draw=#1, line width=0.9pt, rounded corners=5pt, fill=#1!8,
    minimum height=1.25cm, minimum width=2.9cm, align=center, font=\small, text=ink, inner sep=4pt},
  arr/.style={-{Stealth[length=2.6mm]}, draw=#1, line width=1.1pt},
  elab/.style={font=\scriptsize, text=inksoft, align=center, fill=white, inner sep=2pt},
]
\node[phase=cViolet] (explore)  at (0,2.8)    {\textcolor{cViolet}{\textbf{Explore}}\\Multiple reasoning\\paths via MCTS};
\node[phase=cBlue]   (evaluate) at (-4,-0.5)  {\textcolor{cBlue}{\textbf{Evaluate}}\\Score trajectories\\with reward models};
\node[phase=cAmber]  (refine)   at (4,-0.5)   {\textcolor{cAmber}{\textbf{Refine}}\\Self-critique and\\revise errors};
\draw[arr=cViolet] (explore.south west) -- node[elab] {LLaMA-Berry, rStar-Math\\{\scriptsize\citep{zhang2024llama,guan2025rstar}}} (evaluate.north east);
\draw[arr=cBlue]   (evaluate.south east) -- node[elab, below] {Easy-to-Hard\\{\scriptsize\citep{sun2024easy}}} (refine.south west);
\draw[arr=cAmber]  (refine.north west) -- node[elab] {Self-rewarding Correction\\{\scriptsize\citep{xiong2025self}}} (explore.south east);
\end{tikzpicture}
\caption{The inference-time feedback loop for mathematical reasoning. Exploration generates candidate paths (MCTS), evaluation scores them (reward models that transfer from easy to hard), and refinement revises errors (self-rewarding correction). The cycle repeats until a satisfactory solution emerges.}
\label{fig:inference-time-loop}
\end{figure}

Solving complex math problems requires active exploration, evaluation, and revision at inference time. Recent work converges on a closed feedback loop (Figure~\ref{fig:inference-time-loop}) where these three phases reinforce each other.

LLaMA-Berry~\citep{zhang2024llama} and rStar-Math~\citep{guan2025rstar} both leverage Monte Carlo Tree Search to probe multiple reasoning paths. LLaMA-Berry pairs exploration with iterative self-critique and preference-based ranking to prioritize promising trajectories. rStar-Math demonstrates that sustained cycles of exploration and reward-guided refinement allow even smaller models to rival much larger systems. In both cases, scale is redefined as the breadth of reasoning space explored at inference.

~\citep{xiong2025self} embed self-correction as an intrinsic capability: the model generates its own feedback signals through rejection sampling and lightweight RL, becoming both solver and critic. ~\citep{sun2024easy} show that reward models trained on simple tasks transfer to harder ones, reducing dependence on human supervision for complex domains.

\subsubsection{Evaluation and Robustness Assessment}
Robust evaluation now probes resilience to perturbations, generalization across problem variants, and reliability on out-of-distribution inputs. Recent frameworks operate at three distinct levels:

\begin{itemize}
\item \textbf{Perturbation benchmarks.} MathCheck~\citep{zhou2024your} perturbs existing problems and tests how models adapt to equivalent but slightly altered forms. Many state-of-the-art systems collapse under minor rephrasings, exposing brittle correlations. GSM-Symbolic~\citep{mirzadeh2024gsm} reinforces this by constructing variant-controlled problem sets through symbolic templates, where simple alterations like clause insertion or value substitution trigger accuracy drops exceeding 60\%.

\item \textbf{Frontier-difficulty evaluation.} OlymMATH~\citep{sun2025challenging} curates Olympiad-level problems across multiple domains. Where perturbation frameworks reveal brittleness in everyday problem-solving, OlymMATH maps the upper limits of what even the strongest models can achieve, showing that high-stakes mathematical reasoning remains far from solved.

\item \textbf{Intrinsic distributional analysis.} ~\citep{wang2024embedding} introduce the TV score, which analyzes the volatility of embedding trajectories in latent space to detect out-of-distribution inputs. This reframes evaluation as a diagnostic tool that signals when a model is venturing outside its reliable operating range.
\end{itemize}

\begin{figure}[htbp]
\centering
\definecolor{ink}{RGB}{38,41,54}
\definecolor{inksoft}{RGB}{92,97,115}
\definecolor{cTeal}{RGB}{46,125,111}
\definecolor{cAmber}{RGB}{168,120,43}
\definecolor{cViolet}{RGB}{103,87,140}
\begin{tikzpicture}[>=stealth,
  pillar/.style={rectangle, rounded corners=5pt, line width=0.9pt,
    minimum height=1.6cm, minimum width=3.6cm, align=center, font=\small, text=ink, inner sep=5pt},
  elab/.style={font=\scriptsize, text=inksoft, align=center, fill=white, inner sep=2pt},
]
\node[pillar, draw=cTeal, fill=cTeal!8] (proc) at (0,0)
  {\textcolor{cTeal}{\textbf{Process Supervision}}\\[0pt]{\scriptsize OmegaPRM, BoostStep, AlphaMath}};
\node[pillar, draw=cViolet, fill=cViolet!8] (inf) at (6.0,0)
  {\textcolor{cViolet}{\textbf{Inference-Time Optimization}}\\[0pt]{\scriptsize LLaMA-Berry, rStar-Math, Self-rewarding Correction}};
\node[pillar, draw=cAmber, fill=cAmber!8] (eval) at (3.0,-3.2)
  {\textcolor{cAmber}{\textbf{Evaluation \& Robustness}}\\[0pt]{\scriptsize MathCheck, GSM-Symbolic, OlymMATH, TV Score}};
\draw[->, draw=cTeal, line width=1.1pt] (proc.east) -- node[elab, above] {step rewards} (inf.west);
\draw[->, draw=cViolet, line width=1.1pt] (inf.south) -- node[elab] {exposes gaps} (eval.north east);
\draw[->, draw=cAmber, line width=1.1pt] (eval.north west) -- node[elab] {targets weak steps} (proc.south);
\end{tikzpicture}
\caption{Three reinforcing pillars of mathematical reasoning improvement. Process supervision provides step-level rewards that enable inference-time search and refinement. Evaluation exposes where models fail, directing supervision toward weak steps. The cycle tightens as each pillar informs the others.}
\label{fig:math-three-pillars}
\end{figure}
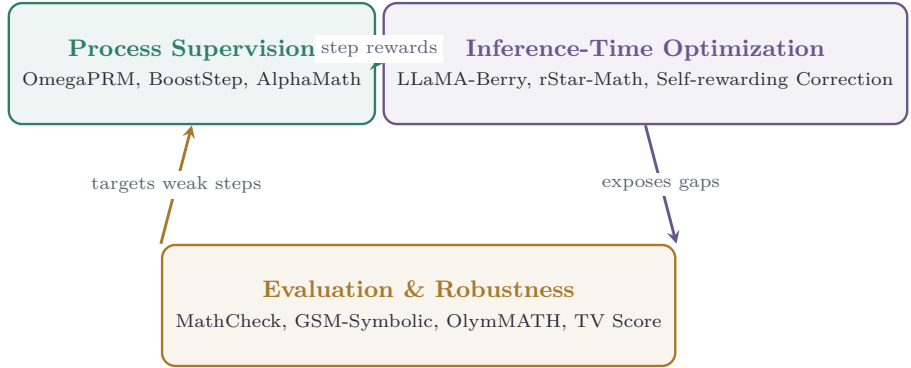

\noindent Figure~\ref{fig:math-three-pillars} shows how process supervision, inference-time optimization, and evaluation interact as three reinforcing phases of mathematical reasoning improvement.

\subsubsection{Reinforcement Learning and Reward-Based Training}
Reinforcement learning offers a natural paradigm for mathematical reasoning, where correctness is often binary and intermediate supervision scarce. For a fuller treatment of RL for reasoning, including search policies, stepwise preference optimization, process rewards, and pure-RL emergence, see Section~\ref{reason:X}. The challenge lies in making sparse outcome signals usable without sacrificing efficiency. Three recent directions address this:

\begin{itemize}
\item \textbf{Sharpening sparse outcome rewards.} ~\citep{lyu2025exploring} prove that behavior cloning from best-of-N samples converges to optimal policies under KL regularization. Their OREAL framework extends this with token-level reward estimation, mitigating sparsity and driving a 7B model to near-perfect performance on MATH-500, rivaling systems many times larger. Outcome rewards, when refined through the right optimization lens, can become precise learning signals.

\item \textbf{Domain-specific structure for reward efficiency.} ~\citep{li2025sos1} demonstrate this in the context of sum-of-squares optimization, a mathematically demanding task linked to Hilbert's 17th problem. By constructing SoS-1K with progressive difficulty levels and embedding reasoning instructions into prompts, a smaller model masters a domain previously inaccessible to LLMs and outperforms GPT-4o-mini in both accuracy and efficiency. Coupling outcome-driven training with structured datasets unlocks capabilities disproportionate to model size.

\item \textbf{Optimizing reasoning efficiency, not just correctness.} ~\citep{chen2024not} highlight that powerful o1-like models tend to `overthink,'' expending unnecessary computation on trivial problems. By introducing outcome and process-based efficiency metrics and proposing self-training strategies to regulate reasoning depth, they show that reward shaping can reduce wasted effort without sacrificing performance, addressing an overlooked cost of large-scale reasoning systems.
\end{itemize}

\noindent The unifying thread is the strategic use of rewards as levers for shaping reasoning quality, specialization, and computational economy. The future of RL in mathematical reasoning lies in smarter reward design that balances accuracy, generalization, and efficiency.

\subsubsection{Metacognitive and Self-Reflective Reasoning}
An emerging frontier in mathematical reasoning involves teaching models to reflect on their own thought processes. By eliciting metacognitive awareness, such as labeling skills, evaluating solution strategies, and reflecting beyond final answers, LLMs become not just solvers but self-aware learners, capable of adapting reasoning strategies to new challenges.~\citep{didolkar2024metacognitive} demonstrate this by prompting models to assign skill labels to math problems, clustering them into interpretable categories that capture the underlying cognitive demands. By linking new problems to previously solved skill clusters, models gain a scaffold for selecting relevant strategies, effectively mimicking how humans recall and reuse domain knowledge. This metacognitive structuring leads to measurable gains on benchmarks like GSM8K~\citep{cobbe2021training} and MATH~\citep{hendrycks2021measuring}, suggesting that reflective categorization helps bridge the gap between isolated problem-solving and strategic reasoning.
~\citep{zhang2024learn} extend this notion of reflection from categorization to deliberate abstraction and critique. Their Reflective Augmentation framework trains models not just to deliver answers, but to articulate and reconsider solution approaches, weigh alternatives, and distill reusable reasoning patterns. This process shifts the emphasis from correctness alone to learning how to reason better over time, which proves especially valuable for complex, multi-hop tasks. Unlike conventional data augmentation, reflective training creates a feedback loop where models practice meta-level thinking, enhancing robustness and adaptability. By teaching models to organize their knowledge into skills, reflect on their strategies, and generalize beyond immediate answers, researchers are embedding a form of ``learning-to-learn'' within LLMs. As such, metacognitive reasoning does not replace process supervision or reinforcement learning but complements them by equipping models with the self-awareness needed to choose, adapt, and refine reasoning strategies on their own.

\begin{figure}[htbp]
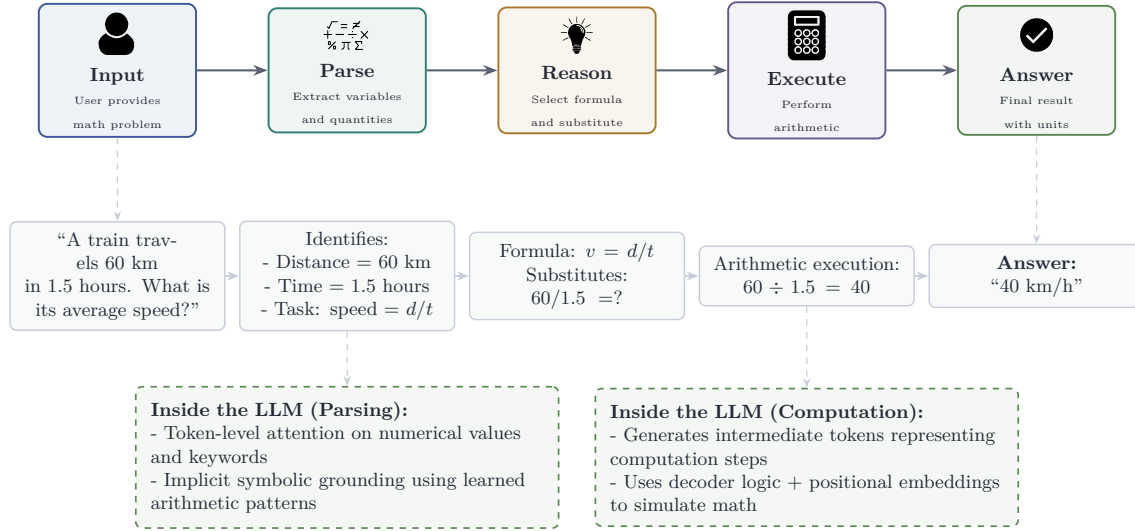

\centering
\definecolor{ink}{RGB}{38,41,54}
\definecolor{inksoft}{RGB}{92,97,115}
\definecolor{hair}{RGB}{200,205,216}
\definecolor{cBlue}{RGB}{60,87,150}
\definecolor{cTeal}{RGB}{46,125,111}
\definecolor{cAmber}{RGB}{168,120,43}
\definecolor{cViolet}{RGB}{103,87,140}
\definecolor{cGreen}{RGB}{92,136,85}
\scalebox{0.8}{
\begin{tikzpicture}[
 pstep/.style={rectangle, draw, line width=0.9pt, rounded corners=4pt, minimum height=1.5cm, minimum width=2.6cm, align=center, font=\small, text=ink},
 ex/.style={rectangle, draw=hair, line width=0.7pt, rounded corners=3pt, fill=ink!3, minimum height=0.7cm, align=center, font=\small, text width=3.2cm, inner sep=5pt, text=ink},
 det/.style={rectangle, draw=cGreen, line width=0.8pt, dashed, rounded corners=3pt, fill=cGreen!7, align=left, font=\small, text width=6.5cm, inner sep=7pt, text=ink},
 arr/.style={-{Stealth[length=2.4mm]}, draw=inksoft, line width=1pt},
 exarr/.style={-{Stealth[length=2mm]}, draw=hair, line width=0.9pt},
 link/.style={-{Stealth[length=1.6mm]}, draw=hair, line width=0.6pt, dashed},
]
\node[pstep, draw=cBlue, fill=cBlue!8] (s1) at (0,0) {\includegraphics[width=0.8cm]{user.PNG}\\\textbf{Input}\\\tiny User provides\\\tiny math problem};
\node[pstep, draw=cTeal, fill=cTeal!8] (s2) at (3.8,0) {\includegraphics[width=0.8cm]{symbol.PNG}\\\textbf{Parse}\\\tiny Extract variables\\\tiny and quantities};
\node[pstep, draw=cAmber, fill=cAmber!8] (s3) at (7.6,0) {\includegraphics[width=0.8cm]{lightbulb_icon.PNG}\\\textbf{Reason}\\\tiny Select formula\\\tiny and substitute};
\node[pstep, draw=cViolet, fill=cViolet!8] (s4) at (11.4,0) {\includegraphics[width=0.8cm]{calculator_icon.PNG}\\\textbf{Execute}\\\tiny Perform\\\tiny arithmetic};
\node[pstep, draw=cGreen, fill=cGreen!8] (s5) at (15.2,0) {\includegraphics[width=0.8cm]{check_icon.PNG}\\\textbf{Answer}\\\tiny Final result\\\tiny with units};
\draw[arr] (s1)--(s2); \draw[arr] (s2)--(s3); \draw[arr] (s3)--(s4); \draw[arr] (s4)--(s5);
\node[ex] (e1) at (0,-3.4) {``A train travels 60 km\\in 1.5 hours. What is\\its average speed?''};
\node[ex] (e2) at (3.8,-3.4) {Identifies:\\- Distance = 60 km\\- Time = 1.5 hours\\- Task: speed = $d/t$};
\node[ex] (e3) at (7.6,-3.4) {Formula: $v = d/t$\\Substitutes:\\$60 / 1.5 = ?$};
\node[ex] (e4) at (11.4,-3.4) {Arithmetic execution:\\$60 \div 1.5 = 40$};
\node[ex] (e5) at (15.2,-3.4) {\textbf{Answer:}\\``40 km/h''};
\draw[exarr] (e1)--(e2); \draw[exarr] (e2)--(e3); \draw[exarr] (e3)--(e4); \draw[exarr] (e4)--(e5);
\draw[link] (s1)--(e1); \draw[link] (s5)--(e5);
\node[det] (d1) at (3.8,-6.4) {\textbf{Inside the LLM (Parsing):}\\- Token-level attention on numerical values and keywords\\- Implicit symbolic grounding using learned arithmetic patterns};
\node[det] (d2) at (11.4,-6.4) {\textbf{Inside the LLM (Computation):}\\- Generates intermediate tokens representing computation steps\\- Uses decoder logic + positional embeddings to simulate math};
\draw[link] (e2)--(d1); \draw[link] (e4)--(d2);
\end{tikzpicture}
}
\caption{Mathematical reasoning flow. Row~1: generic LLM reasoning pipeline (shared across domains). Row~2: math-specific example showing symbol extraction, formula application, and arithmetic execution. Row~3: what happens inside the model at key stages.}
\label{fig:math-reasoning-flow}
\end{figure}

\begin{table}[htbp]
\centering
\small
\begin{tabularx}{\textwidth}{ >{\raggedright\arraybackslash}p{0.25\textwidth} >{\raggedright\arraybackslash\hsize=0.8\hsize\linewidth=\hsize}X >{\raggedright\arraybackslash\hsize=0.8\hsize\linewidth=\hsize}X >{\raggedright\arraybackslash\hsize=1.4\hsize\linewidth=\hsize}X }
\toprule
\textbf{Author} & \textbf{Datasets} & \textbf{Models} & \textbf{Evaluation Setup} \\
\midrule

\multicolumn{4}{l}{\textbf{Architectural Innovation and Model Development}} \\

Shao et al.~\citep{shao2024deepseekmath} & DeepSeekMath (120B tokens), GSM8K, MATH, OCW, SAT-MMLU, STEM, CMATH, Gaokao & DeepSeek-LLM 1.3B, DeepSeekMath-7B & Few-shot CoT across all benchmarks \\

Ying et al.~\citep{ying2024internlm} & GSM8K, MATH, MiniF2F, Hungary Exam, MathBench-ZH & InternLM2-Math (7B, 20B) & In-context majority voting, LEAN verification, ORM/PRM reward \\

Zeng et al.~\citep{zeng2024skywork} & GSM8K, MATH & LLaMA2-7B, Mistral-7B, DeepSeekMath-7B & Strict regex evaluation, vLLM decoding \\

Zhang et al.~\citep{zhang2024infinitymath} & InfinityMath, GSM8K, MATH, AQuA, SVAMP, MMLU-Math & LLaMA2, CodeLlama, Aquila2-7B & 3-epoch SFT, docstrings+PoT boost \\

Wang et al.~\citep{wang2024mathpile} & MATHPILE (9.45B tokens), GSM8K, AQuA, MATH, STEM & Mistral-7B, LLaMA3, DeepSeekMath & Few-shot CoT, strict evaluation \\

Zhang et al.~\citep{zhang2024initialization} & Synthetic anchor mapping dataset & GPT-2, Transformer & Accuracy on held-out pairs, t-SNE activation analysis \\

Liu et al.~\citep{liu2025quantization} & AIME-120, MATH-500, GSM8K, GPQA-Diamond & DeepSeek, QwQ-32B, LLaMA (quantized) & AWQ/GPTQ/KVQuant accuracy impact \\

\midrule
\multicolumn{4}{l}{\textbf{Data Scaling and Synthetic Data Generation}} \\

Li et al.~\citep{li2024common} & GSM8K, MATH, SVAMP, ASDiv & Xwin-Math (7B–70B) & SFT with GPT-4 Turbo synthetic data \\

Tang et al.~\citep{tang2024mathscale} & MathScaleQA (2M), GSM8K, MATH & MathScale models & Alpaca-style SFT, macro/micro accuracy \\

Toshniwal et al.~\citep{toshniwal2024openmathinstruct} & OpenMath (1.8M), GSM8K, MATH & LLaMA2, Mistral, CodeLLaMA & Zero-shot evaluation across benchmarks \\

Tong et al.~\citep{tong2024dart} & GSM8K, MATH, CollegeMath & LLaMA3, Mistral, DeepSeekMath & DART-Hard dataset with MCTS-sampled CoT \\

\midrule
\multicolumn{4}{l}{\textbf{Process Supervision and Step-by-Step Reasoning}} \\

Luo et al.~\citep{luo2024improve} & MATH500, GSM8K & Gemini Pro, Gemma2-27B & PRM supervision, majority voting pass@64 \\

Zhang et al.~\citep{zhang2025booststep} & MATH500, AQuA, AMC & GPT-4o, Qwen2.5-72B & BoostStep single-step CoT improvement \\

Chen et al.~\citep{chen2024alphamath} & GSM8K, MATH, OCW & DeepSeekMath-7B & Step-by-step supervision with MCTS search \\

\bottomrule
\end{tabularx}
\caption{Experimental setup overview of mathematical reasoning papers (Part~1).}
\label{tab:mathreasoning-summary-part1}
\end{table}

\begin{table}[htbp]
\centering
\small
\begin{tabularx}{\textwidth}{ >{\raggedright\arraybackslash}p{0.25\textwidth} >{\raggedright\arraybackslash\hsize=0.8\hsize\linewidth=\hsize}X >{\raggedright\arraybackslash\hsize=0.8\hsize\linewidth=\hsize}X >{\raggedright\arraybackslash\hsize=1.4\hsize\linewidth=\hsize}X }
\toprule
\textbf{Author} & \textbf{Datasets} & \textbf{Models} & \textbf{Evaluation Setup} \\
\midrule

\multicolumn{4}{l}{\textbf{Search, Self-Refinement, and Inference-Time Optimization}} \\

Zhang et al.~\citep{zhang2024llama} & GSM8K, MATH500, AIME24 & LLaMA-3.1-8B, Gemma2-2B & SR-MCTS search, rm@16, maj@k \\

Guan et al.~\citep{guan2025rstar} & GSM8K, MATH-500, AIME24 & Qwen2.5-Math, Phi3-mini & Pass@1 with rollout search \\

Xiong et al.~\citep{xiong2025self} & MATH500, OlympiadBench & Qwen2.5-Math-7B & Self-correction metrics \\

Sun et al.~\citep{sun2024easy} & MATH500 & PRM800K, MetaMATH & BoN, Maj@16/256 evaluation \\

\midrule
\multicolumn{4}{l}{\textbf{Evaluation and Robustness Assessment}} \\

Zhou et al.~\citep{zhou2024your} & MATHCHECK-GSM, MATHCHECK-GEO & 43 LLMs/MLLMs & Accuracy and F1 across reasoning tasks \\

Mirzadeh et al.~\citep{mirzadeh2024gsm} & GSM-Symbolic & 25+ LLMs & Robustness drop vs GSM8K \\

Sun et al.~\citep{sun2025challenging} & OlymMATH benchmark & Qwen3, Gemini, DeepSeek & Pass@1, Consistency@64 \\

Wang et al.~\citep{wang2024embedding} & MultiArith, GSM8K & LLaMA2-7B, GPT2-XL & TV score and AUROC evaluation \\

\midrule
\multicolumn{4}{l}{\textbf{Reinforcement Learning and Reward-Based Training}} \\

Lyu et al.~\citep{lyu2025exploring} & MATH500, AIME2024 & GPT-4o, Claude-3.5 & Outcome reward models \\

Li et al.~\citep{li2025sos1} & MATH500 benchmarks & Qwen2.5-Math-7B & PPO, DPO, reward model comparison \\

Chen et al.~\citep{chen2024not} & ASDIV, GSM8K, MATH500 & Qwen-QwQ-32B, DeepSeek-R1 & Overthinking analysis \\

\midrule
\multicolumn{4}{l}{\textbf{Metacognitive and Self-Reflective Reasoning}} \\

Didolkar et al.~\citep{didolkar2024metacognitive} & GSM8K, MATH, SVAMP & GPT-4, Mixtral & Skill-based CoT reasoning \\

Zhang et al.~\citep{zhang2024learn} & GSM8K, MATH, MAWPS & Mistral-7B, LLaMA-3-8B & RefAug + QA augmentation evaluation \\

\bottomrule
\end{tabularx}
\caption{Experimental setup overview of mathematical reasoning papers (Part~2).}
\label{tab:mathreasoning-summary-part2}
\end{table}

\subsection{Commonsense Reasoning}
\label{reason:IV}
\begin{figure}[htbp]
\centering
\definecolor{ink}{RGB}{38,41,54}
\definecolor{inksoft}{RGB}{92,97,115}
\definecolor{hair}{RGB}{200,205,216}
\definecolor{cBlue}{RGB}{60,87,150}
\definecolor{cTeal}{RGB}{46,125,111}
\definecolor{cAmber}{RGB}{168,120,43}
\definecolor{cViolet}{RGB}{103,87,140}
\definecolor{cGreen}{RGB}{92,136,85}
\scalebox{0.8}{
\begin{tikzpicture}[
 pstep/.style={rectangle, draw, line width=0.9pt, rounded corners=4pt, minimum height=1.5cm, minimum width=2.6cm, align=center, font=\small, text=ink},
 ex/.style={rectangle, draw=hair, line width=0.7pt, rounded corners=3pt, fill=ink!3, minimum height=0.7cm, align=center, font=\small, text width=3.2cm, inner sep=5pt, text=ink},
 det/.style={rectangle, draw=cGreen, line width=0.8pt, dashed, rounded corners=3pt, fill=cGreen!7, align=left, font=\small, text width=6.5cm, inner sep=7pt, text=ink},
 arr/.style={-{Stealth[length=2.4mm]}, draw=inksoft, line width=1pt},
 exarr/.style={-{Stealth[length=2mm]}, draw=hair, line width=0.9pt},
 link/.style={-{Stealth[length=1.6mm]}, draw=hair, line width=0.6pt, dashed},
]
\node[pstep, draw=cBlue, fill=cBlue!8] (s1) at (0,0) {\includegraphics[width=0.8cm]{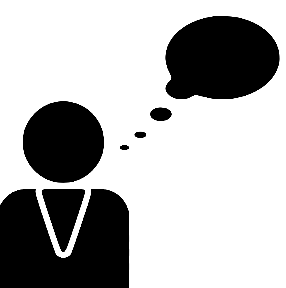}\\\textbf{Input}\\\tiny User poses\\\tiny everyday query};
\node[pstep, draw=cTeal, fill=cTeal!8] (s2) at (3.8,0) {\includegraphics[width=0.8cm]{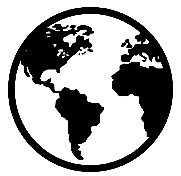}\\\textbf{Parse}\\\tiny Activate world\\\tiny knowledge};
\node[pstep, draw=cAmber, fill=cAmber!8] (s3) at (7.6,0) {\includegraphics[width=0.8cm]{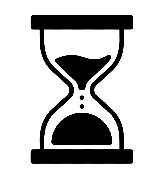}\\\textbf{Reason}\\\tiny Build causal\\\tiny chain};
\node[pstep, draw=cViolet, fill=cViolet!8] (s4) at (11.4,0) {\includegraphics[width=0.8cm]{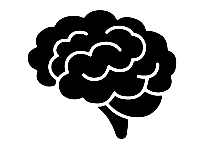}\\\textbf{Infer}\\\tiny Fill implicit\\\tiny steps};
\node[pstep, draw=cGreen, fill=cGreen!8] (s5) at (15.2,0) {\includegraphics[width=0.8cm]{check_icon.PNG}\\\textbf{Answer}\\\tiny Plausible\\\tiny conclusion};
\draw[arr] (s1)--(s2); \draw[arr] (s2)--(s3); \draw[arr] (s3)--(s4); \draw[arr] (s4)--(s5);
\node[ex] (e1) at (0,-3.4) {``If you put ice in\\the sun, what\\happens?''};
\node[ex] (e2) at (3.8,-3.4) {Activates world\\knowledge:\\- Ice melts under heat\\- Sun emits heat\\- Melting is gradual};
\node[ex] (e3) at (7.6,-3.4) {Causal chain:\\Sun $\to$ Heat\\$\to$ Ice melts\\Fills missing steps};
\node[ex] (e4) at (11.4,-3.4) {Predicts outcome\\based on causal\\plausibility};
\node[ex] (e5) at (15.2,-3.4) {\textbf{Answer:}\\``The ice will melt.''};
\draw[exarr] (e1)--(e2); \draw[exarr] (e2)--(e3); \draw[exarr] (e3)--(e4); \draw[exarr] (e4)--(e5);
\draw[link] (s1)--(e1); \draw[link] (s5)--(e5);
\node[det] (d1) at (3.8,-6.4) {\textbf{Inside the LLM (Context):}\\- Embeddings link ``ice'', ``sun'', ``happens'' to environmental and thermal contexts\\- Causal associations retrieved from pretraining corpus};
\node[det] (d2) at (11.4,-6.4) {\textbf{Inside the LLM (Inference):}\\- Uses next-token generation for causally coherent outcomes\\- Leverages masked prediction to guess intermediate states};
\draw[link] (e2)--(d1); \draw[link] (e4)--(d2);
\end{tikzpicture}
}
\caption{Commonsense reasoning flow. Row~1: generic pipeline. Row~2: commonsense example showing world-knowledge activation, causal chaining, and plausibility-based inference. Row~3: internal model mechanisms at context and inference stages.}
\label{fig:commonsense-reasoning-flow}
\end{figure}

Commonsense reasoning in large language models (LLMs) is the ability to draw plausible inferences about everyday situations, physical causality, social norms, temporal order, spatial relations, beyond rote fact recall. It requires contextual abstraction and the integration of heterogeneous signals (text, vision, structure). Figure~\ref{fig:commonsense-reasoning-flow} outlines this process, showing how a model activates world knowledge, builds a causal chain, and fills implicit gaps to reach a plausible conclusion. As LLMs move into real applications, robustness on such reasoning is central to safe, coherent interaction. Below we group recent work into four themes, measurement and audits; multimodal and visual grounding; causal/scientific structure; and knowledge-graph composition, followed by cross-cutting takeaways. Table~\ref{tab:commonsense-summary} lists the experimental setups.

\subsubsection{Causal and Scientific Commonsense}
These approaches shape reasoning around cause and effect or structured scientific thinking, often by introducing frameworks that make intermediate steps more transparent.~\citep{kiciman2023causal} find that GPT-3.5 and GPT-4 perform well on tasks like pairwise cause identification, counterfactuals, and assessing necessity or sufficiency. These models also generalize to new datasets, pointing to a real ability for causal insight and opening up the possibility of combining neural models with symbolic tools. ~\citep{ouyang2023structured} present StructChem, a prompting method that breaks down chemistry questions into layered stages. Without extra training, GPT-4 can use these structured prompts to locate relevant formulas and apply them step by step, leading to as much as 30 percent better performance. ~\citep{chen2024huatuogpt} introduce HuatuoGPT-o1, a medical reasoning model trained in two phases. It first learns from verifier-guided examples, then improves using reinforcement learning based on feedback from a medical verifier across 40K verified problems. This makes the model more accurate and lets users trace its reasoning, which is especially valuable in clinical use. By adding discipline-specific structure, such as verifiers or staged prompts, these methods help language models turn intuition into clear, checkable steps, making them more reliable for tasks where errors carry serious consequences.

\subsubsection{Knowledge Graphs \& Logical Composition}

Language models can gain commonsense abilities by composing facts and relationships from structured knowledge graphs. This often involves rephrasing these graphs into natural language tasks or using them directly during inference. One approach, COM2, introduced by~\citep{fang2024complex}, creates complex, multi-step reasoning questions by sampling logical queries from commonsense knowledge graphs and then turning them into coherent narratives using templates and language models. These queries include multiple reasoning steps, such as identifying joint causes or effects of events. COM2 improves both zero-shot question answering and generative tasks by providing models with richer logical structures embedded in natural text, without needing extensive human annotations. Another method is KnowZRel, ~\citep{khan2025knowzrel}, which targets visual reasoning through scene graph generation. KnowZRel detects objects in images, refines them using spatial and structural cues, and then retrieves relationships between them from a heterogeneous commonsense knowledge graph. This allows the model to predict unseen relationships, achieving strong gains in zero-shot recall and performing well even with rare or underrepresented visual relationships. Another alternative angle on structured commonsense comes from SenticNet 7~\citep{cambria2022senticnet}, which integrates affective and conceptual knowledge into a hybrid commonsense reasoning framework. It constructs a large-scale graph that blends symbolic affective concepts with deep neural features, enabling reasoning that spans both factual and emotional dimensions. By embedding knowledge from multiple sources, including ConceptNet~\citep{speer2017conceptnet}, AffectNet~\citep{mollahosseini2017affectnet}, and Open Mind Common Sense~\citep{singh2002open}, into a unified structure optimized for concept-level sentiment inference, SenticNet~\citep{cambria2022senticnet} enables LLMs and affective computing systems to reason over nuanced human situations, such as intent, mood, and interpersonal context.

\subsubsection{Measuring \& Auditing Commonsense}
This line of research focuses on how we evaluate commonsense in LLMs, critiquing where current benchmarks fall short and introducing cleaner alternatives to avoid overestimating model capabilities. ~\citep{chizhov2025hellaswag} revisit the widely-used HellaSWAG benchmark~\citep{chizhov2025hellaswag} and expose key validity issues, such as ungrammatical choices and misleading distractors. They show that models can often select the same answer even when fed irrelevant input, suggesting shallow pattern matching rather than true reasoning. To address this, they propose GoldenSwag~\citep{zellers2018swag}, a cleaned and more rigorously constructed subset that better reflects genuine commonsense evaluation. ~\citep{davis2023benchmarks} surveys 139 benchmarks across text, vision, and simulation, highlighting systemic issues like annotation noise, narrow coverage, and poor construct clarity. Their work offers concrete principles for building more trustworthy commonsense benchmarks moving forward. 

\subsubsection{Multimodal \& Visual Commonsense}
This area focuses on how models integrate visual signals, images, or video, with language to support commonsense reasoning, requiring coherent alignment between perceptual input and textual inference. ~\citep{wang2023gemini} conduct a comprehensive evaluation of Gemini, assessing its performance across 12 commonsense reasoning datasets spanning both text-only and multimodal formats. When benchmarked against four language-only models and two vision-language models, Gemini demonstrates strong performance, particularly in its ability to handle nuanced multimodal scenarios. The results underscore the importance of broad, unified evaluation frameworks that capture diverse dimensions of commonsense. ~\citep{zhang2024common} reframes deepfake detection as a visual question answering (VQA) task in DD-VQA, introducing a vision–language model trained to identify and explain visual inconsistencies, such as facial warping or unnatural lighting. By generating textual rationales for visual anomalies, the model achieves both higher accuracy and greater interpretability, reflecting a human-like form of commonsense assessment. Grounding commonsense in visual data demands tight alignment between perceptual cues and textual reasoning. When models are trained to bridge this gap explicitly, they become more accurate and transparent in their judgments.

\begin{table}[htbp]
\centering
\small
\begin{tabularx}{\textwidth}{ >{\raggedright\arraybackslash}p{0.25\textwidth} >{\raggedright\arraybackslash\hsize=0.8\hsize\linewidth=\hsize}X >{\raggedright\arraybackslash\hsize=0.8\hsize\linewidth=\hsize}X >{\raggedright\arraybackslash\hsize=1.4\hsize\linewidth=\hsize}X }
\toprule
\textbf{Author} & \textbf{Datasets} & \textbf{Models} & \textbf{Evaluation Setup} \\
\midrule

\multicolumn{4}{l}{\textbf{Causal and Scientific Commonsense}} \\

Kıcıman et al.~\citep{kiciman2023causal} & Tübingen Cause-Effect Pairs (108 pairs across 37 domains) & text-davinci-002/003, GPT-3.5 & Prompt-tuned bidirectional causal queries; weighted and standard accuracy \\

Ouyang et al.~\citep{ouyang2023structured} & SciBench: quan, chemmc, atkins, matter (Pw, Ps splits) & GPT-3.5, GPT-4, Vicuna-13B, LLaMA-2-13B-chat & Zero-/few-shot evaluation with deviation thresholds \\

Chen et al.~\citep{chen2024huatuogpt} & MedQA, MedMCQA, PubMedQA, MMLU-Pro, GPQA & HuatuoGPT-o1 (8B,70B), Qwen2.5, LLaMA, Gemma2 & Two-stage SFT+RLHF with verifier; accuracy metric \\

\midrule
\multicolumn{4}{l}{\textbf{Knowledge Graphs \& Logical Composition}} \\

Fang et al.~\citep{fang2024complex} & COM2, aNLI, CSQA, PIQA, SIQA, WG, MEI, PCD, ATM10x & GPT-3.5/4, DeBERTa-v3, COMET, CAR & Multi-choice QA + generative metrics (BLEU, ROUGE, CIDEr) \\

Khan et al.~\citep{khan2025knowzrel} & VG, GQA, COM2, MedQA, PubMedQA, Tübingen, SciBench & LLaMA-3.1, HuatuoGPT, GPT-4, COMET, TransE & Zero/few-shot CoT, zR@K metrics, SGG evaluation \\

SenticNet 7~\citep{cambria2022senticnet} & CR, MR, Amazon, IMDb, Sanders, SST, STS, SE13, SE15, SE16 & SenticNet 7 (compared with 20 sentiment lexica ) & Binary sentiment classification (positive vs negative) comparing lexicon-based sentiment analysis \\

\midrule
\multicolumn{4}{l}{\textbf{Measuring \& Auditing Commonsense}} \\

Chizhov et al.~\citep{chizhov2025hellaswag} & HellaSwag (annotated) & Claude 3.5 Sonnet, GPT variants & Zero-prompt ablations, shuffled choices, likelihood evaluation \\

\midrule
\multicolumn{4}{l}{\textbf{Multimodal \& Visual Commonsense}} \\

Wang et al.~\citep{wang2023gemini} & 12 commonsense datasets including CommonsenseQA, PIQA, Social IQa, VCR & LLaMA-2-70B, GPT-3.5/4 Turbo, Gemini Pro, GPT-4V & Zero-shot SP and few-shot CoT; accuracy metric \\

Zhang et al.~\citep{zhang2024common} & DD-VQA (FaceForensics++, 2968 images) & Custom VQA model with contrastive learning & Accuracy and explanation plausibility for image-question reasoning \\

\bottomrule
\end{tabularx}
\caption{Experimental setup overview of commonsense reasoning papers.}
\label{tab:commonsense-summary}
\end{table}

\subsection{Visual and Multimodal Reasoning}
\label{reason:V} 
\begin{figure}[htbp]
\centering
\definecolor{ink}{RGB}{38,41,54}
\definecolor{inksoft}{RGB}{92,97,115}
\definecolor{hair}{RGB}{200,205,216}
\definecolor{cBlue}{RGB}{60,87,150}
\definecolor{cTeal}{RGB}{46,125,111}
\definecolor{cAmber}{RGB}{168,120,43}
\definecolor{cViolet}{RGB}{103,87,140}
\definecolor{cGreen}{RGB}{92,136,85}
\scalebox{0.8}{
\begin{tikzpicture}[
 pstep/.style={rectangle, draw, line width=0.9pt, rounded corners=4pt, minimum height=1.5cm, minimum width=2.6cm, align=center, font=\small, text=ink},
 ex/.style={rectangle, draw=hair, line width=0.7pt, rounded corners=3pt, fill=ink!3, minimum height=0.7cm, align=center, font=\small, text width=3.2cm, inner sep=5pt, text=ink},
 det/.style={rectangle, draw=cGreen, line width=0.8pt, dashed, rounded corners=3pt, fill=cGreen!7, align=left, font=\small, text width=6.5cm, inner sep=7pt, text=ink},
 arr/.style={-{Stealth[length=2.4mm]}, draw=inksoft, line width=1pt},
 exarr/.style={-{Stealth[length=2mm]}, draw=hair, line width=0.9pt},
 link/.style={-{Stealth[length=1.6mm]}, draw=hair, line width=0.6pt, dashed},
]
\node[pstep, draw=cBlue, fill=cBlue!8] (s1) at (0,0) {\includegraphics[width=0.8cm]{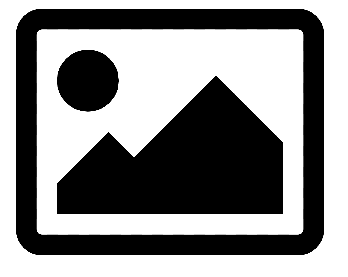}\\\textbf{Input}\\\tiny Image + text\\\tiny prompt};
\node[pstep, draw=cTeal, fill=cTeal!8] (s2) at (3.8,0) {\includegraphics[width=0.8cm]{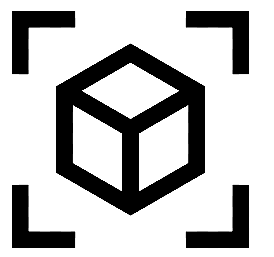}\\\textbf{Encode}\\\tiny Vision encoder\\\tiny extracts features};
\node[pstep, draw=cAmber, fill=cAmber!8] (s3) at (7.6,0) {\includegraphics[width=0.8cm]{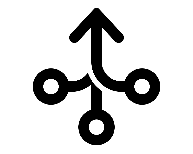}\\\textbf{Fuse}\\\tiny Cross-modal\\\tiny alignment};
\node[pstep, draw=cViolet, fill=cViolet!8] (s4) at (11.4,0) {\includegraphics[width=0.8cm]{brain_icon.PNG}\\\textbf{Reason}\\\tiny Multimodal\\\tiny inference};
\node[pstep, draw=cGreen, fill=cGreen!8] (s5) at (15.2,0) {\includegraphics[width=0.8cm]{check_icon.PNG}\\\textbf{Answer}\\\tiny Verbalized\\\tiny response};
\draw[arr] (s1)--(s2); \draw[arr] (s2)--(s3); \draw[arr] (s3)--(s4); \draw[arr] (s4)--(s5);
\node[ex] (e1) at (0,-3.4) {[Image: Traffic Sign]\\+ ``What should the\\driver do?''};
\node[ex] (e2) at (3.8,-3.4) {Vision encoder:\\- Object detection\\- Scene layout\\- Symbol extraction};
\node[ex] (e3) at (7.6,-3.4) {Cross-attention aligns\\image regions with\\text tokens};
\node[ex] (e4) at (11.4,-3.4) {Infers intent of sign\\$\to$ matches rule\\$\to$ verbalizes};
\node[ex] (e5) at (15.2,-3.4) {\textbf{Output:}\\``The driver\\should stop.''};
\draw[exarr] (e1)--(e2); \draw[exarr] (e2)--(e3); \draw[exarr] (e3)--(e4); \draw[exarr] (e4)--(e5);
\draw[link] (s1)--(e1); \draw[link] (s5)--(e5);
\node[det] (d1) at (3.8,-6.4) {\textbf{Inside the Model (Fusion):}\\- Cross-attention layers align image regions with tokens\\- Hidden states encode joint visual-linguistic semantics};
\node[det] (d2) at (11.4,-6.4) {\textbf{Inside the Model (Reasoning):}\\- Leverages visual-language embeddings to infer intent\\- Simulates human-like decision through next-token generation};
\draw[link] (e3)--(d1); \draw[link] (e4)--(d2);
\end{tikzpicture}
}
\caption{Multimodal reasoning flow. Row~1: generic vision-language pipeline. Row~2: example showing image encoding, cross-modal fusion, and visual intent inference. Row~3: internal mechanisms at fusion and reasoning stages.}
\label{fig:multimodal-reasoning-flow}
\end{figure}

Visual and multimodal reasoning represents a fundamental expansion of LLM capabilities beyond text, requiring models to interpret visual scenes, align language with visual cues, infer spatial relationships, and draw grounded conclusions from multimodal evidence. This capability is increasingly critical as LLMs evolve from conversational agents to comprehensive assistants capable of interacting with visual interfaces, analyzing documents, understanding videos, and supporting real-world applications in robotics and automation. The challenge lies not merely in processing visual inputs, but in achieving genuine reasoning that integrates visual perception with symbolic understanding, spatial cognition, and temporal awareness.
The complexity of visual reasoning stems from several inherent challenges: the need to bridge the gap between continuous visual representations and discrete symbolic reasoning, the requirement to handle multiple modalities simultaneously while maintaining coherent reasoning chains, and the demand for robust performance across diverse visual domains from mathematical diagrams to real-world scenes. Figure~\ref{fig:multimodal-reasoning-flow} depicts this pipeline, tracing how an image and text prompt are encoded, fused through cross-modal alignment, and passed through multimodal inference to produce a verbalized response. Recent research has approached these challenges through complementary strategies that we organize into six key areas, each addressing different aspects of the visual reasoning problem. Tables~\ref{tab:vlm-summary-part1} and~\ref{tab:vlm-summary-part2} detail the experimental setups.

\subsubsection{Modular Controllers \& Active Retrieval}
Rather than treating visual reasoning as a passive perception task, recent approaches have reconceptualized LLMs as active controllers that can orchestrate perception modules and search over intermediate states. This paradigm shift enables more dynamic and flexible visual reasoning by treating perception as an interactive process.
 ~\citep{stanic2024towards} use LLMs as programmers that compose spatial and temporal abstractions with visual tools, removing the need for heavy manual in-context prompt engineering and enabling truly zero-shot compositional reasoning. Their approach demonstrates how LLMs can dynamically construct reasoning pipelines tailored to specific visual tasks. ~\citep{dong2024progressive} propose AR-MCTS, which couples active retrieval with Monte Carlo Tree Search, allowing models to fetch visual and text evidence as they explore different reasoning branches. This approach significantly improves both diversity and accuracy across complex multimodal tasks by treating evidence gathering as an integral part of the reasoning process.
The key insight from these approaches is that treating LLMs as controllers that are capable of calling tools, retrieving information, and exploring multiple reasoning paths can transform static perception into a dynamic search procedure that can generalize beyond training distributions and adapt to novel visual reasoning challenges.

\subsubsection{Math \& Diagrammatic Reasoning}
Mathematical diagrams present a particularly challenging domain for visual reasoning, requiring models to jointly parse spatial information and perform symbolic manipulation. Unlike text-based mathematical problems, diagrammatic reasoning demands faithful interpretation of visual elements rather than relying on textual shortcuts or descriptions.
MathVerse~\citep{zhang2024mathverse} makes this challenge explicit by converting 2,612 visual math problems into nearly 15k variants and introducing chain-of-thought–based evaluation. Results reveal that models frequently produce correct-looking answers without truly using the diagram, exposing how brittle diagram-to-symbol grounding remains. To counteract such gaps, MathV360K~\citep{shi2024math} scales targeted multimodal training, bootstrapping LLaVA-1.5 with a large curated visual math dataset. This yields strong open-source gains and begins to close the gap with GPT-4V, underscoring how domain-specific corpora can realign models toward pixel-faithful reasoning.
But stronger training alone does not ensure deeper reasoning. We-Math~\citep{qiao2024we} introduces concept- and depth-structured diagnostics, showing that while most models plateau at shallow recall, GPT-4o is among the first to demonstrate more generalizable reasoning strategies. This diagnostic approach highlights where diagrammatic reasoning breaks down, often in multi-step abstraction, offering a lens beyond raw accuracy. Complementing this, InfiMM-WebMath-40B~\citep{han2024infimm} shows that curated large-scale multimodal pretraining allows even smaller models to match or surpass larger ones on MathVerse and We-Math, proving that well-structured corpora can substitute for sheer parameter count.
Finally, HumanEval-V~\citep{zhang2024humaneval} extends the challenge by pairing diagrams with code-based unit tests. Across 22 models, performance on spatial and topological reasoning remains strikingly low (max pass@k $\approx$ 36.8\%), emphasizing that even when symbolic checking is available, diagram interpretation remains the bottleneck.
These findings highlight that while visual mathematics improves with targeted data and scaffolding approaches, the fundamental challenge of faithful diagram-to-symbol translation remains a critical bottleneck, particularly for spatial topology and geometric transformations.

\subsubsection{Grounded Reasoning in Structured Visual Media}
Document understanding requires models to handle plots, tables, forms, and multi-page layouts while maintaining coherent reasoning across visual and textual sources. Recent approaches fall into two categories: perceptual pretext tasks that connect visual elements to structured representations, and retrieval pipelines that preserve visual structure across pages.

\begin{itemize}
\item \textbf{Chart-to-table translation.} ~\citep{carbune2024chart} fine-tune PaLI3-5B into ChartPaLI-5B through chart-to-table translation and synthetic trace generation, surpassing much larger models on chart reasoning tasks.

\item \textbf{Direct chart training.} ~\citep{masry2024chartgemma} train ChartGemma directly on chart images without gold table annotations, improving trend reading and low-level detail extraction across summarization, QA, and fact-checking.

\item \textbf{Chart-to-code generation.}~\citep{yang2024chartmimic} introduce ChartMimic with 4,800 curated triplets and multi-level evaluation metrics, exposing alignment gaps between visual parsing and code synthesis.

\item \textbf{Multi-document multimodal retrieval.} M3DocRAG~\citep{cho2024m3docrag} combines multimodal retrieval with generation for multi-hop queries across document pages. ~\citep{suri2024visdom} extend this with VisDoMRAG, achieving 12--20\% improvements over unimodal and long-context baselines on visually rich multi-document QA.
\end{itemize}

\subsubsection{Externalized Visual Reasoning \& Interpretability}
Recent work makes visual reasoning more transparent and more trainable through two directions: externalizing intermediate states and applying reinforcement learning tailored to perception.

\begin{itemize}
\item \textbf{Externalized reasoning traces.} Visual Sketchpad~\citep{hu2024visual} lets models sketch boxes and lines as a visual chain-of-thought, boosting accuracy on spatial tasks while providing human-interpretable traces. Sparse autoencoders applied to LLaVA-NeXT-8B~\citep{zhang2024large} complement this by revealing when hidden representations align with semantic parts, exposing how much structure the model already encodes internally.

\item \textbf{Reinforcement-based perception training.} Visual-RFT~\citep{liu2025visual} introduces perception-verifiable rewards, showing that RL fine-tuning outperforms supervised training on few-shot visual reasoning. OThink-MR1~\citep{liu2025othink} extends this with GRPO-D (Group Relative Policy Optimization with dropout), yielding over 60\% average improvements across multimodal benchmarks.
\end{itemize}

\noindent Making intermediate reasoning visible, whether through external sketches or internal feature analysis, strengthens both performance and trust in multimodal systems.

\subsubsection{Temporal Reasoning Across Long-Horizon Video}
Temporal reasoning in video requires coherence over long sequences, causal attribution across frames, and strict computational efficiency. Recent work addresses three complementary fronts:

\begin{itemize}
\item \textbf{Evaluation.} TOMATO~\citep{shangguan2024tomato} introduces fine-grained metrics (Multi-Frame Gain, Frame Order Sensitivity) and exposes a 57.3\% gap between human and model performance, showing that models often fail to track basic temporal order.

\item \textbf{Learning.} Video-R1~\citep{feng2025video} applies Temporal Group Relative Policy Optimization (T-GRPO) to train models to assign credit over sequences, outperforming GPT-4o on VSI-Bench~\citep{yang2025thinking} and MVBench~\citep{li2024mvbench}.

\item \textbf{Efficiency.} STORM~\citep{jiang2025token} combines temporal encoders with token-reduction strategies, preserving temporal order while cutting compute by up to $8\times$ and latency by nearly $3\times$, with measurable accuracy gains on long-video benchmarks.
\end{itemize}

\noindent Closing the long-horizon video reasoning gap requires progress on all three fronts: evaluation that makes deficiencies visible, RL that supplies temporal credit assignment, and architectures that manage efficiency trade-offs.

\subsubsection{Evaluation and Benchmarks}
Visual reasoning evaluation requires benchmarks that resist language-only shortcuts and test genuine perception and abstraction. Table~\ref{tab:visual-eval-benchmarks} organizes recent benchmarks by the layer of difficulty they target.

\begin{table}[htbp]
\centering
\small
\begin{tabularx}{\textwidth}{ >{\raggedright\arraybackslash}p{0.18\textwidth} >{\raggedright\arraybackslash}p{0.17\textwidth} >{\raggedright\arraybackslash}X >{\raggedright\arraybackslash}p{0.22\textwidth} }
\toprule
\textbf{Benchmark} & \textbf{Domain} & \textbf{What it probes} & \textbf{Key finding} \\
\midrule
Image--dialogue games \citep{hakimov2024two} & Grounded interaction & Situation dynamics in goal-oriented visual games & Open models fail at dynamics; closed models rely on superficial captioning \\
\addlinespace
MapEval \citep{dihan2024mapeval} & Geo-spatial reasoning & Map-based QA across text, visual, and API modes & Models trail human performance by over 20\% \\
\addlinespace
MM-IQ \citep{cai2025mm} & Abstract reasoning & Higher-order abstraction across 8 reasoning paradigms & State-of-the-art MLLMs perform at near-chance levels \\
\addlinespace
LEGO-Puzzles \citep{tang2025lego} & Spatial reasoning & Multi-step spatial assembly and manipulation & Models score far below human accuracy \\
\addlinespace
SWE-bench Multimodal \citep{tang2025lego} & Visual UI debugging & JavaScript bug repair from visual interfaces & Leading systems solve only 12\% of cases \\
\addlinespace
RPM-style \citep{zhang2024far} & Visual deduction & Abstract pattern recognition (Raven's Progressive Matrices) & Neither CoT nor in-context learning overcomes perception limits \\
\bottomrule
\end{tabularx}
\caption{Visual reasoning benchmarks by difficulty layer. Across grounded interaction, spatial reasoning, and abstract pattern recognition, models remain brittle when reasoning depends on visual evidence.}
\label{tab:visual-eval-benchmarks}
\end{table}

\noindent The consistent finding is that multimodal models remain brittle when reasoning genuinely depends on visual evidence. Progress in visual reasoning cannot be measured by language-side performance alone.

\subsubsection{Summary \& Future Directions}
Visual and multimodal reasoning research has converged on three key strategies: developing modular controllers that can actively retrieve and compose evidence through tool use and search procedures; creating domain-specific scaffolds including mathematical corpora, chart-to-code translation systems, and document retrieval pipelines that force pixel-grounded reasoning steps; and implementing sequence-aware training approaches through reinforcement learning and temporal encoding that reward correct multi-step reasoning under computational constraints.
Despite significant progress, critical challenges remain. Faithful diagram parsing and symbol grounding continue to limit mathematical reasoning capabilities. Temporally coherent video understanding under computational budgets remains largely unsolved. Most importantly, current evaluation approaches struggle to create assessments that resist language-only shortcuts while remaining practical for iterative development. Future advances will likely require continued innovation across all three strategic dimensions while developing more sophisticated evaluation frameworks that can guide progress toward genuine visual reasoning capabilities.

\begin{table}[htbp]
\centering
\small
\begin{tabularx}{\textwidth}{ >{\raggedright\arraybackslash}p{0.25\textwidth} >{\raggedright\arraybackslash\hsize=0.8\hsize\linewidth=\hsize}X >{\raggedright\arraybackslash\hsize=0.8\hsize\linewidth=\hsize}X >{\raggedright\arraybackslash\hsize=1.4\hsize\linewidth=\hsize}X }
\toprule
\textbf{Author} & \textbf{Datasets} & \textbf{Models} & \textbf{Evaluation Setup} \\
\midrule

\multicolumn{4}{l}{\textbf{Modular Controllers \& Active Retrieval}} \\

Stani\'c et al.~\citep{stanic2024towards} & RefCOCO, RefCOCO+, GQA, NExT-QA & Code-Bison (PaLM-2 based) with vision models & Visual grounding evaluated using IoU on RefCOCO/RefCOCO+, and compositional QA accuracy on GQA and NExT-QA\\

Dong et al.~\citep{dong2024progressive} & MATHVISTA, WE-MATH, GAOKAO-MM & GPT-4o, Qwen2-VL, InternVL2, LLaVA-OneVision & Multi-step QA; PRM vs ORM reasoning strategies \\

\midrule
\multicolumn{4}{l}{\textbf{Math \& Diagrammatic Reasoning}} \\

Zhang et al.~\citep{zhang2024mathverse} & MathVerse benchmark & Various VLMs incl. GPT-4V, Gemini &Zero-shot evaluation with Chain-of-Thought prompting on visual mathematical reasoning task \\

Shi et al.~\citep{shi2024math} & MathVista, Math-V, MathVerse, MMMU & Math-LLaVA (LLaVA-1.5-13B) & Zero-shot accuracy evaluation on multimodal mathematical reasoning tasks. \\

Qiao et al.~\citep{qiao2024we} & WE-MATH & GPT-4o, Qwen2-VL, InternVL & Multiple-choice accuracy evaluation on one-step, two-step, and three-step visual math problems \\

Han et al.~\citep{han2024infimm} & MathVerse, We-Math & InfiMM-Math & Chain-of-Thought prompting with answer accuracy evaluation \\

Zhang et al.~\citep{zhang2024humaneval} & HumanEval-V & 19 LLMs incl. GPT-4o, Claude 3.5, Gemini & pass@k, execution success rate for visual code reasoning \\

\midrule
\multicolumn{4}{l}{\textbf{Grounded Reasoning in Structured Visual Media}} \\

Carbune et al.~\citep{carbune2024chart} & ChartQA, FigureQA, PlotQA & ChartPaLI-5B (PaLI-3 based) & Reasoning accuracy (RA\%) evaluation with zero-shot, quick-adaptation, and fine-tuning settings. \\

Masry et al.~\citep{masry2024chartgemma} & ChartQA, ChartFC, ChartCheck, OpenCQA, Chart2Text & ChartGemma models & Relaxed accuracy for ChartQA, accuracy for ChartFC, and GPT-4–based evaluation (1–5 score) for open-ended tasks. \\

Yang et al.~\citep{yang2024chartmimic} & ChartMimic dataset & GPT-4o, Claude-3-Opus, Gemini-Pro-Vision & Direct Mimic and Customized Mimic tasks measuring chart understanding\\

Cho et al.~\citep{cho2024m3docrag} & M3DOCVQA, MP-DocVQA, MMLongBench-Doc & Qwen2-VL, Idefics 2/3, InternVL2 & EM, F1, ANLS, Recall@1 \\

Suri et al.~\citep{suri2024visdom} & PaperTab, MPNet-RAG, OCR-QA datasets & Qwen2-VL, ColQwen, Gemini & ANLCS, F1, visual RAG evaluation \\

\bottomrule
\end{tabularx}
\caption{Experimental setup overview of visual and multimodal reasoning papers (Part~1).}
\label{tab:vlm-summary-part1}
\end{table}

\begin{table}[htbp]
\centering
\small
\begin{tabularx}{\textwidth}{ >{\raggedright\arraybackslash}p{0.25\textwidth} >{\raggedright\arraybackslash\hsize=0.8\hsize\linewidth=\hsize}X >{\raggedright\arraybackslash\hsize=0.8\hsize\linewidth=\hsize}X >{\raggedright\arraybackslash\hsize=1.4\hsize\linewidth=\hsize}X }
\toprule
\textbf{Author} & \textbf{Datasets} & \textbf{Models} & \textbf{Evaluation Setup} \\
\midrule

\multicolumn{4}{l}{\textbf{Externalized Visual Reasoning \& Interpretability}} \\

Hu et al.~\citep{hu2024visual} & Geometry3K, IsoBench, V*Bench, MMVP, BLINK & GPT-4 Turbo, GPT-4o & Iterative Thought–Action–Observation reasoning with sketch generation\\

Zhang et al.~\citep{zhang2024large} & LLaVA-NeXT finetuning data & LLaVA-NeXT, GPT-4o & IoU, CLIP similarity for reasoning feature alignment \\

Liu et al.~\citep{liu2025visual} & Flower102, Pets37, COCO, LISA & Qwen2-VL (SFT, Visual-RFT) & Accuracy, mIoU comparisons \\

Liu et al.~\citep{liu2025othink} & GeoQA, SuperClevR & Qwen2-VL-Instruct & Few-shot RL reasoning accuracy \\

\midrule
\multicolumn{4}{l}{\textbf{Temporal Reasoning Across Long-Horizon Video}} \\

Shangguan et al.~\citep{shangguan2024tomato} & TOMATO dataset & GPT-4o, Gemini 1.5, Qwen2-VL & Temporal reasoning accuracy \\

Feng et al.~\citep{feng2025video} & VSI-Bench, MVBench, VideoMME & Qwen2.5-VL, Video-R1 & Accuracy with 64-frame inference \\

Jiang et al.~\citep{jiang2025token} & EgoSchema, LongVideoBench, VideoMME & STORM, GPT-4o, Qwen2-VL & Long video reasoning accuracy and latency \\

\midrule
\multicolumn{4}{l}{\textbf{Evaluation and Benchmarks}} \\

Hakimov et al.~\citep{hakimov2024two} & ADE20K, DOCCI, CLEVR, Visual Genome, Pentomino datasets & GPT-4o, GPT-4V, Claude-3.5-Sonnet, Gemini-1.5-Flash, & Multimodal game-based evaluation \\

Dihan et al.~\citep{dihan2024mapeval} & MapEval & GPT-4o, Claude 3.5, Gemini & Accuracy across textual, visual, API reasoning tasks \\

Cai et al.~\citep{cai2025mm} & MM-IQ benchmark & Qwen2.5-VL, Deepseek-VL, GPT-4o & Zero-shot accuracy across 8 reasoning paradigms \\

Tang et al.~\citep{tang2025lego} & LEGO-Puzzles, LEGO-Lite & MiniCPM, Qwen2.5-VL, GPT-4o & Spatial reasoning accuracy vs human baseline \\

Zhang et al.~\citep{zhang2024far} & Mensa, IT, RAVEN & GPT-4V, Gemini Pro, LLaVA & Visual IQ pattern reasoning accuracy \\

\bottomrule
\end{tabularx}
\caption{Experimental setup overview of visual and multimodal reasoning papers (Part~2).}
\label{tab:vlm-summary-part2}
\end{table}

\subsection{Temporal Reasoning}
\label{reason:VI}
\begin{figure}[htbp]
\centering
\definecolor{ink}{RGB}{38,41,54}
\definecolor{inksoft}{RGB}{92,97,115}
\definecolor{hair}{RGB}{200,205,216}
\definecolor{cBlue}{RGB}{60,87,150}
\definecolor{cTeal}{RGB}{46,125,111}
\definecolor{cAmber}{RGB}{168,120,43}
\definecolor{cViolet}{RGB}{103,87,140}
\definecolor{cGreen}{RGB}{92,136,85}
\scalebox{0.8}{
\begin{tikzpicture}[
 pstep/.style={rectangle, draw, line width=0.9pt, rounded corners=4pt, minimum height=1.5cm, minimum width=2.6cm, align=center, font=\small, text=ink},
 ex/.style={rectangle, draw=hair, line width=0.7pt, rounded corners=3pt, fill=ink!3, minimum height=0.7cm, align=center, font=\small, text width=3.2cm, inner sep=5pt, text=ink},
 det/.style={rectangle, draw=cGreen, line width=0.8pt, dashed, rounded corners=3pt, fill=cGreen!7, align=left, font=\small, text width=6.5cm, inner sep=7pt, text=ink},
 arr/.style={-{Stealth[length=2.4mm]}, draw=inksoft, line width=1pt},
 exarr/.style={-{Stealth[length=2mm]}, draw=hair, line width=0.9pt},
 link/.style={-{Stealth[length=1.6mm]}, draw=hair, line width=0.6pt, dashed},
]
\node[pstep, draw=cBlue, fill=cBlue!8] (s1) at (0,0) {\includegraphics[width=0.8cm]{user.PNG}\\\textbf{Input}\\\tiny Temporal\\\tiny question};
\node[pstep, draw=cTeal, fill=cTeal!8] (s2) at (3.8,0) {\includegraphics[width=0.8cm]{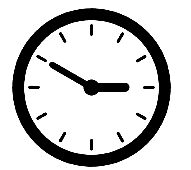}\\\textbf{Parse}\\\tiny Extract time\\\tiny expressions};
\node[pstep, draw=cAmber, fill=cAmber!8] (s3) at (7.6,0) {\includegraphics[width=0.8cm]{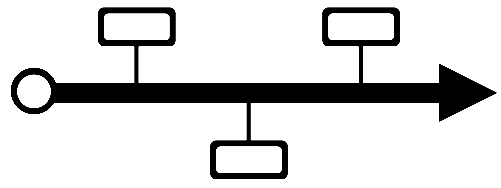}\\\textbf{Represent}\\\tiny Build timeline\\\tiny structure};
\node[pstep, draw=cViolet, fill=cViolet!8] (s4) at (11.4,0) {\includegraphics[width=0.8cm]{calculator_icon.PNG}\\\textbf{Compute}\\\tiny Time\\\tiny arithmetic};
\node[pstep, draw=cGreen, fill=cGreen!8] (s5) at (15.2,0) {\includegraphics[width=0.8cm]{check_icon.PNG}\\\textbf{Answer}\\\tiny Duration or\\\tiny ordering};
\draw[arr] (s1)--(s2); \draw[arr] (s2)--(s3); \draw[arr] (s3)--(s4); \draw[arr] (s4)--(s5);
\node[ex] (e1) at (0,-3.4) {``John started cooking\\at 5 PM and finished\\at 6:30 PM. How long?''};
\node[ex] (e2) at (3.8,-3.4) {Recognizes:\\- Start = 5 PM\\- End = 6:30 PM};
\node[ex] (e3) at (7.6,-3.4) {Converts to 24h:\\17:00 $\to$ 18:30\\Determines interval};
\node[ex] (e4) at (11.4,-3.4) {Time arithmetic:\\$18{:}30 - 17{:}00$\\$= 1$h $30$m};
\node[ex] (e5) at (15.2,-3.4) {\textbf{Answer:}\\``1 hour and\\30 minutes''};
\draw[exarr] (e1)--(e2); \draw[exarr] (e2)--(e3); \draw[exarr] (e3)--(e4); \draw[exarr] (e4)--(e5);
\draw[link] (s1)--(e1); \draw[link] (s5)--(e5);
\node[det] (d1) at (3.8,-6.4) {\textbf{Inside the Model (Temporal Repr.):}\\- Encodes temporal tokens in latent space\\- May simulate a mental timeline};
\node[det] (d2) at (11.4,-6.4) {\textbf{Inside the Model (Arithmetic):}\\- Uses positional embedding patterns to simulate time subtraction\\- Autoregressively generates reasoning chain};
\draw[link] (e2)--(d1); \draw[link] (e4)--(d2);
\end{tikzpicture}
}
\caption{Temporal reasoning flow. Row~1: generic pipeline. Row~2: temporal example showing time parsing, timeline construction, and duration computation. Row~3: internal mechanisms for temporal encoding and arithmetic.}
\label{fig:temporal-reasoning-flow}
\end{figure}

Temporal reasoning spans event ordering, durations, overlaps, and causal chains. It underpins timeline construction, multi-hop narratives, video understanding, and dialog memory. Figure~\ref{fig:temporal-reasoning-flow} shows a representative pipeline where time expressions are parsed, organized into a timeline, and resolved through temporal arithmetic. Recent work clusters into (i) evaluations that isolate temporal skills, (ii) structured representations (graphs/KGs) for explicit reasoning, (iii) explainable prediction pipelines, (iv) concurrency-aware benchmarks and unified training frameworks, and (v) cross-modal transfer and agent memory. We group the papers accordingly. The experimental details are provided in Table~\ref{tab:temporalreasoning-summary}.

\subsubsection{Benchmarks \& Audits of Temporal Skill}
This area focuses on disentangling true temporal understanding from dataset artifacts, offering more controlled and diagnostic evaluations of LLMs' temporal reasoning. ~\citep{fatemi2024test} introduce Test of Time, a synthetic benchmark that systematically varies structure, question type, and fact ordering to probe specific dimensions of temporal skill. It reveals where models succeed, struggle, or rely on shortcuts, and provides a framework for more targeted, deeper evaluations. ~\citep{chu2023timebench} propose TimeBench, a broad, hierarchical suite spanning phenomena such as event ordering, duration, and frequency. Their results highlight persistent gaps between LLMs and human reasoning, especially as task complexity increases.~\citep{wang2023tram} contribute Tram, a ten-dataset benchmark aggregating diverse temporal challenges. Despite covering a wide range of tasks, their results underscore that models still underperform humans by a significant margin across most temporal dimensions. 

\subsubsection{Structured Temporal Representations (Graphs \& KGs)}
Incorporating explicit temporal structure into language model reasoning enables better compositional generalization, transferability, and adaptability to evolving knowledge. Two recent approaches exemplify this trend by using structured representations, such as temporal graphs and temporal knowledge graphs (TKGs). TG-LLM by ~\citep{xiong2024large} introduces a two-step temporal reasoning framework in which natural language inputs are first translated into a temporal graph (TG), followed by structured reasoning over this graph. The model is trained using a synthetic, minimally supervised dataset (TGQA), with fine-tuning performed on a text-to-graph translation task. Temporal reasoning is further enhanced through Chain-of-Thought (CoT) bootstrapping and graph data augmentation, both of which improve the consistency and utility of intermediate reasoning steps. This structured pipeline yields significant improvements across temporal QA benchmarks by enabling the LLM to reason deliberately over graph-based temporal abstractions. LLM-DA, proposed by ~\citep{wang2024large}, targets dynamic temporal knowledge graph reasoning by leveraging LLMs to extract interpretable temporal logical rules from historical data and iteratively adapt them to new events, without fine-tuning the base model. The framework integrates a contextual relation selector to guide rule generation and introduces a dynamic adaptation strategy to maintain rule relevance as the underlying TKG evolves. This method not only enhances interpretability and adaptability but also achieves strong generalization on standard TKG reasoning tasks. These graph-centric approaches, whether through temporal graphs or TKGs, provide reusable, structured scaffolds for temporal reasoning. By encoding constraints and enabling rule-based inference, they help stabilize model behavior under distributional shift and facilitate continual learning with minimal supervision.

\subsubsection{Explainable Temporal Prediction}
This line of work aims to not only determine what will happen and when but also to provide clear explanations for why these outcomes are expected, often in the form of process traces. ~\citep{yuan2024back} present ExpTime, a framework that emphasizes interpretability in temporal forecasting. They introduce TimeLLaMA, a model trained on diverse temporal knowledge graph datasets that generates both future event predictions and accompanying explanations. By combining temporal reasoning with explanation generation, the model ensures that its outputs are aligned with human-understandable rationales. 

\subsubsection{Concurrency \& Unified Temporal Frameworks}
Understanding concurrent and overlapping events continues to challenge current language models, especially as temporal reasoning tasks grow more diverse and realistic. Standard benchmarks often focus on isolated events, which limits progress in more complex temporal scenarios. ~\citep{su2024living}introduce CoTempQA, a benchmark specifically designed to evaluate co-temporal reasoning through four types of overlapping relations. Despite improvements using Chain of Thought prompting, language models still struggle to reason over simultaneous events. Their analysis shows that these difficulties stem from deep mathematical constraints that current models fail to handle effectively. To address this, the same authors propose Timo, a unified framework that combines temporal reasoning with mathematical capabilities. Timo is trained across 38 distinct temporal tasks and uses a self-critic optimization loop to improve reasoning performance. This loop encourages the model to revise and refine its own outputs during training. The combination of mathematical pretraining and iterative self-correction enables the model to generalize across a wide range of temporal tasks without relying on task-specific adaptations. Effectively handling concurrency requires models to reason with explicit temporal structures and to integrate mathematical intuition with learning signals. Unified frameworks that incorporate both structured optimization and domain knowledge show promise in bridging this gap across a broad set of reasoning challenges.

\subsubsection{Cross-Modal Transfer \& Long-Horizon Memory}
Temporal reasoning must extend beyond text to modalities like video and persist reliably over extended interactions for effective agent use. ~\citep{li2024temporal} propose T3, a framework that uses text-based temporal reasoning datasets to improve temporal understanding in video language models. Rather than relying on video pretraining, T3 adapts temporal knowledge learned from text to enhance video QA and forecasting. The model achieves competitive results by mapping textual supervision into temporal patterns that generalize across modalities. ~\citep{ge2025tremu} introduce TReMu, a framework designed for multi-session dialogues that require long-range temporal tracking. It uses timeline summarization to create time-aware memory, allowing large language models to associate events with inferred dates and recover them in later sessions. To improve reasoning, TReMu incorporates a neuro-symbolic approach where the model generates and executes Python code to solve temporal queries. This pipeline boosts accuracy in benchmarks that require managing cross-session dependencies and relative time expressions. Transferring temporal skill from text to video provides an efficient path to multimodal competence. At the same time, neurosymbolic memory systems enhance reasoning over fragmented interactions by preserving and computing with temporal structure across sessions.

\subsubsection{Summary \& Open Questions}
Temporal reasoning improves when models are guided by two key principles. The first is structural representation through temporal graphs, knowledge graphs, and algebraic constraints, which help models organize and adapt their understanding over time. The second is process-level supervision using explanations and self-critique to ensure predictions align with observable temporal patterns.
Challenges remain in scaling to complex concurrent events, transferring temporal skill from text to high-dimensional video data without high cost, and defining standard metrics to evaluate the quality and faithfulness of temporal explanations.

\begin{table}[htbp]
\centering
\small
\begin{tabularx}{\textwidth}{ >{\raggedright\arraybackslash}p{0.25\textwidth} >{\raggedright\arraybackslash\hsize=0.8\hsize\linewidth=\hsize}X >{\raggedright\arraybackslash\hsize=0.8\hsize\linewidth=\hsize}X >{\raggedright\arraybackslash\hsize=1.4\hsize\linewidth=\hsize}X }
\toprule
\textbf{Author} & \textbf{Datasets} & \textbf{Models} & \textbf{Evaluation Setup} \\
\midrule

\multicolumn{4}{l}{\textbf{Benchmarks \& Audits of Temporal Skill}} \\

Fatemi et al.~\citep{fatemi2024test} & ToT (Semantic + Arithmetic) & Claude-3-Sonnet, GPT-4, GPT-4-Turbo, Gemini 1.5 Pro & Accuracy, precision, recall, and rank across multiple graph structures and temporal question types \\

Chu et al.~\citep{chu2023timebench} & TimeBench (10+ datasets incl. TRACIE, McTACO, MenatQA) & GPT-4, GPT-3.5, LLaMA2, Mistral, Baichuan2, Vicuna, ChatGLM, FLAN-T5 & Few/zero-shot w/ and w/o CoT; symbolic/commonsense/event TR; EM-based metrics; scaling analysis \\

Wang et al.~\citep{wang2023tram} & TRAM (10 temporal tasks) & GPT-4, GPT-3.5, Gemini, LLaMA2, BERT, RoBERTa, RST & Zero/5-shot SP + CoT; 300 samples/task; accuracy + F1; human expert upper bound \\

\midrule
\multicolumn{4}{l}{\textbf{Structured Temporal Representations (Graphs \& KGs)}} \\

Xiong et al.~\citep{xiong2024large} & TGQA (proposed), TimeQA, TempReason & LLaMA2 (7B/13B/70B), GPT-3.5, GPT-4, T5, Temp-T5, REMEMO & EM, token-F1, accuracy via ICL/SFT; CoT bootstrapping and graph data augmentation \\

Wang et al.~\citep{wang2024large} & ICEWS14, ICEWS05--15 & RE-NET, RE-GCN, TLogic, GPT-NeoX, LLaMA2-CoH, Vicuna, Mixtral, LLM-DA & Link prediction (MRR, Hits@K); dynamic ablations using historical/current/combined data \\

\midrule
\multicolumn{4}{l}{\textbf{Explainable Temporal Prediction}} \\

Yuan et al.~\citep{yuan2024back} & Gold TR dataset (w/ explanations) & Flan-T5, BART, MPT-7B, Falcon-7B, Vicuna, GPT-3.5, TimeLlama, ChatTimeLlama & Precision/recall/F1, BLEU/ROUGE/BERTScore; human evaluation of correctness and fluency \\

\midrule
\multicolumn{4}{l}{\textbf{Concurrency \& Unified Temporal Frameworks}} \\

Su et al.~\citep{su2024living} & Co-Temporal Reasoning Benchmark & GPT-4, GPT-3.5, LLaMA2, WizardMath/Coder & Closed-Book and Open-Book QA; Equal/Overlap/During/Mix categories with Accuracy and F1 \\

Ge et al.~\citep{ge2025tremu} & LoCoMo (multi-session dialogues) & GPT-4o, GPT-4o-mini, GPT-3.5 + MemoChat/Timeline/TReMu & Accuracy, F1, precision/recall on TA/TP/TI; symbolic CoT vs memory evaluation \\

\bottomrule
\end{tabularx}
\caption{Experimental setup overview of temporal reasoning papers.}
\label{tab:temporalreasoning-summary}
\end{table}

\subsection{Code/Algorithmic Reasoning}
\label{reason:VII}
\begin{figure}[htbp]
\centering
\definecolor{ink}{RGB}{38,41,54}
\definecolor{inksoft}{RGB}{92,97,115}
\definecolor{hair}{RGB}{200,205,216}
\definecolor{cBlue}{RGB}{60,87,150}
\definecolor{cTeal}{RGB}{46,125,111}
\definecolor{cAmber}{RGB}{168,120,43}
\definecolor{cViolet}{RGB}{103,87,140}
\definecolor{cGreen}{RGB}{92,136,85}
\scalebox{0.8}{
\begin{tikzpicture}[
 pstep/.style={rectangle, draw, line width=0.9pt, rounded corners=4pt, minimum height=1.5cm, minimum width=2.6cm, align=center, font=\small, text=ink},
 ex/.style={rectangle, draw=hair, line width=0.7pt, rounded corners=3pt, fill=ink!3, minimum height=0.7cm, align=center, font=\small, text width=3.2cm, inner sep=5pt, text=ink},
 det/.style={rectangle, draw=cGreen, line width=0.8pt, dashed, rounded corners=3pt, fill=cGreen!7, align=left, font=\small, text width=6.5cm, inner sep=7pt, text=ink},
 arr/.style={-{Stealth[length=2.4mm]}, draw=inksoft, line width=1pt},
 exarr/.style={-{Stealth[length=2mm]}, draw=hair, line width=0.9pt},
 link/.style={-{Stealth[length=1.6mm]}, draw=hair, line width=0.6pt, dashed},
]
\node[pstep, draw=cBlue, fill=cBlue!8] (s1) at (0,0) {\includegraphics[width=0.8cm]{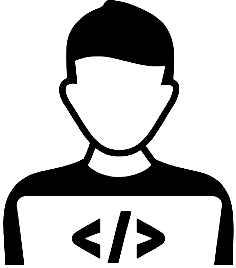}\\\textbf{Input}\\\tiny Coding task\\\tiny description};
\node[pstep, draw=cTeal, fill=cTeal!8] (s2) at (3.8,0) {\includegraphics[width=0.8cm]{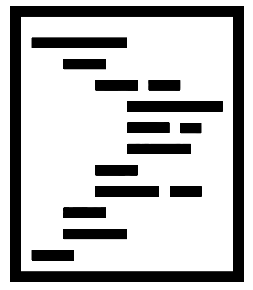}\\\textbf{Understand}\\\tiny Map intent\\\tiny to task type};
\node[pstep, draw=cAmber, fill=cAmber!8] (s3) at (7.6,0) {\includegraphics[width=0.8cm]{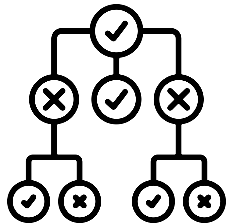}\\\textbf{Design}\\\tiny Plan algorithm\\\tiny and structure};
\node[pstep, draw=cViolet, fill=cViolet!8] (s4) at (11.4,0) {\includegraphics[width=0.8cm]{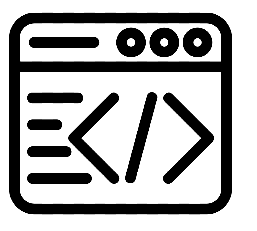}\\\textbf{Generate}\\\tiny Produce\\\tiny source code};
\node[pstep, draw=cGreen, fill=cGreen!8] (s5) at (15.2,0) {\includegraphics[width=0.8cm]{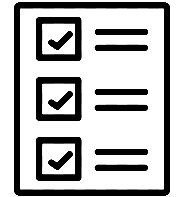}\\\textbf{Verify}\\\tiny Run tests\\\tiny and validate};
\draw[arr] (s1)--(s2); \draw[arr] (s2)--(s3); \draw[arr] (s3)--(s4); \draw[arr] (s4)--(s5);
\node[ex] (e1) at (0,-3.4) {``Write a function to\\check if a number\\is a palindrome.''};
\node[ex] (e2) at (3.8,-3.4) {Identifies task type:\\- Input = number\\- Output = True/False\\- Op: reverse-check};
\node[ex] (e3) at (7.6,-3.4) {Internally plans:\\- Convert to string\\- Reverse string\\- Compare original};
\node[ex] (e4) at (11.4,-3.4) {\ttfamily\footnotesize def palindrome(n):\\~~return str(n)\\~~== str(n)[::-1]};
\node[ex] (e5) at (15.2,-3.4) {\textbf{Passes all}\\test cases:\\Correct logic};
\draw[exarr] (e1)--(e2); \draw[exarr] (e2)--(e3); \draw[exarr] (e3)--(e4); \draw[exarr] (e4)--(e5);
\draw[link] (s1)--(e1); \draw[link] (s5)--(e5);
\node[det] (d1) at (3.8,-6.4) {\textbf{Inside the Model (Understanding):}\\- Maps keywords to common programming task templates\\- Activates relevant logic patterns from training corpus};
\node[det] (d2) at (11.4,-6.4) {\textbf{Inside the Model (Generation):}\\- Maintains intermediate logical plan across tokens\\- Follows common code scaffolding patterns};
\draw[link] (e2)--(d1); \draw[link] (e4)--(d2);
\end{tikzpicture}
}
\caption{Code reasoning flow. Row~1: generic pipeline. Row~2: code-specific example showing intent mapping, algorithm design, code generation, and test verification. Row~3: internal mechanisms.}
\label{fig:code-reasoning-flow}
\end{figure}
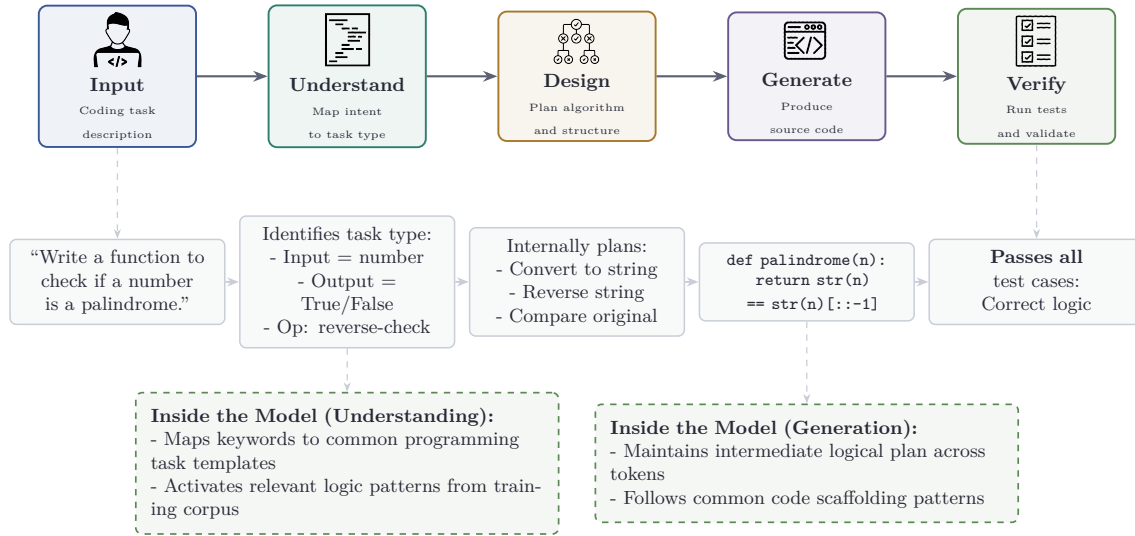

Code and algorithmic reasoning represent perhaps the most complex and rapidly evolving frontier in large language model capabilities, where models must master an intricate interplay of natural language understanding, formal logic, computational thinking, and real-world software engineering practices. Unlike mathematical reasoning, which operates within well-established formal systems, code reasoning must navigate the messy realities of software development: managing dependencies across massive codebases, debugging in complex execution environments, translating between multiple programming languages and abstraction levels, handling legacy systems, and reasoning about security vulnerabilities and performance optimizations. This multifaceted challenge requires models to function simultaneously as algorithm designers, code generators, debuggers, translators, and security analysts.
The emergence of code reasoning as a critical capability has been accelerated by the practical demands of modern software development, where LLMs are increasingly deployed as programming assistants, automated testing systems, code review tools, and even autonomous software agents. This real-world deployment pressure has driven rapid innovation across model architectures, training methodologies, evaluation frameworks, and safety mechanisms, making code reasoning one of the most actively researched areas in LLM development. Figure~\ref{fig:code-reasoning-flow} traces the typical pipeline from task description through intent mapping, algorithm design, code generation, and test verification. The following subsections trace how researchers have approached this multidimensional challenge through specialized strategies that collectively push toward truly capable computational reasoning systems. Tables~\ref{tab:code-reasoning-part1} and~\ref{tab:code-reasoning-part2} summarize the experimental setups.

\subsubsection{Benchmarks \& Audits of Code Reasoning}
Traditional benchmarks for code generation have largely relied on static pass@k metrics (e.g., HumanEval~\citep{zhang2024humaneval}), which conflate pattern memorization with genuine algorithmic reasoning. Recent work has moved toward more rigorous evaluations that distinguish between code generation, execution understanding, efficiency, and real-world usability.

 ~\citep{gu2024cruxeval} introduce \emph{CRUXEval}, a dataset of 800 short Python programs where models must predict input-output behavior rather than merely generate code. Their findings reveal that models performing well on HumanEval can still fail basic execution reasoning, with even GPT-4 and chain-of-thought prompting leaving notable comprehension gaps. Moving beyond static tasks, ~\citep{wei2025codearc} present \emph{CodeARC}, an interactive synthesis benchmark of 1,114 tasks with differential testing that evaluates iterative querying and self-correction. This framework highlights the mismatch between one-shot code generation and the debugging-oriented workflows required in practice.

Beyond correctness, other efforts focus on the efficiency and verifiability of generated code. EffiBench-X~\citep{qing2026effibench} evaluates runtime efficiency across six programming languages (Python, C++, Java, JavaScript, Ruby, Golang), exposing large performance disparities between human- and model-written solutions. Similarly, \emph{COFFE}~\citep{peng2025coffe} introduces the `efficient@k` metric based on CPU instruction counts, revealing that many LLMs produce unnecessarily slow code despite syntactic correctness. These efficiency-oriented audits reflect a growing concern that code reasoning systems should optimize computational performance, not just functional accuracy. 

A different dimension of rigor comes from verifiability. The \emph{VERINA}~\citep{ye2025verina} benchmark combines code, formal specifications, and Lean proofs, requiring models to produce not only correct implementations but also machine-checkable correctness guarantees. Results show that even state-of-the-art models achieve only 61.4\% correctness and a mere 3.6\% proof success rate, underscoring how far current systems are from generating formally trustworthy code.

Finally, researchers have begun grounding benchmarks in realistic developer workflows. Frameworks like \emph{SWE-Lancer}~\citep{miserendino2025swe}, which reframe real freelance software tasks as LLM challenges, expose economic and practical gaps in current systems. Complementary datasets such as \emph{BigCodeBench}~\citep{zhuo2024bigcodebench}, \emph{SWE-Bench}~\citep{jimenez2023swe}, \emph{KernelBench}~\citep{ouyang2025kernelbench}, and \emph{SpreadsheetBench}~\citep{ma2024spreadsheetbench} further test reasoning under real-world constraints like large codebases, kernel APIs, and domain-specific requirements.

Together, these evaluation efforts chart a progression: from testing execution comprehension (CRUXEval), to interactive debugging (CodeARC), to efficiency (EffiBench-X, COFFE), to verifiability (VERINA), and ultimately to real-world applicability (SWE-Lancer and others). This layered view of code reasoning benchmarks not only reveals persistent limitations but also provides clearer guidance for developing models that can function as genuine software engineering assistants.

\subsubsection{Programmatic Scaffolds and Algorithmic Abstraction}
Rather than treating each coding problem in isolation, recent approaches have explored how explicit algorithmic abstractions and reusable logical structures can enhance reasoning across diverse programming tasks. ~\citep{chae2024language} propose \emph{Think-and-Execute}, where models first generate shared pseudocode at the task level before simulating executions for individual instances. By separating reusable logic from input-specific details, the method achieves superior generalization compared to instance-only chain-of-thought or program-of-thought strategies. Building on the idea of structured reuse, ~\citep{li2025codei} reframe reasoning as \emph{code input–output prediction}, where test cases themselves serve as natural-language reasoning traces. Their CodeI/O++ framework uses multi-turn verification with input–output exemplars to substantially improve performance across symbolic, numeric, and commonsense tasks.
Other work extends scaffolding to broader data contexts. ~\citep{li2024open} introduce \emph{Open-Book Neural Algorithmic Reasoning}, which allows models to attend over entire training sets during inference on CLRS tasks. This ``open-book'' approach aggregates knowledge across exemplars, improving multi-task performance while offering interpretable insight into the reasoning process. In a complementary direction, ~\citep{lu2024mathcoder2} bridge symbolic mathematics and computation by systematically translating LaTeX reasoning steps into executable programs, creating the MathCode-Pile dataset. Continued pretraining on this coupled text–code representation yields MathCoder2, which demonstrates significantly stronger mathematical reasoning by unifying symbolic abstraction with program execution.
The success of these approaches demonstrates that externalizing algorithmic logic through pseudocode, test cases, exemplars, and executable mathematical representations provides reusable and verifiable scaffolds that enhance generalization across algorithmic reasoning tasks.

\subsubsection{Foundation Models and Training Paradigms}
Strong code reasoning hinges on foundation models designed with intent, not just larger-scale pretraining, but architecture, context length, and training objectives tailored to software intelligence. Recent developments reveal how advances in dataset scale, architectural design, and instruction tuning coalesce to produce models that rival closed systems in both performance and utility.

DeepSeek-Coder~\citep{guo2024deepseek} stands at the forefront of this movement. Built from scratch on 2 trillion tokens, 87 \% code and 13 \% English/Chinese annotations, it combines project-level context windows (initially 16K tokens) with a fill-in-the-blank objective to train models from 1.3B to 33B parameters. Despite being open-source, DeepSeek-Coder-Base-33B(which adds Grouped-Query Attention) outperforms contemporaries like CodeLlama-34B, and its instruction-tuned variants even match or exceed GPT-3.5-Turbo across HumanEval and MBPP benchmarks. The 7B base model is also competitive with substantially larger open models.
The leap to \emph{DeepSeek-Coder-V2}~\citep{zhu2024deepseek} refines this paradigm with a Mixture-of-Experts architecture that activates only subsets of weights per task (2.4B or 21B,  respectively), supports up to 338 languages, and extends context to 128K tokens. This model not only surpasses GPT-4-Turbo on HumanEval, MBPP+, and GSM8K but also achieves leading accuracy on real-world benchmarks like Aider and LiveCodeBench, illustrating the power of selective scale and long-context reasoning.

Beyond DeepSeek, others are building code-centric foundations. OctoPack~\citep{muennighoff2023octopack} (via StarCoder instruction tuning on CommitPack) underscores the impact of project-level code and intent preservation in multilingual code generation. Meanwhile, GLM-4.5~\citep{zeng2025glm} showcases how mixing agentic capabilities with deep training across 23 trillion tokens and with reinforcement fine-tuning can deliver strong performance on reasoning and coding benchmarks like SWE-bench, AIME, and TAU-Bench, even with fewer active parameters.
Open-code foundations like DeepSeek outperform earlier closed models by virtue of targeted corpora and task-aware objectives; MoE architectures, long-context capacity, and multilingual scale further elevate reasoning in diverse programming contexts; and agentic, reinforcement-enhanced training methods promise interactive, adaptive problem-solving capabilities. Far from mere replication of general-purpose LLMs, code foundations now lead in enabling models that reason, compile, and evolve like true software collaborators.

\subsubsection{Long Context \& Low-Level Code Representations}
Programming tasks often require reasoning about extended code contexts and low-level system representations, presenting unique challenges for memory management and specialized knowledge representation. ~\citep{guo2023longcoder} address the first challenge with \emph{LongCoder}, which combines sparse sliding-window attention with specialized bridge and memory tokens to efficiently propagate information across distant regions of code. This enables coherent completion and reasoning even in repository-scale contexts where conventional transformers struggle. At the other end of the abstraction spectrum, ~\citep{learningnova} target the second challenge with \emph{Nova}, a model for assembly-level reasoning. By using hierarchical attention to capture structural patterns in binaries and contrastive objectives to manage compilation-optimization variance, Nova substantially improves decompilation and binary similarity analysis tasks that lie beyond the reach of standard language models.
These approaches demonstrate that structure-aware attention mechanisms and specialized memory systems can effectively maintain global program state during reasoning, while domain-specific objectives unlock reasoning capabilities over assembly code and binary representations that general language training cannot achieve.

\subsubsection{Agents \& RL for End-to-End Programming}
Moving beyond static code generation, recent work has explored how agent-based approaches and reinforcement learning can enable iterative, goal-directed programming that more closely mirrors human development processes. ~\citep{grosnit2024large} introduce Agent K, an agentic framework that combines structured reasoning with both long-term and short-term memory systems. Instead of relying on gradient-based fine-tuning, the system accumulates experience through an experiential learning pipeline, gradually improving its problem-solving skills across diverse tasks. Remarkably, Agent K achieves Kaggle Grandmaster-level performance, showing that persistent memory and self-directed exploration can yield expert-level programming capabilities without continuous retraining.
Beyond competition platforms, similar ideas have been tested in software engineering settings. ~\citep{yang2024swe} extend SWE-bench into the multimodal domain and demonstrate that agent-based strategies, such as SWE-agent, outperform static generation approaches when tackling real-world GitHub issues. By explicitly structuring reasoning as plan–execute–verify cycles over repositories and test suites, these systems close part of the gap between synthetic benchmarks and practical development workflows.
In competitive programming, ~\citep{el2025competitive} show that a general-purpose RL-trained model (o3) surpasses a domain-specialized fine-tuned system (o1-ioi). This result underscores that broad reinforcement learning objectives, combined with scale, may be more effective than handcrafted domain-specific strategies for achieving robust performance across diverse algorithmic challenges.
Reinforcement learning has also been extended with tool-augmented agents: ~\citep{liu2025visual} and ~\citep{liu2025othink} propose reinforcement fine-tuning with verifiable rewards in multimodal reasoning contexts, but their principles, rewarding correctness through external feedback, map naturally to coding agents that can self-check via compilers, test cases, or static analyzers. These advances suggest that external feedback loops, coupled with RL, provide the backbone for scaling coding agents beyond imitation learning.
These findings highlight a turning point: by treating code generation as an agentic, iterative process, driven by memory, external feedback, and RL-based optimization, models can move beyond prompt-level tricks and toward autonomous planning–execution–verification loops. Such systems are better aligned with the demands of real-world programming, where debugging, refactoring, and strategy adaptation are as essential as initial code synthesis.

\subsubsection{Verification, Security, and Code Translation}
Trustworthy code generation requires mechanisms for functional verification, vulnerability detection, and reliable cross-language translation. Table~\ref{tab:code-verification-security} organizes recent systems by their primary task.

\begin{table}[htbp]
\centering
\small
\begin{tabularx}{\textwidth}{ >{\raggedright\arraybackslash}p{0.15\textwidth} >{\raggedright\arraybackslash}p{0.12\textwidth} >{\raggedright\arraybackslash}X >{\raggedright\arraybackslash}p{0.2\textwidth} }
\toprule
\textbf{System} & \textbf{Task} & \textbf{Approach} & \textbf{Key result} \\
\midrule
VERT \citep{yang2024vert} & Transpilation verification & Formal verification of Rust transpilation against WebAssembly oracles; refines until equivalence proofs hold & Verified functional equivalence for C-to-Rust translation \\
\addlinespace
SLFHunter \citep{ye2024detecting} & Vulnerability detection & Reasons about external library functions to find flaws taint analysis misses, including CI workflow issues & Discovers vulnerabilities invisible to traditional analysis \\
\addlinespace
Binary taint analysis \citep{liu2023harnessing} & Vulnerability detection & LLMs replace handcrafted taint-analysis rules for binary code & Finds previously unknown vulnerabilities \\
\addlinespace
GRACE \citep{lu2024grace} & Vulnerability detection & Fuses code property graphs with in-context retrieval & Stronger detection across diverse datasets \\
\addlinespace
DeGPT \citep{hu2024degpt} & Decompilation & Orchestrates referee, advisor, and operator roles to iteratively refine decompiler output & Improved semantic fidelity in reverse engineering \\
\addlinespace
Refining Decompiled C \citep{wong2023refining} & Decompilation & Focuses on recompilability of recovered source code & Recovery success rates above 75\% \\
\addlinespace
CodeRosetta \citep{tehrani2024coderosetta} & HPC translation & Parallelism-aware pretraining for C-to-CUDA translation & Outperforms general-purpose LLMs on performance portability \\
\bottomrule
\end{tabularx}
\caption{Verification, security, and code translation systems. Formal verification, structured program analysis, and domain-specific pretraining enable auditable outputs and reliable cross-language translation.}
\label{tab:code-verification-security}
\end{table}
\subsubsection{Integration and Future Directions}
 ~\citep{jin2025towards} provide a systematic framework for understanding LLM-based code generation through layered phases encompassing Input Processing, Orchestration, Development, and Validation. Their analysis identifies reliability, evaluation methodology, and deployment practices as core bottlenecks while mapping practical research directions for the field.
Code and algorithmic reasoning research has converged on three fundamental strategies: making reasoning processes explicit through programmatic scaffolds, test cases, and executable representations; rewarding complete reasoning processes through interactive benchmarks, reinforcement learning, and formal verification rather than focusing solely on final outputs; and adapting model architectures and training objectives to fit the specific demands of programming tasks, including long-context reasoning, low-level representations, and structured program analysis.
Despite significant progress, critical challenges remain in developing scalable post-hoc guarantees for general code generation, enabling faithful long-horizon planning under computational constraints, and establishing standardized interactive evaluations that correlate with real-world programming reliability. Future advances will likely require continued innovation across all three strategic dimensions while developing more sophisticated frameworks that can ensure both capability and trustworthiness in AI-assisted programming.

\begin{table}[htbp]
\centering
\small
\begin{tabularx}{\textwidth}{ >{\raggedright\arraybackslash}p{0.2\textwidth} >{\raggedright\arraybackslash\hsize=0.85\hsize\linewidth=\hsize}X >{\raggedright\arraybackslash\hsize=0.85\hsize\linewidth=\hsize}X >{\raggedright\arraybackslash\hsize=1.3\hsize\linewidth=\hsize}X }
\toprule
\textbf{Author} & \textbf{Datasets} & \textbf{Models} & \textbf{Evaluation Setup} \\
\midrule

\multicolumn{4}{l}{\textbf{Benchmarks \& Audits of Code Reasoning}} \\

Gu et al.~\citep{gu2024cruxeval} & CRUXEval (~1700 Python functions) & CodeLlama, GPT-4, StarCoder, others & pass@1/pass@5, bootstrap CI, CoT comparison \\

Wei et al.~\citep{wei2025codearc} & HumanEval+, MBPP+, APPS & GPT-4o, LLaMA-3.1 & Oracle interaction protocol, pass@1 \\

Qing et al.~\citep{qing2026effibench} & EFFIBENCH-X & DeepSeek-R1, DeepSeek-V3-0324, Llama-4 (Scout, Maverick) & Sandboxed execution with 100 test cases per problem \\

Peng et al.~\citep{peng2025coffe} & COFFE & Phi-3, CodeLlama, StarCoder, Mixtral & Stressful test case generation using STGen \\

Ye et al.~\citep{ye2025verina} & VERINA & GPT-4o-mini, DeepSeek-V3, GPT-4o, Qwen-3-235B-A22B & 2-shot prompting with pass@1 and pass@k metrics evaluating Code Generation \\

Miserendino et al.~\citep{miserendino2025swe} & SWE-Lancer (IC SWE and SWE Manager tasks) & GPT-4o-2024-08-06, o1, Claude-3.5-Sonnet, DeepSeek-R1, Llama-3.3 & Agents run in a Docker sandbox without internet \\

Zhuo et al.~\citep{zhuo2024bigcodebench} & BigCodeBench & 60 LLMs including GPT-4o, GPT-4-Turbo, GPT-4 & Zero-shot code generation using greedy decoding; evaluated with Pass@1 and Pass@5 \\

Jimenez et al.~\citep{jimenez2023swe} & SWE-bench (and SWE-bench Lite) &ChatGPT-3.5, GPT-4, GPT-4-Turbo, Claude 2 &performance measured using \% Resolved and \% Apply based on generated patches \\

Ouyang et al.~\citep{ouyang2025kernelbench} & KernelBench & Llama-3.1-70B, DeepSeek-V3, DeepSeek-R1, OpenAI o1 & GPU kernel generation evaluated using fastp@k / fast1 metrics \\

Ma et al.~\citep{ma2024spreadsheetbench} & SPREADSHEETBENCH & TaPEx, TaPas, Binder (GPT-3.5), CodeQwen-7B, DeepSeekCoder-33B & Spreadsheet manipulation tasks evaluated using OJ-style testing with multiple spreadsheet test cases per instruction \\

\midrule
\multicolumn{4}{l}{\textbf{Programmatic Scaffolds and Algorithmic Abstraction}} \\

Chae et al.~\citep{chae2024language} & 7 BBH tasks & GPT-3.5, CodeLlama & Zero-shot, CoT, PoT evaluation \\

Li et al.~\citep{li2025codei} & DROP, GSM8K, MMLU-STEM & Qwen2.5, LLaMA3.1 & Two-stage instruction tuning evaluation \\

Li et al.~\citep{li2024open} & CLRS benchmark (30 tasks) & PGN, MPNN, Triplet-GMPNN & F1 score over algorithmic tasks \\

Lu et al.~\citep{lu2024mathcoder2} & GSM8K, MATH, SAT-Math, MMLU-Math & MathCoder2, DeepSeekMath & 4-shot and zero-shot evaluation \\

\midrule
\multicolumn{4}{l}{\textbf{Foundation Models and Training Paradigms}} \\

Guo et al.~\citep{guo2024deepseek} & HumanEval, MBPP & DeepSeekCoder, GPT-4 & Zero-shot and few-shot evaluation \\

Zhu et al.~\citep{zhu2024deepseek} &HumanEval, MBPP+, LiveCodeBench & CodeLlama (7B–70B), StarCoder, StarCoder2 & Evaluated across code generation, code completion, code fixing, and code reasoning tasks \\

Muennighoff et al.~\citep{muennighoff2023octopack} & COMMITPACK, HUMANEVALPACK & OctoPack, GPT-4 & pass@1 multilingual evaluation \\

Zeng et al.~\citep{zeng2025glm} & SimpleQA, BBH, MMLU & GLM-4.5, GLM-4.5-Air, Qwen3-235B & Evaluated across agentic abilities, reasoning, coding, and multilingual knowledge tasks \\

\bottomrule
\end{tabularx}
\caption{Experimental setup overview of code/algorithmic reasoning papers (Part~1).}
\label{tab:code-reasoning-part1}
\end{table}

\begin{table}[htbp]
\centering
\small
\begin{tabularx}{\textwidth}{ >{\raggedright\arraybackslash}p{0.25\textwidth} >{\raggedright\arraybackslash\hsize=0.8\hsize\linewidth=\hsize}X >{\raggedright\arraybackslash\hsize=0.8\hsize\linewidth=\hsize}X >{\raggedright\arraybackslash\hsize=1.4\hsize\linewidth=\hsize}X }
\toprule
\textbf{Author} & \textbf{Datasets} & \textbf{Models} & \textbf{Evaluation Setup} \\
\midrule

\multicolumn{4}{l}{\textbf{Long Context \& Low-Level Code Representations}} \\

Guo et al.~\citep{guo2023longcoder} & LCC, PY150, JavaCorpus & LongCoder, UniXcoder & Exact match, edit similarity \\

Jiang et al.~\citep{learningnova} & AnghaBench, BinaryCorp-3M & Nova, GPT-4o, CodeLlama & Binary similarity and decompilation evaluation \\

\midrule
\multicolumn{4}{l}{\textbf{Agents \& RL for End-to-End Programming}} \\

Grosnit et al.~\citep{grosnit2024large} & Kaggle competitions & Agent K & Leaderboard Elo-MMR evaluation \\

Yang et al.~\citep{yang2024swe} & SWE-bench M & GPT-4o (gpt-4o-2024-08-06), Claude 3.5 Sonnet (claude-3-5-sonnet-20240620) & Evaluated using RAG, SWE-agent (Base, JS, Multimodal), and Agentless JS frameworks \\

El-Kishky et al.~\citep{el2025competitive} & IOI 2024 tasks & o1, o3 models & Live competition evaluation \\

Liu et al.~\citep{liu2025visual} & Flower102, Pets37, FGVC-Aircraft & Qwen2-VL-2B, Qwen2-VL-7B & Evaluated on few-shot visual perception tasks, including fine-grained image classification, object detection \\

Liu et al.~\citep{liu2025othink} & GeoQA-Train (8K), GeoQA-Test (753), R1 Distilled CLEVR (37K), SuperCLEVR & Qwen2-VL-2B-Instruct, Qwen2-VL-7B-Instruct & Accuracy on visual counting and geometry reasoning tasks, along with cross-task generalization ability \\

\midrule
\multicolumn{4}{l}{\textbf{Verification, Security, and Code Translation}} \\

Yang et al.~\citep{yang2024vert} & TransCoder-IR, CROWN & CodeLlama, StarCoder & Verified Rust transpilation \\

Ye et al.~\citep{ye2024detecting} & Embedded Linux firmware samples & GPT-4 + analysis tools & Vulnerability detection evaluation \\

Liu et al.~\citep{liu2023harnessing} & Juliet v1.3 dataset & GPT-4, LATTE & Precision, Recall, F1 metrics \\

Lu et al.~\citep{lu2024grace} & Devign, Reveal datasets & GRACE (LLM + GNN) & Binary classification F1 \\

Hu et al.~\citep{hu2024degpt} & Decompiled program datasets & DeGPT & Decompilation readability metrics \\

Wong et al.~\citep{wong2023refining} & CodeContest dataset & DecGPT & Chain length to successful recompilation \\

TehraniJamsaz et al.~\citep{tehrani2024coderosetta} & BabelTower, Stack V2 & CodeRosetta, GPT-4 & BLEU, CodeBLEU, compilation accuracy \\

\bottomrule
\end{tabularx}

\caption{Experimental setup overview of code/algorithmic reasoning papers (Part~2).}
\label{tab:code-reasoning-part2}
\end{table}

\subsection{Retrieval-Augmented Generation (RAG)-based Reasoning}
\label{reason:VIII}
\begin{figure}[htbp]
\centering
\definecolor{ink}{RGB}{38,41,54}
\definecolor{inksoft}{RGB}{92,97,115}
\definecolor{hair}{RGB}{200,205,216}
\definecolor{cBlue}{RGB}{60,87,150}
\definecolor{cTeal}{RGB}{46,125,111}
\definecolor{cAmber}{RGB}{168,120,43}
\definecolor{cViolet}{RGB}{103,87,140}
\definecolor{cGreen}{RGB}{92,136,85}
\scalebox{0.8}{
\begin{tikzpicture}[
 pstep/.style={rectangle, draw, line width=0.9pt, rounded corners=4pt, minimum height=1.5cm, minimum width=2.6cm, align=center, font=\small, text=ink},
 ex/.style={rectangle, draw=hair, line width=0.7pt, rounded corners=3pt, fill=ink!3, minimum height=0.7cm, align=center, font=\small, text width=3.2cm, inner sep=5pt, text=ink},
 det/.style={rectangle, draw=cGreen, line width=0.8pt, dashed, rounded corners=3pt, fill=cGreen!7, align=left, font=\small, text width=6.5cm, inner sep=7pt, text=ink},
 arr/.style={-{Stealth[length=2.4mm]}, draw=inksoft, line width=1pt},
 exarr/.style={-{Stealth[length=2mm]}, draw=hair, line width=0.9pt},
 link/.style={-{Stealth[length=1.6mm]}, draw=hair, line width=0.6pt, dashed},
]
\node[pstep, draw=cBlue, fill=cBlue!8] (s1) at (0,0) {\includegraphics[width=0.8cm]{user.PNG}\\\textbf{Query}\\\tiny User asks a\\\tiny factual question};
\node[pstep, draw=cTeal, fill=cTeal!8] (s2) at (3.8,0) {\includegraphics[width=0.8cm]{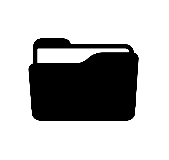}\\\textbf{Retrieve}\\\tiny Search vector\\\tiny database};
\node[pstep, draw=cAmber, fill=cAmber!8] (s3) at (7.6,0) {\includegraphics[width=0.8cm]{lightbulb_icon.PNG}\\\textbf{Rerank}\\\tiny Score and filter\\\tiny passages};
\node[pstep, draw=cViolet, fill=cViolet!8] (s4) at (11.4,0) {\includegraphics[width=0.8cm]{fusion.PNG}\\\textbf{Fuse}\\\tiny Combine docs\\\tiny with prompt};
\node[pstep, draw=cGreen, fill=cGreen!8] (s5) at (15.2,0) {\includegraphics[width=0.8cm]{check_icon.PNG}\\\textbf{Generate}\\\tiny Evidence-based\\\tiny answer};
\draw[arr] (s1)--(s2); \draw[arr] (s2)--(s3); \draw[arr] (s3)--(s4); \draw[arr] (s4)--(s5);
\node[ex] (e1) at (0,-3.4) {``What are the health\\benefits of turmeric\\according to studies?''};
\node[ex] (e2) at (3.8,-3.4) {Encodes query into\\vector; sends to\\retriever (e.g., FAISS)};
\node[ex] (e3) at (7.6,-3.4) {Filters top-$k$\\relevant passages;\\scores by relevance};
\node[ex] (e4) at (11.4,-3.4) {Combines [Retrieved\\Docs] + Original\\Prompt $\to$ LLM input};
\node[ex] (e5) at (15.2,-3.4) {\textbf{Output:}\\Evidence-based\\grounded answer};
\draw[exarr] (e1)--(e2); \draw[exarr] (e2)--(e3); \draw[exarr] (e3)--(e4); \draw[exarr] (e4)--(e5);
\draw[link] (s1)--(e1); \draw[link] (s5)--(e5);
\node[det] (d1) at (3.8,-6.4) {\textbf{Inside the Model (Reranking):}\\- Reranks using cross-attention or similarity scores\\- Narrows down hallucination risk};
\node[det] (d2) at (11.4,-6.4) {\textbf{Inside the Model (Fusion):}\\- Decoder uses retrieved facts + language priors\\- Balances factual relevance with fluent response};
\draw[link] (e3)--(d1); \draw[link] (e4)--(d2);
\end{tikzpicture}
}
\caption{Retrieval-augmented reasoning flow. Row~1: generic RAG pipeline. Row~2: example showing query embedding, document retrieval, reranking, prompt fusion, and evidence-grounded generation. Row~3: internal mechanisms.}
\label{fig:rag-reasoning-flow}
\end{figure}

Retrieval-Augmented Generation (RAG) extends the reasoning capabilities of LLMs by integrating external knowledge retrieval into the generation process. This paradigm has proven especially effective for reasoning tasks that require factual accuracy, up-to-date information, and grounding in long or multi-document contexts. As reasoning often involves synthesizing evidence, handling complex queries, or navigating multi-hop chains of logic, RAG serves as a natural fit for augmenting LLMs' capabilities in these areas. Figure~\ref{fig:rag-reasoning-flow} illustrates the end-to-end RAG pipeline, from query encoding and document retrieval through reranking and prompt fusion to evidence-grounded generation. In this survey, we focus on how RAG methods have been leveraged and extended for various forms of reasoning. Tables~\ref{tab:rag-reasoning-part1} and~\ref{tab:rag-reasoning-part2} summarize the experimental setups.

\subsubsection{Long-Context, Memory \& Efficiency}
Reasoning over long contexts requires models to use available information efficiently, produce extended outputs, and manage computational cost. Five recent systems address different bottlenecks:

\begin{itemize}
\item \textbf{Context utilization.} BABILong~\citep{kuratov2024babilong} benchmarks 20 reasoning tasks in ultra-long documents, finding that standard LLMs use only 10--20\% of available context. Fine-tuned recurrent memory transformers process up to 50M tokens effectively.

\item \textbf{Window extension.} ChatQA 2~\citep{xu2024chatqa} extends LLaMA 3 to 128K tokens through continued pretraining and three-stage instruction tuning, showing that retrieval scale and context length synergize to improve reasoning accuracy.

\item \textbf{Memory architecture.} MemoRAG~\citep{qian2024memorag} adopts a dual-system design: a lightweight module constructs global memory with key-value compression, while a generator uses feedback-driven clues to retrieve and reason over relevant content, improving performance on unstructured and query-implicit tasks.

\item \textbf{Output length.} LongWriter/AgentWrite~\citep{bai2024longwriter} identifies that generation lengths are bounded by fine-tuning data. Creating training data with 10K--20K word outputs enables extended generation with high coherence and factuality.

\item \textbf{Inference cost.} RetrievalAttention~\citep{liu2024retrievalattention} indexes key-value caches via attention-aware vector search, retrieving only the most relevant tokens per generation step. This supports 128K-token inference with near full-attention accuracy and up to 5$\times$ latency reduction.
\end{itemize}

\subsubsection{Unified Retrieval--Generation Architectures}
Unified architectures integrate retrieval and generation within a single model, removing the latency and fragmentation of multi-stage pipelines. Table~\ref{tab:unified-rag-architectures} compares four approaches.

\begin{table}[htbp]
\centering
\small
\begin{tabularx}{\textwidth}{ >{\raggedright\arraybackslash}p{0.15\textwidth} >{\raggedright\arraybackslash}p{0.2\textwidth} >{\raggedright\arraybackslash}X >{\raggedright\arraybackslash}p{0.2\textwidth} }
\toprule
\textbf{System} & \textbf{Unification strategy} & \textbf{How it works} & \textbf{Key advantage} \\
\midrule
ReLiK \citep{orlando2024relik} & Structural: fused retriever-reader & Merges entity linking and QA into a single-pass model & Up to 40$\times$ inference speedup \\
\addlinespace
OneGen \citep{zhang2024onegen} & Structural: autoregressive retrieval tokens & Retrieval tokens emitted during generation in one forward pass & Simultaneous retrieval and generation; outperforms GRIT on multi-hop \\
\addlinespace
Promptriever \citep{weller2024promptriever} & Behavioral: instruction-following retriever & Trains a retriever to follow prompts like an LM, generalizing across tasks and unseen instructions & Strong few-shot and zero-shot retrieval \\
\addlinespace
RankRAG \citep{yu2024rankrag} & Structural: joint ranking and answering & Single instruction-tuned LLM ranks passages and generates answers & Outperforms task-specific rankers on knowledge-intensive QA \\
\bottomrule
\end{tabularx}
\caption{Unified retrieval-generation architectures. Whether through structural fusion or behavioral alignment, these systems eliminate pipeline fragmentation and enable end-to-end reasoning over external knowledge.}
\label{tab:unified-rag-architectures}
\end{table}

\subsubsection{Planning \& Multi-hop Chain-of-Retrieval}

Recent RAG advances treat retrieval as an iterative, procedural process integrated with reasoning. Chain-of-Retrieval (CoRAG)~\citep{wang2025chain} trains a model to iteratively retrieve evidence through reformulated sub-queries, guided by rejection sampling. It allows test-time control over retrieval steps and decoding strategies, achieving over 10-point EM gains on MuSiQue. DeepRAG~\citep{guan2025deeprag} frames RAG as a Markov Decision Process where the model learns policies for when to use parametric memory versus external retrieval, improving reasoning fidelity by over 20\% on complex tasks. ReaRAG, proposed by ~\citep{lee2025rearag}, introduces a lightweight retrieval planner that alternates between `Search' and `Finish' actions, strengthening factual grounding in large retrieval-augmented models. These works collectively transform RAG into a procedural reasoning engine that bridges symbolic-like control with neural generation.

\subsubsection{Document-Centric \& Multimodal RAG}
Reasoning over structured or visually rich documents suffers when content is flattened to plain text. Recent work preserves layout, visual signals, and structural fidelity:

\begin{itemize}
\item \textbf{Image-based retrieval.} VisRAG~\citep{yu2024visrag} treats document pages as images and embeds visual and spatial information via VLMs, achieving up to 40\% QA gains over text-only pipelines.

\item \textbf{HTML structure preservation.} HtmlRAG~\citep{tan2025htmlrag} retains HTML tags and layout through cleaning and block-tree pruning, improving QA accuracy across six datasets.

\item \textbf{OCR noise benchmarking.} OHRBench~\citep{zhang2024ocr} quantifies how OCR artifacts (wrong words, layout distortions) degrade both retrieval relevance and generation faithfulness.

\item \textbf{Multimodal medical RAG.} MMed-RAG~\citep{xia2024mmed} integrates domain-aware retrieval, adaptive context selection, and instruction-tuned generation to reduce hallucinations in clinical document QA.
\end{itemize}

\subsubsection{Domain-Specific RAG}
RAG gains traction in domains requiring up-to-date grounding and interpretability. Table~\ref{tab:domain-specific-rag} summarizes four applications.

\begin{table}[htbp]
\centering
\small
\begin{tabularx}{\textwidth}{ >{\raggedright\arraybackslash}p{0.14\textwidth} >{\raggedright\arraybackslash}p{0.16\textwidth} >{\raggedright\arraybackslash}X >{\raggedright\arraybackslash}p{0.18\textwidth} }
\toprule
\textbf{Domain} & \textbf{System} & \textbf{Retrieval strategy} & \textbf{Key result} \\
\midrule
Fact-checking & Few-shot RAG \citep{singhal2024evidence} & Evidence-backed social media verification & 22\% accuracy improvement over baselines \\
\addlinespace
Medical QA & SearchRAG \citep{shi2025searchrag} & Query synthesis with uncertainty-based evidence selection & Higher quality real-time, fine-grained answers \\
\addlinespace
Finance & StockLLM + FinSeer \citep{xiao2025retrieval} & 1B LLM with LLM-guided retriever for historical sequences & Up to 8\% gain on stock prediction benchmarks \\
\addlinespace
Code-mixed QA & RetrieveGPT \citep{deroy2024retrievegpt} & LLM prompting with mathematical relevance modeling & Improved bilingual (Bengali-English) social media QA \\
\bottomrule
\end{tabularx}
\caption{Domain-specific RAG applications. Each domain benefits from adaptive retrieval strategies tailored to its grounding and reliability requirements.}
\label{tab:domain-specific-rag}
\end{table}

\subsubsection{Graphs \& Multi-Corpora, Multimodal RAG}
GraphRAG~\citep{peng2024graph} constructs knowledge graphs on top of documents, improving semantic retrieval for factual grounding and response accuracy through entity-relationship exploitation. UniversalRAG, proposed by ~\citep{yeo2025universalrag}, extends RAG to heterogeneous corpora spanning text, images, and videos at multiple granularities. Its modality and granularity-aware routing mechanism selects the most suitable corpus for each query type, avoiding retrieval noise from irrelevant modalities. UniversalRAG outperforms both modality-specific and unified baseline methods across eight benchmarks.

\subsubsection{Evaluations \& Frameworks}
RAG evaluation must capture retrieval quality, factual grounding, and multi-step inference. Table~\ref{tab:rag-eval-frameworks} compares five recent frameworks and tools.

\begin{table}[htbp]
\centering
\small
\begin{tabularx}{\textwidth}{ >{\raggedright\arraybackslash}p{0.16\textwidth} >{\raggedright\arraybackslash}p{0.14\textwidth} >{\raggedright\arraybackslash}X >{\raggedright\arraybackslash}p{0.2\textwidth} }
\toprule
\textbf{Framework} & \textbf{Scope} & \textbf{What it evaluates} & \textbf{Key contribution} \\
\midrule
FRAMES \citep{krishna2024fact} & Multi-hop QA & Factuality, retrieval, and reasoning jointly (824 questions) & Multi-hop pipelines outperform no-retrieval baselines by 0.66 vs. 0.408 accuracy \\
\addlinespace
CORAL \citep{cheng2024coral} & Conversational RAG & Passage retrieval, answer generation, and citation labeling & Fine-grained inspection of RAG behavior over discourse \\
\addlinespace
OmniEval \citep{wang2024omnieval} & Finance domain & Retrieval and generation with task-specific metrics & Domain-specific evaluation suite \\
\addlinespace
RAG Foundry \citep{fleischer2024rag} & General & Full-stack open-source experimentation platform & Consistent pipelines from raw data to metric computation \\
\addlinespace
RE-AdaptIR \citep{fleshman2024re} & Retriever adaptation & Reverse-engineering adaptation using unlabeled in-domain data & Scalable retriever improvement without annotation \\
\bottomrule
\end{tabularx}
\caption{RAG evaluation frameworks and tools. Multi-dimensional assessment across retrieval, reasoning, and factuality is needed for reliable deployment.}
\label{tab:rag-eval-frameworks}
\end{table}

\begin{figure}[htbp]
\centering
\definecolor{ink}{RGB}{38,41,54}
\definecolor{inksoft}{RGB}{92,97,115}
\definecolor{hair}{RGB}{200,205,216}
\definecolor{cBlue}{RGB}{60,87,150}
\definecolor{cTeal}{RGB}{46,125,111}
\definecolor{cAmber}{RGB}{168,120,43}
\definecolor{cViolet}{RGB}{103,87,140}
\definecolor{cGreen}{RGB}{92,136,85}
\definecolor{cSlate}{RGB}{85,92,112}
\begin{tikzpicture}[
  block/.style={rectangle, draw=#1, line width=0.9pt, rounded corners=4pt, fill=#1!8,
    minimum height=1.0cm, align=center, font=\scriptsize, text width=2.5cm, inner sep=4pt, text=ink},
  arr/.style={-{Stealth[length=2mm]}, draw=inksoft, line width=0.9pt},
  feedback/.style={-{Stealth[length=1.8mm]}, draw=hair, line width=0.8pt, dashed},
  elab/.style={font=\tiny, text=inksoft, fill=white, inner sep=1.5pt},
]
\node[block=cSlate]  (lc)   at (0,3)    {\textbf{Long-Context \& Memory}\\BABILong, MemoRAG,\\RetrievalAttention};
\node[block=cGreen]  (uni)  at (5.2,3)  {\textbf{Unified Ret-Gen}\\ReLiK, OneGen,\\RankRAG};
\node[block=cAmber]  (plan) at (10.4,3) {\textbf{Planning \&}\\\textbf{Multi-hop CoR}\\CoRAG, DeepRAG};
\node[block=cViolet] (doc)  at (0,0)    {\textbf{Document \&}\\\textbf{Multimodal}\\VisRAG, HtmlRAG};
\node[block=cTeal]   (dom)  at (5.2,0)  {\textbf{Domain-Specific}\\Fact-check, Medical,\\Finance};
\node[block=cBlue]   (eval) at (10.4,0) {\textbf{Evaluation}\\FRAMES, CORAL,\\OmniEval};
\draw[arr] (lc)   -- node[elab, above] {feeds context to} (uni);
\draw[arr] (uni)  -- node[elab, above] {enables iterative} (plan);
\draw[arr] (lc)   -- node[elab] {preserves structure} (doc);
\draw[arr] (doc)  -- node[elab, below] {grounds} (dom);
\draw[arr] (dom)  -- node[elab, below] {tested by} (eval);
\draw[arr] (plan.south) -- node[elab] {assessed by} (eval.north);
\draw[feedback] (eval.north west) -- node[elab, sloped] {exposes gaps in} (lc.south east);
\end{tikzpicture}
\caption{How RAG research directions connect. Long-context methods feed unified architectures, which enable iterative multi-hop retrieval. Document-centric approaches preserve structure for domain applications. Evaluation frameworks assess all components and expose gaps that loop back to long-context and planning research.}
\label{fig:rag-subsection-map}
\end{figure}
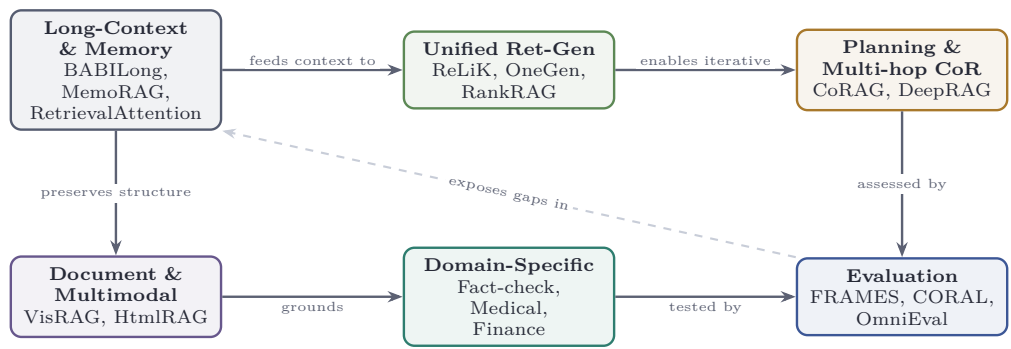

\noindent Figure~\ref{fig:rag-subsection-map} shows how these six research directions connect within the RAG ecosystem.

\subsubsection{Safety \& Robustness}
RAG pipelines, with their growing complexity, are exposed to adversarial attacks and denial-of-service-type failures. ~\citep{liang2025saferag} proposed the SafeRAG benchmark, which consists of various categories of attacks (e.g., silver noise, white DoS) to detect system-level weaknesses and stress-test the RAG system. This work thus points to the importance of evaluating the security of RAGs alongside their accuracy, emphasizing that RAG systems must incorporate defenses against retrieval layer attacks and noise injection.

\subsubsection{Summary \& Open Questions}
We have identified three levers that are employed to achieve robustness: (1) \emph{procedural retrieval} (chain/MDP/action loops) that plans evidence gathering; (2) \emph{structure preservation} (HTML/layout/graphs) and modality-aware routing to avoid lossy flattening; (3) \emph{efficiency + safety} (KV retrieval, memory, security stress tests) for practical and robust systems. The challenges still include the integration of planner-style retrieval with long-context memory, the development of principled defenses against retrieval attacks, and the establishment of standardized multi-turn metrics having high correlation with downstream reliability.

\begin{table}[htbp]
\centering
\small
\begin{tabularx}{\textwidth}{ >{\raggedright\arraybackslash}p{0.25\textwidth} >{\raggedright\arraybackslash\hsize=0.8\hsize\linewidth=\hsize}X >{\raggedright\arraybackslash\hsize=0.8\hsize\linewidth=\hsize}X >{\raggedright\arraybackslash\hsize=1.4\hsize\linewidth=\hsize}X }
\toprule
\textbf{Author} & \textbf{Datasets} & \textbf{Models} & \textbf{Evaluation Setup} \\
\midrule

\multicolumn{4}{l}{\textbf{Long-Context, Memory \& Efficiency}} \\

Kuratov et al.~\citep{kuratov2024babilong} & BABILong (QA tasks with PG19 distractors) & GPT-4, Gemini, LLaMA-3.1, Phi, Qwen & QA accuracy across increasing context sizes (up to millions of tokens) \\

Xu et al.~\citep{xu2024chatqa} & InfiniteBench, QMSum, HotpotQA, ChatRAG Bench & LLaMA3-ChatQA, GPT-4-Turbo, Qwen2 & 128K context RAG; metrics: ROUGE-L, F1, EM \\

Qian et al.~\citep{qian2024memorag} & NarrativeQA, GovReport, UltraDomain & MemoRAG (Mistral), LLaMA3 & Memory-guided retrieval; F1 across domains \\

Bai et al.~\citep{bai2024longwriter} & LongBench-Write & LongWriter models, GPT-4o & Output length compliance and quality metrics \\

Liu et al.~\citep{liu2024retrievalattention} & $\infty$-Bench, RULER, Needle-in-Haystack & RetrievalAttention (LLaMA-based) & KV-cache retrieval efficiency and latency \\

\midrule
\multicolumn{4}{l}{\textbf{Unified Retrieval–Generation Architectures}} \\

Orlando et al.~\citep{orlando2024relik} & AIDA-CoNLL, KILT datasets & ReLiK models & Entity linking micro-F1 via GERBIL \\

Zhang et al.~\citep{zhang2024onegen} & LongBench-Write & GPT-4o, LLaMA-3.1, GLM-4 & Human preference win-rate and quality metrics \\

Weller et al.~\citep{weller2024promptriever} & MS MARCO, BEIR & Promptriever, RepLLaMA & MRR, nDCG@10 retrieval performance \\

Yu et al.~\citep{yu2024rankrag} & NQ, TriviaQA, FEVER & RankRAG with LLaMA3 & EM, F1 across open-domain QA tasks \\

\midrule
\multicolumn{4}{l}{\textbf{Planning \& Multi-hop Chain-of-Retrieval}} \\

Wang et al.~\citep{wang2025chain} & HotpotQA, 2Wiki, MuSiQue & CoRAG (LLaMA3.1) & EM, F1 across multi-hop QA \\

Guan et al.~\citep{guan2025deeprag} & HotpotQA, MuSiQue & LLaMA3, Qwen2.5 & EM and F1 with structured retrieval reasoning \\

\midrule
\multicolumn{4}{l}{\textbf{Document-Centric \& Multimodal RAG}} \\

Yu et al.~\citep{yu2024visrag} & ArxivQA, DocVQA, ChartQA & VisRAG, GPT-4o & Retrieval accuracy and QA accuracy \\

Tan et al.~\citep{tan2025htmlrag} & HotpotQA, TriviaQA, ELI5 & HtmlRAG & EM, Hit@1, ROUGE-L \\

Zhang et al.~\citep{zhang2024ocr} & OHRBench & Qwen-VL, LLaMA models & OCR reasoning metrics (edit distance, F1) \\

Xia et al.~\citep{xia2024mmed} & MIMIC-CXR, IU-Xray & MMed-RAG, Med-Flamingo & VQA metrics (F1, AUROC) and report generation \\

\bottomrule
\end{tabularx}
\caption{Experimental setup overview of RAG-based reasoning papers (Part~1).}
\label{tab:rag-reasoning-part1}

\end{table}

\begin{table}[htbp]
\centering
\small
\begin{tabularx}{\textwidth}{ >{\raggedright\arraybackslash}p{0.25\textwidth} >{\raggedright\arraybackslash\hsize=0.8\hsize\linewidth=\hsize}X >{\raggedright\arraybackslash\hsize=0.8\hsize\linewidth=\hsize}X >{\raggedright\arraybackslash\hsize=1.4\hsize\linewidth=\hsize}X }

\toprule
\textbf{Author} & \textbf{Datasets} & \textbf{Models} & \textbf{Evaluation Setup} \\

\midrule
\multicolumn{4}{l}{\textbf{Evaluations \& Frameworks}} \\

Krishna et al.~\citep{krishna2024fact} & FRAMES benchmark & RAG pipelines & Retrieval, reasoning, and factuality evaluation \\

Cheng et al.~\citep{cheng2024coral} & Multi-turn conversation datasets & GPT-3.5, Qwen models & Citation-aware ROUGE evaluation \\

Wang et al.~\citep{wang2024omnieval} & Financial QA datasets & Llama3.1, DeepSeek models & MAP, MRR, F1 and hallucination metrics \\

Fleischer et al.~\citep{fleischer2024rag} & TriviaQA, PubMedQA & LLaMA models & RAG vs CoT comparison with RAGAS metrics \\

Fleshman et al.~\citep{fleshman2024re} & NQ, TriviaQA, HotpotQA & RankRAG variants & Retrieval ranking improvements \\

\midrule
\multicolumn{4}{l}{\textbf{Applications: Legal, Medical Search \& Finance}} \\

Singhal et al.~\citep{singhal2024evidence} & Fact-checking dataset & Mixtral, LLaMA models & METEOR, Accuracy, F1 \\

Shi et al.~\citep{shi2025searchrag} & MedQA, MMLU-Med & LLaMA3 models & Accuracy comparison vs MedRAG \\

Han et al.~\citep{xiao2025retrieval} & Financial stock datasets & StockLLM & ACC and MCC metrics \\

Deroy et al.~\citep{deroy2024retrievegpt} & Bengali-English code-mixed queries & GPT-3.5 models & Relevance detection evaluation \\

\midrule
\multicolumn{4}{l}{\textbf{Safety \& Robustness}} \\

Liang et al.~\citep{liang2025saferag} & Adversarial RAG benchmarks & GPT-4, Qwen models & F1 and attack failure rate metrics \\

\midrule
\multicolumn{4}{l}{\textbf{Graphs \& Multi-Corpora Multimodal RAG}} \\

Yeo et al.~\citep{yeo2025universalrag} & MMLU, WebQA, LVBench & UniversalRAG models & Accuracy, EM/F1 across modalities \\

\bottomrule
\end{tabularx}

\caption{Experimental setup overview of RAG-based reasoning papers (Part~2).}
\label{tab:rag-reasoning-part2}

\end{table}

\subsection{Tool-Augmented Reasoning / Agentic Reasoning}
\label{reason:IX}
\begin{figure}[htbp]
\centering
\definecolor{ink}{RGB}{38,41,54}
\definecolor{inksoft}{RGB}{92,97,115}
\definecolor{hair}{RGB}{200,205,216}
\definecolor{cBlue}{RGB}{60,87,150}
\definecolor{cTeal}{RGB}{46,125,111}
\definecolor{cAmber}{RGB}{168,120,43}
\definecolor{cViolet}{RGB}{103,87,140}
\definecolor{cRose}{RGB}{169,79,103}
\definecolor{cSlate}{RGB}{85,92,112}
\begin{tikzpicture}[
 layer/.style={rectangle, draw=#1, line width=1pt, rounded corners=5pt, fill=#1!8,
   text width=11.4cm, minimum height=1.05cm, align=left, font=\small, text=ink,
   inner xsep=10pt, inner ysep=6pt, anchor=west},
 up/.style={-{Stealth[length=2.6mm]}, draw=inksoft, line width=1.1pt},
]
\node[layer=cRose]   (l5) at (0,5.6) {\textcolor{cRose}{\textbf{Safety \& Governance}}\\[1pt]GuardReasoner, Compute Budgets, Adversarial Shields};
\node[layer=cAmber]  (l4) at (0,4.2) {\textcolor{cAmber}{\textbf{Collaboration}}\\[1pt]Generator $\leftrightarrow$ Verifier $\leftrightarrow$ Refiner \quad$\mid$\quad Distill $\rightarrow$ Single Model};
\node[layer=cTeal]   (l3) at (0,2.8) {\textcolor{cTeal}{\textbf{Orchestration}}\\[1pt]Abstract Planner $\rightarrow$ A*/MCTS Search $\rightarrow$ Tool Cards $\rightarrow$ Executor};
\node[layer=cViolet] (l2) at (0,1.4) {\textcolor{cViolet}{\textbf{Interface}}\\[1pt]Agent--Computer Interface, Document Parser, Web Browser};
\node[layer=cSlate]  (l1) at (0,0.0) {\textcolor{cSlate}{\textbf{Environment}}\\[1pt]Calculator, Web Search, Code Executor, APIs, Databases};
\draw[up] (l1.north -| 5.7,0) -- (l2.south -| 5.7,0);
\draw[up] (l2.north -| 5.7,0) -- (l3.south -| 5.7,0);
\draw[up] (l3.north -| 5.7,0) -- (l4.south -| 5.7,0);
\draw[up] (l4.north -| 5.7,0) -- (l5.south -| 5.7,0);
\end{tikzpicture}
\caption{Layered architecture of tool-augmented and agentic reasoning systems. From bottom to top: external tools provide capabilities; interfaces enable interaction; orchestration plans and sequences actions; collaboration coordinates multiple agents; safety and governance ensure assurance and compute control.}
\label{fig:agentic-architecture}
\end{figure}
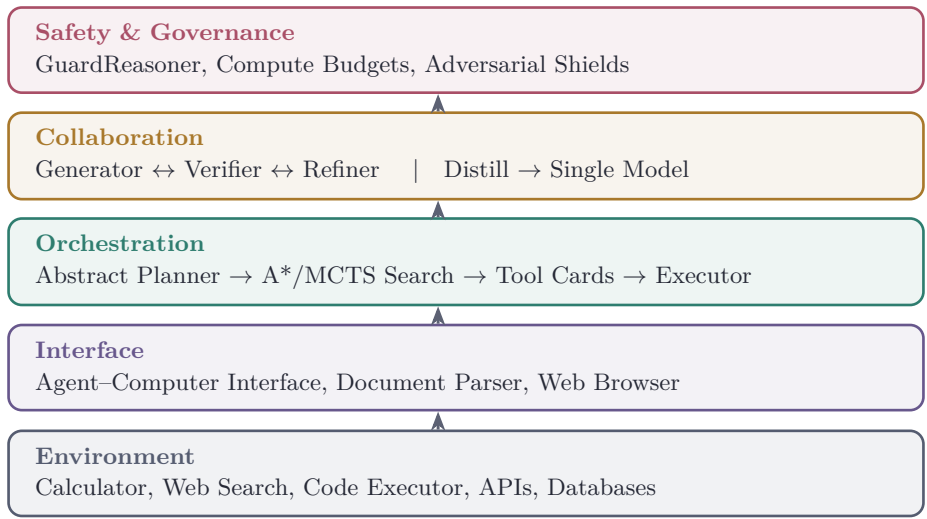

Tool-augmented and agentic reasoning moves LLMs from passive text generation toward interactive, goal-directed problem solving in the external world. Instead of relying solely on what was learned during pretraining, an agentic LLM can plan a sequence of actions, call specialized tools and APIs, retrieve up-to-date information, execute code, and adapt its strategy based on the responses it receives. This addresses several well-known limitations of standard LLM architectures: the inability to access current information, perform precise calculations, interact with software systems, or maintain long-term memory across a task. Figure~\ref{fig:agentic-architecture} presents the layered architecture that organizes these systems, from the environment and interface layers through orchestration and collaboration to safety and governance. In practice, agentic reasoning has been applied to mathematical computation, web research, software development, scientific discovery, and multi-step interactive problem solving. The following sections examine the approaches researchers have developed to enable these capabilities. Tables~\ref{tab:agent-planning-efficiency} and~\ref{tab:agent-multiagent-safety} compile the experimental details.

\subsubsection{Planning and Tool Orchestration}
Agentic systems depend on structured planning: breaking a problem into steps, choosing the right tool at each step, and sequencing actions deliberately.

Early work separated planning from execution. Chain-of-Abstraction (CoA)~\citep{gao2024efficient} generates an abstract plan with placeholders, then fills them by invoking tools. This two-phase approach improves accuracy on mathematical reasoning and question answering. TOOLCHAIN~\citep{zhuang2023toolchain} frames tool use as a search problem, applying A* search to explore different call sequences. It finds better plans while also reducing execution time.

More recent systems adopt multi-agent and hierarchical designs. OctoTools~\citep{lu2025octotools} introduces standardized ``tool cards,'' a planner, and an executor in a training-free framework, producing consistent accuracy gains across 16 domains including mathematics, medicine, and general knowledge. TAPAS~\citep{babu2025adaptive} combines symbolic planning with LLM-based agents, automatically generating task structures and executing them. This lets it generalize to new environments without manual rules.

Other approaches learn planning strategies from experience. AVATAR~\citep{wu2024avatar} refines prompts by learning from both successful and failed examples, improving performance on question answering and retrieval. START~\citep{li2025start} uses hint-based prompting and self-supervised training to encourage effective tool usage across math, science, and coding tasks without manually curated examples. What recurs across these methods is the separation of high-level planning from low-level execution and the systematic exploration of action sequences.

\subsubsection{Efficiency and Resource Management}
As agentic systems scale, managing computation matters as much as improving accuracy. ~\citep{krishnamurthy2024can} examine in-context learning and show that effective exploratory behavior in GPT-4 emerges only when interaction histories are summarized and paired with chain-of-thought prompting. ~\citep{cuadron2025danger} study the ``overthinking'' problem in software engineering tasks. They find that excessive reasoning often delays decisive action, while simple, low-overthinking selection strategies improve task success rates by nearly 30\%.

Resource allocation adds another dimension. ~\citep{fu2024efficiently} introduce Dynasor with Certaindex, a progress-aware system that dynamically assigns compute and terminates low-potential queries early, saving up to 50\% compute and lowering latency. ~\citep{li2025rethinking} question mixture-of-agents assumptions with Self-MoA, which ensembles multiple runs of a single strong model and outperforms heterogeneous model mixtures while being more efficient. These results suggest that allocating resources based on progress and favoring well-designed self-ensembling over model diversity are productive directions.

\subsubsection{Multi-Agent Collaboration and Training}
Researchers have explored how multiple agents can collaborate to solve problems that exceed the capacity of any single model. The common thread across this work is treating reasoning as a collective process, with role specialization, structured debate, and knowledge distillation as recurring design elements.

Much of the effort has gone into designing collaboration frameworks. ~\citep{jaiswal2024improving} propose MoRA, which employs a mixture of ``refinement agents,'' each specializing in diagnosing and correcting a specific error type such as miscomprehension or computation mistakes, allowing them to iteratively fix flaws in a reasoning chain. MALT~\citep{motwani2024malt} generalizes this idea by training heterogeneous agents to follow a generation-verification-refinement workflow within a multi-agent search tree, improving performance on complex reasoning tasks without human labels. CIPHER~\citep{pham2023let} takes a different route, enabling agents to ``debate'' by sharing richer belief distributions through embedding-space message passing rather than token-based communication, which leads to more accurate outcomes. MultiAgentBench~\citep{zhu2025multiagentbench} provides a dedicated benchmark with milestone-based metrics to evaluate these collaborative and competitive strategies across various network topologies.

Deploying multiple agents, though effective, is expensive. One response has been to distill the collaborative capabilities of a multi-agent system into a single, more competent model. The MAGDI~\citep{chen2024magdi} framework does this by training a student model to replicate the reasoning process of a multi-agent interaction graph, transferring collaborative problem-solving skills into one model.

Training these agents has also driven innovation in data generation and learning algorithms. SynWorld~\citep{fang2025synworld} addresses the scarcity of complex, interactive training data by synthesizing diverse interactive scenarios, allowing agents to be refined using Monte Carlo Tree Search over multi-step action paths in novel environments. Open-Reasoner-Zero~\citep{hu2025open} shows that lightweight PPO recipes, using simple rules and no complex penalties, can match or exceed the performance of heavier reinforcement learning pipelines with significantly fewer resources.

\subsubsection{Domain Applications and Interface Design}
How well an agentic system performs in practice depends largely on the interfaces connecting it to real-world tools and environments. An LLM's internal, parametric knowledge is often fragile and insufficient on its own, which makes external tools and memory systems essential for robust reasoning~\citep{wang2024knowledge}.

Several recent systems illustrate what becomes possible with well-designed interfaces. Open Deep Search~\citep{alzubi2025open} provides an open-source agent and search stack that can orchestrate dynamic web queries, supporting more complex information gathering than single-shot searches. SWE-agent~\citep{yang2024swe} goes further by introducing an Agent-Computer Interface (ACI) that lets an agent navigate a computer environment autonomously, editing code, managing repositories, and executing programs, achieving state-of-the-art performance on software engineering benchmarks. ~\citep{wu2025agentic} combine web search and code execution with a mind-map construction tool that helps organize multi-step reasoning, outperforming strong baselines on scientific synthesis tasks. In biomedical research, an AI co-scientist~\citep{gottweis2025towards} interfaces with the scientific method through generate-debate-evolve cycles, proposing and validating novel hypotheses. TheoremExplainAgent~\citep{ku2025theoremexplainagent} extends agentic reasoning into creative domains by interfacing with the Manim animation engine to plan and generate multimodal videos explaining mathematical theorems.

A less visible but equally important interface layer is document parsing. ~\citep{zhang2024document} survey this area and highlight that converting documents into a machine-usable format remains a significant bottleneck for downstream agentic systems.

\subsubsection{Safety and Security}
As agentic systems gain autonomy, their safety and security require approaches that go beyond traditional content filtering.
~\citep{liu2025guardreasoner} introduce GuardReasoner, a reasoning-capable safety system trained on 460,000 interaction steps using supervised fine-tuning and hard-sample direct preference optimization. It improves both safety performance and explainability compared to static filtering. The threat model, though, has grown in parallel with defenses. The X-Teaming~\citep{rahman2025x} framework shows how collaborative agents can escalate harmful behavior across a multi-turn conversation, achieving near-perfect success rates at bypassing safeguards. The authors also released XGuard-Train, a large-scale dataset of defensive dialogues for training countermeasures against these conversational attacks.

Vulnerabilities extend to system infrastructure as well. ~\citep{gu2025auditing} conduct systematic audits of prompt caching in LLM APIs, revealing timing side-channel attacks and cross-user cache sharing that can leak behavioral patterns and architectural details. On a separate front, TP-LLaMA~\citep{chen2024advancing} targets the agent's own reliability by learning from both successful and failed tool-use trajectories through stepwise direct preference optimization, improving generalization and making the agent a more predictable component within a larger system.

\subsubsection{Integration and Future Directions}
Research in tool-augmented and agentic reasoning has coalesced around three capabilities: systematic planning and search over tool actions, paired with agent debate and role specialization; progress-aware resource allocation, overthinking avoidance, and calibrated self-ensembling; and reasoning-capable safety systems, multi-turn adversarial evaluation, and system-level security auditing.
Open problems remain in building unified planning systems that jointly optimize tool selection, cost, and risk; in creating interfaces that blend perception, memory, and formal verification; and in establishing benchmarks that measure end-to-end success under realistic constraints such as latency limits, system failures, and safety requirements.

\begin{table}[htbp]
\centering
\small
\begin{tabularx}{\textwidth}{ >{\raggedright\arraybackslash}p{0.25\textwidth} >{\raggedright\arraybackslash\hsize=0.8\hsize\linewidth=\hsize}X >{\raggedright\arraybackslash\hsize=0.8\hsize\linewidth=\hsize}X >{\raggedright\arraybackslash\hsize=1.4\hsize\linewidth=\hsize}X }

\toprule
\textbf{Author} & \textbf{Datasets} & \textbf{Models} & \textbf{Evaluation Setup} \\
\midrule

\multicolumn{4}{l}{\textbf{Planning and Tool Orchestration}} \\

Gao et al.~\citep{gao2024efficient} & GSM8K, SVAMP, HotpotQA, NQ & LLaMA-2, Toolformer & CoA vs CoT; EM accuracy with tool integration \\

Zhuang et al.~\citep{zhuang2023toolchain} & ToolBench, GSM8K & GPT-3.5, GPT-4 & Planning with MCTS; performance vs baseline planners \\

Lu et al.~\citep{lu2025octotools} & AlgoPuzzleVQA, Hallusion-VD, PuzzleVQA & GPT-4o (base model) with OctoTools framework & Evaluated the OctoTools agentic framework across 16 benchmarks covering vision and text modalities. \\

Babu et al.~\citep{babu2025adaptive} & LLM+P benchmark domains and VirtualHome simulation tasks & GPT-4o (primary), Claude 3.7 Sonnet, GPT-4o Mini & Evaluated the TAPAS planning framework on classical planning benchmarks and simulated environments. \\

Wu et al.~\citep{wu2024avatar} & STARK, HotpotQA, ArxivQA & GPT-4, Claude 3 & Tool orchestration with multiple APIs; EM + judge metrics \\

Li et al.~\citep{li2025start} & GPQA, MATH500, LiveCodeBench & Qwen2.5, LLaMA3.3, GPT-4o & Long CoT with tool reasoning; Pass@1 \\

\midrule
\multicolumn{4}{l}{\textbf{Efficiency and Resource Management}} \\

Krishnamurthy et al.~\citep{krishnamurthy2024can} & Synthetic Bandit Instances & GPT-3.5, GPT-4, LLaMA-2 & Reward optimization metrics \\

Cuadron et al.~\citep{cuadron2025danger} & SWE-Bench & GPT-4o, Claude 3.5 & Overthinking score vs task success \\

Fu et al.~\citep{fu2024efficiently} & AIME24, GSM8K, LiveCodeBench & DeepSeek, LLaMA3 & Token efficiency and SLO improvements \\

Li et al.~\citep{li2025rethinking} & AlpacaEval 2.0, MT-Bench, MMLU & WizardLM, Gemma, Qwen & Win-rate comparisons across MoA strategies \\

\midrule

\multicolumn{4}{l}{\textbf{Multi-Agent Collaboration and Training}} \\

Anand et al.~\citep{jaiswal2024improving} & SciEval, MMLU Physics & LLaMA3, GPT-4 & Physics reasoning accuracy \\

Motwani et al.~\citep{motwani2024malt} & GSM8K, MATH & LLaMA-3.1 & Multi-agent reasoning; accuracy improvements \\

Pham et al.~\citep{pham2023let} & GSM8K, MMLU & LLaMA2, Falcon & Multi-agent debate with majority voting \\

Zhu et al.~\citep{zhu2025multiagentbench} & Simulation tasks, coding tasks & LLaMA3, GPT-4o & Task score and coordination metrics \\

Chen et al.~\citep{chen2024magdi} & StrategyQA, ARC, GSM8K & LLaMA2, Mistral & Graph-based reasoning training \\

Fang et al.~\citep{fang2025synworld} & ToolBench, HotpotQA & GPT-4, Qwen & MCTS optimization for tool workflows \\

Hu et al.~\citep{hu2025open} & MATH500, GPQA & Open-Reasoner-Zero & RLHF-based reasoning training \\

\bottomrule
\end{tabularx}
\caption{Experimental setup overview of tool-augmented/agentic reasoning papers (Part~1).}
\label{tab:agent-planning-efficiency}
\end{table}

\begin{table}[htbp]
\centering
\small
\begin{tabularx}{\textwidth}{ >{\raggedright\arraybackslash}p{0.25\textwidth} >{\raggedright\arraybackslash\hsize=0.8\hsize\linewidth=\hsize}X >{\raggedright\arraybackslash\hsize=0.8\hsize\linewidth=\hsize}X >{\raggedright\arraybackslash\hsize=1.4\hsize\linewidth=\hsize}X }

\toprule
\textbf{Author} & \textbf{Datasets} & \textbf{Models} & \textbf{Evaluation Setup} \\

\midrule
\multicolumn{4}{l}{\textbf{Domain Applications and Interface Design}} \\

Alzubi et al.~\citep{alzubi2025open} & FRAMES, SimpleQA & GPT-4o, DeepSeek-R1 & Search-augmented agent evaluation \\

Yang et al.~\citep{yang2024swe} & SWE-Bench & GPT-4 Turbo, Claude 3 & Automated code patching performance \\

Wu et al.~\citep{wu2025agentic} & GPQA, Deep Research datasets & GPT-4o, LLaMA3 & Agentic reasoning evaluation \\

Gottweis et al.~\citep{gottweis2025towards} & Biomedical tasks & Gemini 2.0 & Multi-agent hypothesis generation \\

Ku et al.~\citep{ku2025theoremexplainagent} & TheoremExplainBench & GPT-4o, Claude 3.5 & Multimodal explanation evaluation \\

\midrule
\multicolumn{4}{l}{\textbf{Safety and Security}} \\

Liu et al.~\citep{liu2025guardreasoner} & HarmBench, SafeRLHF & GuardReasoner, GPT-4o & Weighted F1 safety metrics \\

Rahman et al.~\citep{rahman2025x} & HarmBench, SafeMTData & GPT-4o, Qwen2.5 & Adversarial attack success rate \\

Gu et al.~\citep{gu2025auditing} & Synthetic prompt caching dataset & GPT-4o, Claude, Gemini & Prompt caching detection accuracy \\

Chen et al.~\citep{chen2024advancing} & ToolBench & ChatGPT, ToolLLaMA & Completion/pass metrics \\

\bottomrule
\end{tabularx}

\caption{Experimental setup overview of tool-augmented/agentic reasoning papers (Part~2).}
\label{tab:agent-multiagent-safety}

\end{table}

\subsection{Reinforcement Learning for Reasoning}
\label{reason:X}
\begin{figure}[htbp]
\centering
\definecolor{ink}{RGB}{38,41,54}
\definecolor{inksoft}{RGB}{92,97,115}
\definecolor{hair}{RGB}{200,205,216}
\definecolor{cBlue}{RGB}{60,87,150}
\definecolor{cAmber}{RGB}{168,120,43}
\definecolor{cGreen}{RGB}{92,136,85}
\definecolor{cRose}{RGB}{169,79,103}
\definecolor{panel}{RGB}{244,245,248}
\begin{tikzpicture}[
 hub/.style={circle, draw=ink, line width=1.1pt, fill=ink, text=white,
   minimum size=1.9cm, align=center, font=\small\bfseries},
 spoke/.style={rectangle, draw=#1, line width=0.9pt, rounded corners=5pt, fill=#1!8,
   minimum height=1.35cm, align=center, font=\scriptsize, text width=3.0cm, text=ink, inner sep=5pt},
 axis/.style={{Stealth[length=2.4mm]}-{Stealth[length=2.4mm]}, draw=#1, line width=1.1pt},
]
\node[hub] (c) {LLM\\Policy};
\node[spoke=cGreen] (s1) at (0, 2.7)    {\textcolor{cGreen}{\textbf{Learn to Search}}\\MCTS, A*, internalized planning\\(Searchformer, COAT/Satori)};
\node[spoke=cBlue]  (s2) at (-3.9,-1.7) {\textcolor{cBlue}{\textbf{Optimize the Chain}}\\Stepwise DPO, process rewards\\(IRPO, Flow-DPO, RM-R1)};
\node[spoke=cAmber] (s3) at (3.9,-1.7)  {\textcolor{cAmber}{\textbf{Govern Compute}}\\Adaptive budgets, offline critics,\\domain RL (Dynasor, SWE-RL)};
\draw[axis=cGreen] (c) -- (s1);
\draw[axis=cBlue]  (c) -- (s2);
\draw[axis=cAmber] (c) -- (s3);
\node[rectangle, draw=cRose, line width=0.9pt, fill=panel, rounded corners=5pt,
  minimum width=11cm, align=center, font=\scriptsize, text=ink, inner sep=7pt] (em) at (0,-4.0)
  {\textcolor{cRose}{\textbf{Pure RL Emergence}}~~ Random Init $\rightarrow$ Verifiable Rewards $\rightarrow$ Emergent Reasoning (DeepSeek-R1, Absolute Zero, RLSP)};
\draw[-{Stealth[length=2mm]}, draw=hair, line width=0.8pt, dashed] (c) -- (em);
\end{tikzpicture}
\caption{Three levers of reinforcement learning for reasoning. The LLM policy is refined through: (1)~learning to search via MCTS and internalized planning, (2)~optimizing reasoning chains via stepwise preferences and process rewards, and (3)~governing compute via adaptive budgets and offline critics. Pure RL emergence (bottom) shows structured reasoning arising from scratch with verifiable rewards alone.}
\label{fig:rl-reasoning-loop}
\end{figure}
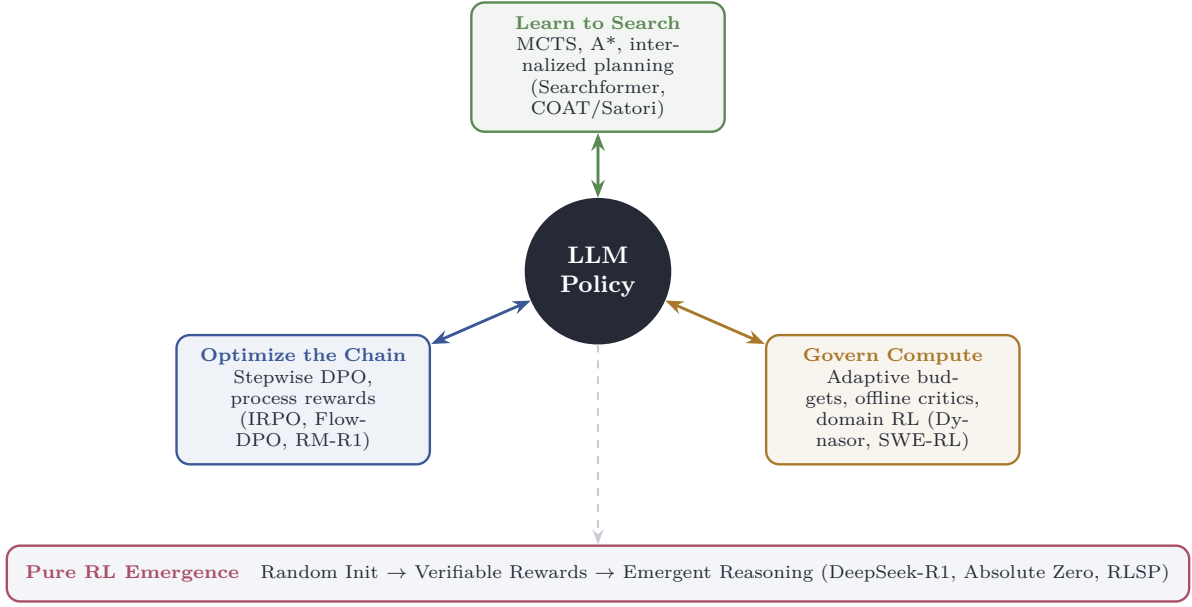

RL for reasoning moves LLMs beyond passive next-token prediction toward \emph{procedural} problem solving: learning to search, allocate compute, prefer good chains over bad ones, and verify along the way. Figure~\ref{fig:rl-reasoning-loop} maps the three main levers: learning to search via MCTS and internalized planning, optimizing reasoning chains via stepwise preferences and process rewards, and governing compute via adaptive budgets and offline critics. Recent work clusters into (i) search/planning policies, (ii) stepwise preference learning, (iii) process rewards and tree search, (iv) pure-RL emergence and self-play, (v) efficiency and compute governance, (vi) offline critics and domain RL, (vii) preference modeling robustness, and (viii) surveys/blueprints. Tables~\ref{tab:rl-reasoning-methods} and~\ref{tab:rl-reasoning-alignment} detail the experimental setups.

\subsubsection{Search \& Internalized Planning}
 ~\citep{lehnert2024beyond} train \emph{Searchformer} to imitate and improve A* dynamics on Sokoban, bootstrapping a neural planner that solves with fewer steps and smaller models. 
 ~\citep{shen2025satori} propose \emph{COAT} (Chain-of-Action-Thought): a 7B model learns internal autoregressive search via format tuning then RL, iterating reflect$\rightarrow$act$\rightarrow$explore and reaching SOTA on math. 
\emph{Takeaway:} Teaching models the \emph{shape} of search (explicitly or internalized) yields reliable, reusable planning behavior.

\subsubsection{Stepwise Preference Optimization}
Recent preference optimization work has shifted toward step-level supervision, targeting individual reasoning steps rather than comparing full traces. ~\citep{pang2024iterative} introduce Iterative Reinforcement of Preference Optimization (IRPO), an extension of DPO that contrasts entire CoT traces while penalizing incorrect intermediate steps. It improves performance on GSM8K, MATH, and ARC without additional annotations or external data.
~\citep{lu2024step} present Step-Controlled DPO, which generates synthetic negatives at pre-specified reasoning steps, enabling more precise credit assignment and achieving strong results across open models. ~\citep{deng2024flow} extend the idea into a collaborative setting with Flow-DPO, an online multi-agent framework where LLMs iteratively co-construct solutions and DPO is applied to evolving trajectories rather than static outputs. Across these methods, step-aware preference signals, whether from trace alignment, targeted error insertion, or agent-driven feedback, produce more faithful and interpretable reasoning than optimizing for final answers alone.

\subsubsection{Process Rewards \& Tree-Search Ensembles}

Rather than scoring only final answers, several methods now align generation with stepwise reward signals, using the structure of intermediate reasoning to guide search. ~\citep{park2024ensembling} propose LE-MCTS, which ensembles multiple LLMs through MCTS guided by process reward models that score intermediate trajectories. This outperforms both single-model baselines and naive ensembles by promoting more coherent solution paths. ~\citep{zhang2024rest} introduce ReST-MCTS, which estimates per-step rewards from reasoning traces and then self-trains both policy and reward models within an MCTS framework. Without manual annotations, it surpasses prior approaches like ReST-EM and Best-of-N.

~\citep{zhang2025lessons} analyze pitfalls in Process Reward Model (PRM) training, including Monte Carlo bias and overfitting in Best-of-N setups, and propose consensus filtering for more robust supervision signals. ~\citep{chen2025rm} take a different angle with RM-R1, which treats reward modeling itself as a reasoning problem: instead of scalar rewards, the model generates textual rubrics and reasoning traces before scoring answers. This generative framing improves both interpretability and benchmark performance on reasoning-intensive tasks.

\subsubsection{Pure RL \& Emergent Reasoning}
Several recent results show that structured reasoning can emerge from reinforcement learning alone, without supervised fine-tuning. ~\citep{guo2025deepseek} train DeepSeek-R1 entirely with RL from a randomly initialized base. A multi-stage recipe that adds readability refinements and distillation yields competitive reasoning on complex benchmarks despite the absence of any supervised starting point. ~\citep{zhao2025absolute} push this further with Absolute Zero, where a model learns through self-play: it generates its own tasks, verifies them using a code executor, and trains on the resulting reward signal. The model reaches state-of-the-art performance in math and programming without external supervision of any kind.

Ye et al.~\citep{ye2025emergence} introduce RLSP, which separates exploration from correctness in the reward design. They find that exploration rewards alone can induce backtracking, verification, and iterative refinement, behaviors typically associated with advanced reasoning. Verifiable, exploration-driven signals appear sufficient to produce structured reasoning patterns from scratch.

\subsubsection{Efficiency \& Compute Governance}
How computation is allocated during multi-step generation matters as much as raw performance. Several recent methods optimize inference-time efficiency through reasoning-aware control. ~\citep{liao2025reward} use a reward model to dynamically steer draft-target switching in speculative decoding, accelerating generation while preserving answer quality. ~\citep{arora2502training} train models to predict instance-specific reasoning budgets, tailoring compute to question difficulty and sustaining accuracy with fewer tokens.

Yin et al.~\citep{yin2024reasoning} take a different approach with UAG, which monitors token-level uncertainty during reasoning. When confidence drops, UAG revisits earlier steps and injects corrective hints, maintaining correctness with minimal overhead. ~\citep{setlur2024rl} fine-tune LLMs on mixed-quality synthetic traces, including structured incorrect ones, using a credit attribution reward. This matches the gains of $8\times$ data augmentation in math reasoning tasks without the corresponding data cost.

\subsubsection{Offline Critics \& Domain-Scale RL}
RL can also be scaled and stabilized through offline supervision and domain-specific reward signals. ~\citep{xiang2024retrospex} introduce Retrospex, which adds a frozen, offline-trained critic on top of an LLM. The critic evaluates each candidate's reasoning step and rescales its influence by predicted value, improving accuracy without inflating the context window or requiring model retraining. Because policy and value modeling are decoupled, the critic can be applied as a plug-and-play overlay to any open-weight LLM.

~\citep{wei2025swe} move RL to a larger scale with SWE-RL, trained over the longitudinal evolution of open-source software repositories. Structured reward functions based on issue resolution, code changes, and pull request workflows guide a 70B-parameter model (LLaMA3-SWE-RL) that reasons across software engineering workflows, surpassing code- and math-only instruction tuning. Offline critics paired with structured, real-world domains provide the feedback loops needed to move RL-based reasoning beyond synthetic benchmarks.

\subsubsection{Preference Modeling \& Alignment Robustness}
The reliability of preference models (PMs) used for alignment is itself an active area of work. ~\citep{go2023compositional} decompose global preference judgments into interpretable features (helpfulness, factuality, etc.) with Compositional Preference Models (CPMs). Feature scores are extracted via a prompted LM and combined through logistic regression. This modular design reduces reward hacking and generalizes better than monolithic scalar PMs. ~\citep{siththaranjan2023distributional} address a different weakness: annotator and context uncertainty. Their Distributional Preference Learning (DPL) framework learns over distributions rather than point estimates, capturing ambiguity in human feedback and reducing overconfidence. The result is stronger resilience against jailbreaks and adversarial attacks.

\subsubsection{Surveys \& Blueprints}
Xu et al.~\citep{xu2025towards} survey the field of reinforced reasoning, organizing it around three axes: data construction (MCTS, process-level annotation), learning paradigms (Process Reward Models, RLHF), and inference-time techniques (tree-of-thought, test-time scaling). They frame this trajectory as a roadmap toward Large Reasoning Models, LLMs designed specifically for long-horizon, structured problem-solving.

Besta et al.~\citep{besta2025reasoning} offer a modular blueprint that breaks reasoning into discrete components: structures (thoughts, plans), search procedures (beam, MCTS), supervision (outcome or process), and learning algorithms (RL, imitation). Both efforts provide useful scaffolding for building reasoning-first LLMs.

\subsubsection{Summary \& Outlook}
Across approaches, three levers dominate: (1) \emph{learn to search} (internal or tree-based with process rewards), (2) \emph{optimize the chain} (stepwise preferences, uncertainty-aware backtracking, reward-guided decoding), and (3) \emph{govern compute} (instance-adaptive budgets, offline critics, realistic domains). Open challenges include unified metrics for process quality, principled reward design without leakage, and scaling self-play to diverse, verifiable tasks beyond math/code.

\begin{table}[htbp]
\centering
\small
\begin{tabularx}{\textwidth}{ >{\raggedright\arraybackslash}p{0.25\textwidth} >{\raggedright\arraybackslash\hsize=0.8\hsize\linewidth=\hsize}X >{\raggedright\arraybackslash\hsize=0.8\hsize\linewidth=\hsize}X >{\raggedright\arraybackslash\hsize=1.4\hsize\linewidth=\hsize}X }

\toprule
\textbf{Author} & \textbf{Datasets} & \textbf{Models} & \textbf{Evaluation Setup} \\
\midrule

\multicolumn{4}{l}{\textbf{Search \& Internalized Planning}} \\

Lehnert et al.~\citep{lehnert2024beyond} & Maze Navigation, Sokoban & Searchformer (45M–757M) & Success Rate, SWC, ILR on optimal path finding \\

Shen et al.~\citep{shen2025satori} & GSM8K, MATH500, AIME2024 & Satori-Qwen-7B & Chain-of-Action-Thought RL with greedy decoding \\

\midrule
\multicolumn{4}{l}{\textbf{Stepwise Preference Optimization}} \\

Pang et al.~\citep{pang2024iterative} & GSM8K & LLaMA-2-70B-chat & Iterative preference optimization with majority voting \\

Lu et al.~\citep{lu2024step} & GSM8K, MATH, APE210K, CMATH & Mistral-7B, InternLM2-20B & Stepwise Contrastive DPO; greedy decoding accuracy \\

Deng et al.~\citep{deng2024flow} & MetaMath (GSM8K+MATH) & LLaMA-3-8B, Phi-3-14B & Flow-SFT training using flow traces \\

\midrule
\multicolumn{4}{l}{\textbf{Process Rewards \& Tree-Search Ensembles}} \\

Park et al.~\citep{park2024ensembling} & GSM8K, MATH500, SVAMP, ASDiv & LLaMA-3-8B, Gemma-2-9B & Process reward guided MCTS evaluation \\

Zhang et al.~\citep{zhang2024rest} & MATH, GSM8K, GPQA & LLaMA-3, Mistral & ReST-MCTS* reasoning with process rewards \\

Zhang et al.~\citep{zhang2025lessons} & GSM8K, MATH, OlympiadBench & Qwen2.5-Math-7B & PRM evaluation using prm@8 and maj@8 \\

Chen et al.~\citep{chen2025rm} & RewardBench, RM-Bench & RM-R1 (Qwen-based) & Reward model evaluation on reasoning benchmarks \\

\midrule
\multicolumn{4}{l}{\textbf{Pure RL \& Emergent Reasoning}} \\

Yang et al.~\citep{guo2025deepseek} & MMLU, GPQA, AIME2024, Codeforces & DeepSeek-R1 & pass@1 and cons@64 across reasoning tasks \\

Zhao et al.~\citep{zhao2025absolute} & LiveCodeBench, HumanEval+, AIME’24 & Qwen2.5, LLaMA3.1 & RL self-play training without curated datasets \\

Ye et al.~\citep{ye2025emergence} & MATH500, AIME2024 & LLaMA-3.1-8B, Qwen2.5-32B & RL search promotion evaluation \\

\bottomrule
\end{tabularx}
\caption{Experimental setup overview of RL for reasoning papers (Part~1).}
\label{tab:rl-reasoning-methods}
\end{table}

\begin{table}[htbp]
\centering
\small
\begin{tabularx}{\textwidth}{ >{\raggedright\arraybackslash}p{0.25\textwidth} >{\raggedright\arraybackslash\hsize=0.8\hsize\linewidth=\hsize}X >{\raggedright\arraybackslash\hsize=0.8\hsize\linewidth=\hsize}X >{\raggedright\arraybackslash\hsize=1.4\hsize\linewidth=\hsize}X }

\toprule
\textbf{Author} & \textbf{Datasets} & \textbf{Models} & \textbf{Evaluation Setup} \\
\midrule

\multicolumn{4}{l}{\textbf{Efficiency \& Compute Governance}} \\

Liao et al.~\citep{liao2025reward} & MATH500, GSM8K & Qwen2.5, LLaMA3 & Reward shaping and FLOPs efficiency analysis \\

Arora et al.~\citep{arora2502training} & GSM8K, AIME & PPO + RLOO LLMs & Optimizing correctness and brevity in reasoning \\

Yin et al.~\citep{yin2024reasoning} & GSM8K, ARC, StrategyQA & LLaMA2, Mistral & Token usage and cost-efficiency evaluation \\

Setlur et al.~\citep{setlur2024rl} & GSM8K, MATH & GPT-4, Gemini 1.5 Pro & Training using synthetic incorrect reasoning traces \\

\midrule
\multicolumn{4}{l}{\textbf{Offline Critics \& Domain-Scale RL}} \\

Xiang et al.~\citep{xiang2024retrospex} & ScienceWorld, ALFWorld, WebShop & Flan-T5, LLaMA3 agents & Offline RL critic improving agent performance \\

Wei et al.~\citep{wei2025swe} & SWE-Bench Verified & LLaMA3-SWE-RL-70B & Software repair success rate \\

\midrule
\multicolumn{4}{l}{\textbf{Preference Modeling \& Alignment Robustness}} \\

Go et al.~\citep{go2023compositional} & HH-RLHF, SHP & Flan-T5, GPT models & Preference modeling with interpretable features \\

Siththaranjan et al.~\citep{siththaranjan2023distributional} & SHP-style theoretical setup & BTL, Least Squares, DPL & Theoretical preference learning analysis \\

\bottomrule
\end{tabularx}

\caption{Experimental setup overview of RL for reasoning papers (Part~2).}

\label{tab:rl-reasoning-alignment}

\end{table}

\subsection{Multilingual Reasoning}
\label{reason:XI}
\begin{figure}[htbp]
\centering
\definecolor{ink}{RGB}{38,41,54}
\definecolor{inksoft}{RGB}{92,97,115}
\definecolor{cBlue}{RGB}{60,87,150}
\definecolor{cAmber}{RGB}{168,120,43}
\definecolor{cGreen}{RGB}{92,136,85}
\definecolor{cViolet}{RGB}{103,87,140}
\definecolor{cRose}{RGB}{169,79,103}
\scalebox{0.8}{
\begin{tikzpicture}[
 node distance=0.7cm,
 box/.style={rectangle, draw=#1, line width=1pt, rounded corners=5pt, fill=#1!8,
   minimum height=1.7cm, align=center, font=\small, text width=2.9cm, text=ink},
 path/.style={rectangle, draw=#1, line width=0.9pt, rounded corners=5pt, fill=#1!8,
   minimum height=0.95cm, align=center, font=\scriptsize, text width=3.6cm, text=ink, inner sep=5pt},
 arr/.style={-{Stealth[length=2.4mm]}, draw=inksoft, line width=1pt},
 elab/.style={font=\scriptsize, text=inksoft, fill=white, inner sep=1.5pt},
]
\node[box=cBlue] (eng) {\textcolor{cBlue}{\textbf{English}}\\Reasoning\\Model};
\node[path=cGreen,  right=1.8cm of eng, yshift=1.8cm]  (p1) {\textcolor{cGreen}{\textbf{Path A: Model Merging}}\\Merge with language LLM\\(low cost, \$120)};
\node[path=cViolet, right=1.8cm of eng]                (p2) {\textcolor{cViolet}{\textbf{Path B: Zero-shot Bridge}}\\LangBridge aligns\\representations};
\node[path=cAmber,  right=1.8cm of eng, yshift=-1.8cm] (p3) {\textcolor{cAmber}{\textbf{Path C: Path Alignment}}\\MAPO / mCoT aligns\\reasoning steps};
\node[box=cRose, right=1.8cm of p2] (tgt) {\textcolor{cRose}{\textbf{Target}}\\Languages\\{\scriptsize AR, ZH, TH,\\HI, FR, \ldots}};
\draw[arr] (eng) -- node[elab, pos=0.55] {answers only}    (p1);
\draw[arr] (eng) -- node[elab, pos=0.55] {representations}  (p2);
\draw[arr] (eng) -- node[elab, pos=0.55] {reasoning process}(p3);
\draw[arr] (p1) -- (tgt);
\draw[arr] (p2) -- (tgt);
\draw[arr] (p3) -- (tgt);
\end{tikzpicture}
}
\caption{Three strategies for cross-lingual reasoning transfer. Path~A merges reasoning-strong and language-specific models at low cost. Path~B aligns multilingual representations via lightweight zero-shot bridges. Path~C directly aligns reasoning processes across languages through preference optimization and structured reasoning formats.}
\label{fig:multilingual-transfer}
\end{figure}
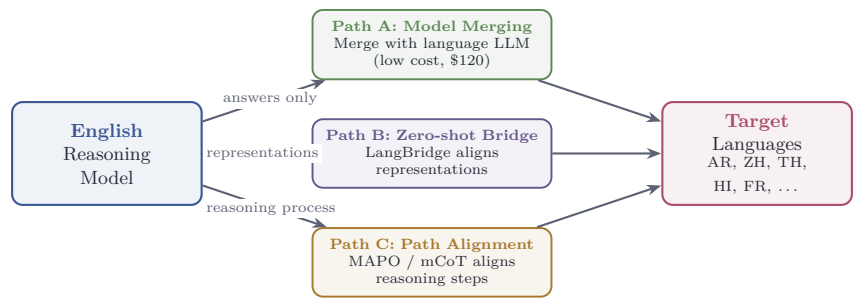

Multilingual reasoning requires models to carry out coherent, step-by-step logical processes across languages that differ in script, grammar, cultural context, and domain vocabulary. Most reasoning capabilities are developed on English data, so transferring them to languages with different morphological patterns, syntactic structures, and cultural reasoning norms is a persistent difficulty.
Deploying AI systems globally makes this more than an academic question, as reliable reasoning across linguistic communities is a practical requirement. Translating reasoning datasets rarely captures language-specific reasoning patterns or the ways logical structure varies across traditions. Figure~\ref{fig:multilingual-transfer} summarizes three strategies for bridging this gap: model merging, zero-shot representation alignment, and direct reasoning-path optimization. Recent work has tackled this through efficient cross-lingual adaptation, modular transfer mechanisms, and multilingual reasoning frameworks that try to preserve both linguistic fidelity and logical coherence. Tables~\ref{tab:multilingual-reasoning-part1} and~\ref{tab:multilingual-reasoning} provide the experimental details.

\subsubsection{Efficient Cross-Lingual Adaptation}
Developing multilingual reasoning capabilities traditionally required extensive retraining on multilingual datasets, but recent approaches have focused on efficient methods that can add reasoning capabilities to existing multilingual models or adapt reasoning-strong models to new languages.
Early approaches to multilingual reasoning often relied on retraining large models from scratch with extensive multilingual datasets, an approach that is both computationally expensive and difficult to scale across the world’s languages. Recent work has shifted toward more efficient methods that adapt existing reasoning-strong models or merge complementary capabilities without incurring the full cost of pretraining.
~\citep{pipatanakul2025open} demonstrate a practical merging strategy, combining reasoning-heavy models such as DeepSeek R1 with language-specific LLMs using open datasets. The process retains native fluency while transferring reasoning ability at a cost of roughly \$120. ~\citep{huang2024mindmerger} propose MindMerger, which parameter-efficiently injects multilingual competence into reasoning-focused models via a lightweight two-step merging process. Their method delivers up to 8\% gains in low-resource languages while leaving the base reasoning model intact.
~\citep{fan2025slam} take a different approach with SLAM, which fine-tunes only the feed-forward sublayers of six transformer layers, amounting to just 6.5–8\% of parameters. This selective update enhances multilingual reasoning across ten languages in a single stage, preserving existing reasoning strength while closing linguistic gaps.

\subsubsection{Cross-Lingual Knowledge Transfer}
Recent research shows that LLM reasoning abilities can be transferred across languages through lightweight architectures, dynamic routing, and multilingual ensembles, without requiring extensive multilingual supervision.
LangBridge~\citep{yoon2024langbridge} implements a zero-shot interface that aligns a multilingual encoder (e.g., mT5) with an English-trained reasoning model such as MetaMath via minimal trainable layers. Without multilingual data, it elevates MGSM accuracy from 40.5 \% to 53.5 \%, matching PaLM-540B, and improves low-resource language performance by exploiting language-agnostic representations.
~\citep{hu2024large} separate reasoning into knowledge retrieval versus knowledge-free reasoning. Factual retrieval falters across languages, but pure reasoning transfers nearly perfectly by sharing neuron activations and hidden patterns. This points to the modularity of reasoning capabilities.
AdaMCoT~\citep{weihua2026adamcot} introduces adaptive chain-of-thought routing, where the model thinks through intermediate language-agnostic steps before responding in the target language. This improves factual reasoning and cross-lingual consistency in low-resource settings without additional pretraining.

\subsubsection{Reasoning Path Alignment Across Languages}
Recent work has shifted attention from answer accuracy alone to aligning the reasoning processes themselves across languages, so that logical steps remain consistent regardless of the language used.
MAPO (Multilingual Alignment through Preference Optimization)~\citep{she2024mapo} frames this as a preference optimization problem, using DPO and PPO with translation-consistency rewards. It yields substantial improvements without stepwise supervision by directly optimizing for reasoning process similarity across languages. mCoT (multilingual Chain-of-Thought)~\citep{lai2024mcot} instruction-tunes models on multilingual mathematical reasoning data to improve consistency across 11 languages, with strong benefits for low-resource settings.
~\citep{payoungkhamdee2025towards} evaluate Program-of-Thought (PoT) reasoning in multilingual contexts and find that PoT fine-tuning consistently outperforms traditional chain-of-thought methods. They propose code quality as a reliable proxy for reasoning quality, suggesting that program-based reasoning provides a more stable mechanism for cross-lingual transfer than natural language chains alone.

\subsubsection{Datasets and Benchmarks}
Foundational math-oriented resources like MGSM~\citep{shi2022language}, a translation of GSM8K into 10 languages, revealed that chain-of-thought reasoning scales with model size and transfers across tasks. MGSM8KInstruct and the xMR (MathOctopus) models~\citep{chen2023breaking} leverage rejection sampling and supervised fine-tuning to push multilingual mathematical reasoning further, in some cases surpassing ChatGPT. Beyond math, the XCOPA benchmark~\citep{ponti2020xcopa} enables evaluation of causal commonsense reasoning across 11 languages and shows that translation-based datasets can outperform naive multilingual pretraining.
More recent benchmarks have expanded both scope and cultural grounding. BenchMAX~\citep{huang2025benchmax} provides a large-scale evaluation suite spanning reasoning, long-context understanding, instruction following, and code generation across 16 languages. P-MMEval~\citep{zhang2024p} introduces a parallel, multi-task framework covering mathematics, logic, code generation, and instruction-following in over ten languages, ensuring consistent cross-lingual comparison. MultiNRC~\citep{fabbri2025multinrc} targets culturally grounded reasoning in French, Spanish, and Chinese, including linguistic puzzles, wordplay, and math with cultural relevance, revealing performance gaps that simple dataset translation cannot bridge. mSCoRe~\citep{ngo2025mscore} focuses on skill-based multilingual commonsense reasoning with a taxonomy of reasoning skills and progressive complexity scaling; it remains highly challenging for even the strongest models.

\subsubsection{Multilingual \emph{Multimodal} Reasoning}
Models that must reason across both languages and modalities face compounded difficulty: linguistic diversity and visual understanding must be handled together while maintaining logical coherence. M4U~\citep{wang2024m4u} evaluates this across 64 academic disciplines and six languages. Even frontier models like GPT-4o, which perform well on monolingual multimodal tasks, show large accuracy drops when required to reason across languages and modalities simultaneously. ~\citep{bang2023multitask} evaluate ChatGPT across 23 datasets spanning multitask, multilingual, and multimodal settings. They find strengths in deductive reasoning alongside systematic weaknesses in handling non-Latin scripts and frequent hallucinations, exposing how fragile current approaches are when scaling across both language and modality.
To better understand these limitations, ~\citep{song2024missing} disentangle the contributions of multilinguality, complex reasoning, and multimodality. Their controlled experiments reveal that targeted interventions such as careful translation pipelines, visual programming methods, and improved image captioning can substantially boost open-source models’ zero-shot performance in multilingual multimodal reasoning. The results suggest that modular approaches combining translation, visual programming, and multimodal alignment can outperform end-to-end training when visual information is preserved across linguistic variation.

\subsubsection{Integration and Future Directions}
Work in this area has been organized around three directions: efficient cross-lingual adaptation through model merging and selective parameter tuning; alignment techniques that optimize reasoning processes rather than just answers, using preference optimization and structured reasoning formats; and modular solutions for multimodal settings that preserve both visual information and linguistic authenticity. ~\citep{ghosh2025multilingual} survey multilingual reasoning datasets, techniques, and open challenges, identifying issues in alignment methodology, bias mitigation, and low-resource language support.
Persistent challenges remain in developing language-agnostic reasoning representations, creating evaluation methods that avoid cultural and linguistic biases, and scaling reasoning capabilities to low-resource languages. Cost-effective adaptation, cross-lingual process alignment, and targeted multimodal solutions form the practical stack, but progress depends on multilingual benchmarks rigorous enough to expose where current methods fall short.

\begin{table}[htbp]
\centering
\small
\begin{tabularx}{\textwidth}{ >{\raggedright\arraybackslash}p{0.25\textwidth} >{\raggedright\arraybackslash\hsize=0.8\hsize\linewidth=\hsize}X >{\raggedright\arraybackslash\hsize=0.8\hsize\linewidth=\hsize}X >{\raggedright\arraybackslash\hsize=1.4\hsize\linewidth=\hsize}X }

\toprule
\textbf{Author} & \textbf{Datasets} & \textbf{Models} & \textbf{Evaluation Setup} \\
\midrule

\multicolumn{4}{l}{\textbf{Efficient Cross-Lingual Adaptation}} \\

Pipatanakul et al.~\citep{pipatanakul2025open} & IFEval, MT-Bench, WangchanThaiInstruct & Typhoon2-70B Instruct, DeepSeek R1-70B Distill & Reasoning performance is measured on math and coding benchmarks, while language ability is assessed using instruction-following, MT-Bench judging\\

Huang et al.~\citep{huang2024mindmerger} & MGSM, MSVAMP, X-CSQA, XNLI & LLaMA 2-7B, mT5-XL, NLLB-200 3.3B & Experiments evaluate MindMerger (Soft and Hard variants) against multilingual reasoning baselines \\

Fan et al.~\citep{fan2025slam} & MGSM, MSVAMP & LLaMA 2 (7B, 13B), MetaMath & SLAM is compared with multilingual reasoning baselines (MAmmoTH, WizardMath, MetaMath, MathOctopus) \\

\midrule
\multicolumn{4}{l}{\textbf{Cross-Lingual Knowledge Transfer}} \\

Yoon et al.~\citep{yoon2024langbridge} & MGSM, HumanEval-MT, BBH, BBH-BN & LLaMA 2, MetaMath, Orca 2, Code Llama, mT5 & Evaluated across four reasoning tasks: mathematical reasoning, code completion, logical reasoning, and commonsense reasoning. \\

Hu et al.~\citep{hu2024large} & StrategyQA, KFRD, QASC, ASDiv, Coin Flip, ProofWriter & LLaMA 2-7B-Chat, BLOOMZ-MT-7B, Mistral-7B-Instruct-v0.1, Qwen-1.5-7B-Chat & Experiments analyze cross-lingual transfer in reasoning tasks under varying knowledge retrieval requirements. \\


\midrule
\multicolumn{4}{l}{\textbf{Reasoning Path Alignment Across Languages}} \\

She et al.~\citep{she2024mapo} & MGSM, MSVAMP, MNumGLUESub & LLaMA 2 (7B, 13B), MetaMath, Mistral, MathOctopus & Accuracy for reasoning performance, Perplexity-based Alignment Score (PPL)\\

Lai et al.~\citep{lai2024mcot} & GSM8K, MGSM, MSVAMP, MetaMathQA, MathInstruct & Mistral-7B, LLaMA 2, Qwen, WizardMath, MathOctopus, xCoT, mCoT & Evaluated on multilingual mathematical reasoning tasks using few-shot Chain-of-Thought prompting \\

Payoungkhamdee et al.~\citep{payoungkhamdee2025towards} & MGSM, GSM8K, DGSM8KPoT, MGSM8KPoT & Llama 2 (7B, 13B), CodeLlama (7B), Llama 3 (8B), & Evaluated on MGSM across 10 languages in zero-shot settings. \\

\bottomrule
\end{tabularx}
\caption{Experimental setup overview of multilingual reasoning papers (Part~1).}
\label{tab:multilingual-reasoning-part1}
\end{table}

\begin{table}[htbp]
\centering
\small
\begin{tabularx}{\textwidth}{ >{\raggedright\arraybackslash}p{0.25\textwidth} >{\raggedright\arraybackslash\hsize=0.8\hsize\linewidth=\hsize}X >{\raggedright\arraybackslash\hsize=0.8\hsize\linewidth=\hsize}X >{\raggedright\arraybackslash\hsize=1.4\hsize\linewidth=\hsize}X }

\toprule
\textbf{Author} & \textbf{Datasets} & \textbf{Models} & \textbf{Evaluation Setup} \\

\midrule
\multicolumn{4}{l}{\textbf{Datasets and Benchmarks}} \\

Shi et al.~\citep{shi2022language} & MGSM & GPT-3 (text-davinci-002), PaLM (PaLM-540B) & Models are evaluated on MGSM across 12 languages using few-shot prompting \\

Chen et al.~\citep{chen2023breaking} & MGSM (in-domain), SVAMP,GSM8K & ChatGPT (gpt-3.5-turbo), GPT-4, LLaMA, RFT, MAmmoTH, WizardMath, MathOctopus & Models are evaluated on MGSM and MSVAMP across 10 languages \\

Ponti et al.~\citep{ponti2020xcopa} & XCOPA (based on COPA) & mBERT, XLM-RoBERTa (Base, Large), Universal Sentence Encoder & Evaluated on XCOPA multiple-choice causal reasoning tasks across 11 languages \\

Huang et al.~\citep{huang2025benchmax} & BenchMAX & Llama 3.1, Qwen 2.5, Gemma 2, InternLM 2.5 & Models are evaluated across 17 languages on the BenchMAX multilingual benchmark \\

Zhang et al.~\citep{zhang2024p} & MGSM, MMMLU, MLogiQA, MHellaSwag & GPT-4o, Claude 3.7 Sonnet, Llama 3.1, Llama 3.2 & Models are evaluated on multilingual reasoning, understanding, generation, and translation tasks using three prompting strategies\\

Mares et al.~\citep{fabbri2025multinrc} & MultiNRC & Gemini 2.5 Pro, o3, o4-mini, GPT-4.1 & Models are evaluated on MultiNRC multilingual reasoning tasks across four reasoning categories \\

Ngo et al.~\citep{ngo2025mscore} & mSCoRe-G, mSCoRe-S & GPT-4o, OpenAI o1, OpenAI o1-mini & Models are evaluated on multilingual commonsense and social reasoning tasks across five languages \\

\midrule
\multicolumn{4}{l}{\textbf{Multilingual Multimodal Reasoning}} \\

Wang et al.~\citep{wang2024m4u} & M4U & Qwen-1.5-7B-Chat, Qwen-1.5-14B-Chat, Mistral-Instruct-v0.2-7B ,Gemini 1.0 Pro & Zero-shot multiple-choice reasoning on M4U with both text-only and multimodal inputs \\

Song et al.~\citep{song2024missing} & NLVR2 (English image pairs, train/val/test), MaRVL (multilingual image pairs: id, sw, ta, tr, zh) & Open LMMs (mBLIP, LLaVA, Qwen-VL, Cambrian-8B, Molmo-7B, UNITERs, ViLT) & Zero-shot and finetuned (NLVR2→MaRVL), direct or chain-of-thought prompting, accuracy across languages \\

\bottomrule
\end{tabularx}

\caption{Experimental setup overview of multilingual reasoning papers (Part~2).}

\label{tab:multilingual-reasoning}

\end{table}

\subsection{Meta-Reasoning / Self-Evolving Reasoning}
\label{reason:XII}
\begin{figure}[htbp]
\centering
\definecolor{ink}{RGB}{38,41,54}
\definecolor{inksoft}{RGB}{92,97,115}
\definecolor{cViolet}{RGB}{103,87,140}
\definecolor{cGreen}{RGB}{92,136,85}
\definecolor{cRose}{RGB}{169,79,103}
\begin{tikzpicture}[
 arr/.style={-{Stealth[length=2.2mm]}, draw=cGreen, line width=1pt},
 chainstep/.style={rectangle, draw=cGreen, line width=0.8pt, rounded corners=3pt,
   fill=white, font=\scriptsize, text=ink, minimum width=1.5cm, minimum height=0.6cm},
]
\draw[rounded corners=9pt, draw=cRose,   line width=1pt, fill=cRose!7]   (-5.7,-2.75) rectangle (5.7,3.35);
\draw[rounded corners=7pt, draw=cViolet, line width=1pt, fill=cViolet!7] (-4.7,-1.45) rectangle (4.7,2.30);
\draw[rounded corners=5pt, draw=cGreen,  line width=1pt, fill=cGreen!7]  (-3.5, 0.15) rectangle (3.5,1.55);
\node[font=\small\bfseries, text=cRose,   anchor=north west] at (-5.45,3.12) {OUTER~~$\cdot$~~Meta-Learning};
\node[font=\small\bfseries, text=cViolet, anchor=north west] at (-4.45,2.08) {MIDDLE~~$\cdot$~~Search \& Revision};
\node[font=\small\bfseries, text=cGreen,  anchor=north west] at (-3.25,1.33) {INNER~~$\cdot$~~Reasoning Chain (CoT)};
\node[chainstep] (s1) at (-1.8,0.6) {Step 1};
\node[chainstep] (s2) at ( 0.0,0.6) {Step 2};
\node[chainstep] (s3) at ( 1.8,0.6) {Step N};
\draw[arr] (s1) -- (s2);
\draw[arr, dashed] (s2) -- (s3);
\node[font=\scriptsize, text=inksoft, fill=cGreen!7, inner sep=1pt] at (0.9,0.6) {$\cdots$};
\node[font=\scriptsize, text=inksoft, align=center, text width=8.4cm] at (0,-0.65)
  {MCTS, backtracking, iterative refinement \quad$\cdot$\quad AlphaLLM, ThinkTwice, Mind Evolution};
\node[font=\scriptsize, text=inksoft, align=center, text width=10.6cm] at (0,-2.05)
  {Discovers reasoning structures; refines via meta-judges \quad$\cdot$\quad SELF-DISCOVER, Meta-Rewarding, capacity-matched distillation};
\end{tikzpicture}
\caption{Three nested levels of meta-reasoning. The \emph{inner loop} executes a standard reasoning chain (CoT). The \emph{middle loop} wraps it with search and revision, MCTS, backtracking, and iterative refinement. The \emph{outer loop} discovers reasoning structures, builds reusable templates, matches them to model capacity, and refines via meta-judges.}
\label{fig:meta-reasoning-loops}
\end{figure}
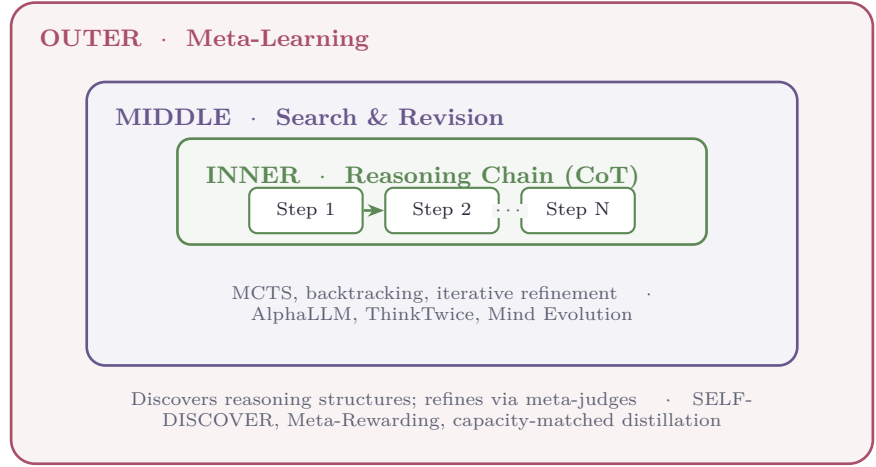

Meta-reasoning treats the model’s \emph{own} thinking as an object to plan, critique, revise, and scale at train- or test-time. Figure~\ref{fig:meta-reasoning-loops} depicts the three nested levels at which this occurs: an inner reasoning chain, a middle search-and-revision loop, and an outer meta-learning loop that discovers structures and refines via meta-judges. Recent work spans (i) discovering/planning reasoning structures, (ii) search-and-revision at inference, (iii) small-data/self-play data creation, (iv) process-aware training signals, (v) exploration dynamics and test-time scaling, (vi) diagnostics and benchmarks, (vii) memory and multimodal self-evolution, (viii) meta-judging and latent control, and (ix) field outlook. Tables~\ref{tab:meta-reasoning-1},~\ref{tab:meta-reasoning-2}, and~\ref{tab:meta-reasoning-3} compile the experimental setups.

\subsubsection{Discovering Structures \& Meta-Planning}

Recent work in meta-reasoning explores how LLMs can uncover and use latent structures to improve complex reasoning. SELF-DISCOVER by ~\citep{zhou2024self} introduces a framework where LLMs autonomously compose task-specific reasoning structures from atomic modules (e.g., critical thinking, subgoal decomposition). It outperforms CoT and self-consistency baselines across challenging benchmarks while requiring far less compute. Eurus by ~\citep{yuan2024advancing} uses interaction data to build structured preference trees, aligned through stepwise reward functions that enable fine-grained reasoning matching user expectations. MPO (Meta-Planner Optimization) from ~\citep{xiong2025mpo} introduces feedback-refined meta-plans that guide agent behavior across tasks. These plans reduce hallucination and stabilize long-horizon reasoning. ~\citep{wang2024understanding} view language models as ''aggregators of reasoning paths,'' formalizing reasoning as the weighted aggregation of random walk paths over knowledge graphs or CoT steps. Training on these distributions enhances logical and multi-step reasoning performance. ~\citep{marjanovic2025deepseek} present Thoughtology, which systematically categorizes reasoning building blocks in DeepSeek-R1, uncovering trade-offs between sequence length, safety, and reliability during task composition.

\subsubsection{Test-Time Search, Revision \& Recombination}
A growing body of work treats inference as an active search process that incorporates revision, recombination, and verification to improve reliability without additional supervision. AlphaLLM by ~\citep{tian2024toward} adapts Monte Carlo Tree Search (MCTS) for test-time self-improvement without labels, operating in an imagination, search, criticize loop. Models refine answers through simulation and evaluation. THOUGHTSCULPT by ~\citep{chi2024thoughtsculpt} augments MCTS with ``revision actions'' that let the model correct intermediate steps, enabling adaptive course correction during reasoning. Mind Evolution by ~\citep{lee2025evolving} introduces an evolutionary search strategy that generates, recombines, and globally refines candidate solutions in the latent space using a solution evaluator. It operates over complete solutions rather than single steps, improving performance on complex planning benchmarks without formal solvers. ThinkTwice by ~\citep{tian2025think} introduces multi-round test-time reasoning, where iterative refinement across thinking loops repeatedly revises answers through additional computation. Step-Back Prompting by ~\citep{yang2025step} equips LLMs with a learned policy for self-backtracking, allowing them to determine when and where to revert previous steps. It yields over 40\% accuracy gains on complex multi-hop reasoning tasks. ~\citep{geiping2025scaling} scale test-time compute through recurrent latent reasoning (Depth-Unrolled Reasoning), unrolling the depth of the reasoning process to improve both depth and reuse of intermediate representations. ~\citep{drori2025diverse} combine multiple models, sampling strategies, and verifiers (e.g., Lean for formal proofs, code execution for ARC puzzles) at inference time, aggregating outputs through rejection sampling, ensemble voting, and meta-learning for cross-benchmark improvements.

\subsubsection{Data Creation \& Capacity-Matched Distillation}
~\citep{ding2024unleashing} introduce ScaleQuest, a fully synthetic data generation pipeline that uses small (7B-scale) models to generate diverse mathematical reasoning problems from scratch. A two-stage tuning process, Question Fine-Tuning (QFT) followed by Question Preference Optimization (QPO), yields a 1M-scale dataset that rivals existing open datasets across in- and out-of-domain reasoning benchmarks. ~\citep{ye2025limo} propose LIMO, showing that fewer than 1,000 hand-curated samples can match the performance of much larger datasets, a result that highlights data selection over scale. ~\citep{li2025small} observe that models with $\leq$3B parameters struggle with long-chain-of-thought distillation. They propose Mix Distillation, which interleaves short and long reasoning chains during training to align with model capacity, improving both convergence and generalization.

\subsubsection{Process-Aware Training Signals}
\citep{chen2024reverse} introduce REVTHINK, a multi-task framework that integrates both forward and reverse reasoning into training. The student model learns to reason step-by-step from question to answer, formulate a backward question from the answer, and then solve it via backward reasoning. This bidirectional learning enables consistency checks and enhances generalization, outperforming unidirectional distillation methods by up to 13.5\% on average across 12 reasoning tasks. ~\citep{lin2024critical} propose identifying critical tokens, i.e., those most essential to final outcomes, and optimizing contrastive critical DPO (cDPO) objectives that highlight and contrast these pivotal steps within reasoning chains, sharpening the model’s attention to what matters most during decision-making. ~\citep{zeng2024b} develop B-STaR, a self-improving framework that balances exploration and exploitation across training iterations. By dynamically reweighting learning examples based on difficulty and novelty, B-STaR produces stable improvements and avoids overfitting to simplistic or repetitive reasoning patterns.

\subsubsection{Exploration Dynamics \& Test-Time Scaling}
LLM reasoning depends on how exploration is governed and how test-time compute is allocated. ~\citep{levy2024same} demonstrate that even before hitting context length limits, longer inputs subtly degrade reasoning quality, pointing to hidden sensitivities in sequence handling. ~\citep{pan2025large} show that large models often rely too heavily on uncertainty signals and fail to robustly explore open-ended tasks, leading to brittle reasoning. ~\citep{wang2025thoughts} propose TIP (Thought-switching Inhibition Penalty), a decoding strategy that discourages rapid shifts between reasoning threads, promoting deeper and more coherent exploration. ~\citep{zhang2025and} survey test-time scaling techniques, analyzing how chain-of-thought sampling, reranking, and verification allocate additional inference-time compute to boost reasoning without retraining.

\subsubsection{Diagnostics, Benchmarks \& Stability Metrics}
 \citep{davidson2024self} show no robust \emph{self-recognition}; models choose plausible answers over authorship. 
~\citep{hosseini2024not} design compositional problem pairs to reveal second-hop failures. 
~\citep{liu2024your} propose \emph{G-Pass@k} and \emph{LiveMathBench} to measure stable vs.\ peak performance. 
~\citep{zeng2024mr} release \emph{MR-Ben} for system-2 style error detection/verification. 
\emph{Takeaway:} Process-centric diagnostics (stability, second-hop, verification) catch weaknesses missed by the outcome metrics.

\subsubsection{Memory \& Multimodal Self-Evolving}

\citep{kang2025lm2} introduce LM2, a Transformer enhanced with a persistent, gated cross-attention memory module. This module decouples memory storage and retrieval from immediate token processing, enabling robust long-context understanding, multi-hop inference, and numerical reasoning. The design maintains general task performance while excelling at memory-intensive benchmarks like BABILong and MMLU, overcoming typical trade-offs in memory-augmented models. ~\citep{liu2024diving} present MSTaR, a framework for self-evolving multimodal reasoning. It discovers and tunes reasoning recipes, including prompt engineering, reward signals, and training curricula, without relying on human annotations. MSTaR demonstrates strong performance across vision-language benchmarks by evolving task-specific agents in a closed-loop setup.

\subsubsection{Meta-Judging, Alignment \& Latent Control}

\citep{wu2024meta} introduce Meta-Rewarding, a self-improvement framework in which language models iteratively refine both their outputs and their judgment processes by evaluating their own evaluations. The model plays three roles: actor, judge, and meta-judge, enabling it to construct reward signals for both responses and judgments without any human supervision. This structure strengthens the reward model's reliability and mitigates common issues like reward hacking and length bias. ~\citep{galichin2025have} employ sparse autoencoders to extract interpretable latent features from LLMs and demonstrate how steering these features, rather than surface-level prompts, can significantly enhance reasoning quality. Their method supports finer control of internal reasoning dynamics. ~\citep{wu2025effectively} present Thinking Intervention, a mechanism that allows models to inject or replace tokens mid-chain, modifying the reasoning trajectory itself. This token-level editing enhances instruction adherence, logical depth, and safety, offering a granular intervention point beyond input/output prompts.

\subsubsection{Summary \& Outlook}
Meta-reasoning advances by (1) \emph{proceduralizing} thought (plans, MCTS, backtracking, recombination), (2) \emph{targeting the process} (critical tokens, reverse steps, stability metrics), and (3) \emph{scaling wisely} (capacity-matched data, test-time compute, memory). Open directions include unifying latent-space search with verifiable edits, standardizing process-first evaluation, and building controllers that jointly optimize exploration depth, compute, and safety across languages and modalities.

\begin{table}[htbp]
\centering
\small
\begin{tabularx}{\textwidth}{ >{\raggedright\arraybackslash}p{0.2\textwidth} >{\raggedright\arraybackslash\hsize=0.8\hsize\linewidth=\hsize}X >{\raggedright\arraybackslash\hsize=0.8\hsize\linewidth=\hsize}X >{\raggedright\arraybackslash\hsize=1.4\hsize\linewidth=\hsize}X }

\toprule
\textbf{Author} & \textbf{Datasets} & \textbf{Models} & \textbf{Evaluation Setup} \\
\midrule

\multicolumn{4}{l}{\textbf{Discovering Structures \& Meta-Planning}} \\

Zhou et al.~\citep{zhou2024self} & BBH, T4D, MATH & GPT-4, GPT-3.5-turbo, PaLM 2-L, Llama2-70B & Zero-shot accuracy; baselines: Direct, CoT, Plan-and-Solve, Self-Consistency, Majority/Best of RMs \\

Yuan et al.~\citep{yuan2024advancing} & ULTRAINTERACT (Math, Coding, Logic) & EURUS-7B-SFT, EURUS-70B-SFT & Single-turn: HumanEval, MBPP, LeetCode; Multi-turn: MINT (Coding and math), Reward modeling \\

Xiong et al.~\citep{xiong2025mpo} & ScienceWorld (textual science experiments), ALFWorld (embodied household tasks), WebShop (online web navigation) & GPT-4o, GPT-4o-mini, Llama-3.1-8B/70B-Instruct, Qwen2.5-7B-Instruct; variants with Reflexion, MPO, SFT, ETO, KnowAgent, WKM & Average reward across seen/unseen test sets; meta plans generated zero-shot (with 1-shot in-context examples for task trajectories) \\

Wang et al.~\citep{wang2024understanding} & GSM8K, AQUA, SVAMP (math word problems with CoT), StrategyQA (multihop QA), LogicalDeduction (logical reasoning)& Gemma 2B, Yi 6B, Llama 2 7B/13B; variants trained with random-walk pretraining + SFT & Average accuracy on test sets; models pre-trained on random walk paths (M=500 steps) then supervised fine-tuned (SFT) \\

Patel et al.~\citep{marjanovic2025deepseek} & AIME-24, MATH500, GSM8k, Multiplications (k×k) & DeepSeek-R1 & Two studies: (1) effect of thought length on performance, (2) token counts in correct vs incorrect thoughts \\

\midrule
\multicolumn{4}{l}{\textbf{Test-Time Search, Revision \& Recombination}} \\

Tian et al.~\citep{tian2024toward} & GSM8K, MATH (subset) & ALPHALLM (Llama-2-70B / WizardMath-70B) & Final-answer accuracy, avg rollouts; self-improvement via $\eta$MCTS; CoT prompting; termination: line/formula rules \\

Chi et al.~\citep{chi2024thoughtsculpt} & Story Outlines, Mini-Crosswords, CommonGen-Hard & THOUGHTSCULPT (GPT-3.5 / GPT-4) & Node evaluation (holistic/itemized), child generation via feedback, MCTS or DFS search \\

Lee et al.~\citep{lee2025evolving} & TravelPlanner, Trip Planning, Meeting Planning & Mind Evolution (Gemini 1.5 Flash / Pro) & Success rate (fully solved instances), LLM calls, input/output tokens, API cost; baselines: 1-Pass, Best-of-N, Sequential-Revision+; two-stage Flash → Pro for unsolved instances \\

Tian et al.~\citep{tian2025think} & AIME 2024, MATH-500, GPQA-Diamond, LiveCodeBench & DeepSeek-R1, QwQ-32B, AM-Distill-Qwen-32B/7B & Pass@1 accuracy; multi-round thinking (Rounds 1–4); token limit 32,768; sampling: temp 0.6, top-p 0.95; metrics: global average accuracy \\

Yang et al.~\citep{yang2025step} & Countdown & Llama3.2-1B / Llama3.2-3B & Accuracy on Seen/ New Targets; greedy and beam search baselines \\

Drori et al.~\citep{drori2025diverse} & IMO (2006–2024), ARC (400), HLE (100) & o1, o3-mini high & Automatic verification (math/code), consensus (HLE), success rate/accuracy \\

\bottomrule
\end{tabularx}
\caption{Experimental setup overview of meta/self-evolving reasoning papers (Part~1).}
\label{tab:meta-reasoning-1}
\end{table}

\begin{table}[htbp]
\centering
\small
\begin{tabularx}{\textwidth}{ >{\raggedright\arraybackslash}p{0.12\textwidth} >{\raggedright\arraybackslash\hsize=0.85\hsize\linewidth=\hsize}X >{\raggedright\arraybackslash\hsize=1\hsize\linewidth=\hsize}X >{\raggedright\arraybackslash\hsize=1.15\hsize\linewidth=\hsize}X }

\toprule
\textbf{Author} & \textbf{Datasets} & \textbf{Models} & \textbf{Evaluation Setup} \\

\midrule
\multicolumn{4}{l}{\textbf{Data Creation \& Capacity-Matched Distillation}} \\

Ding et al.~\citep{ding2024unleashing} & GSM8K, MATH, College Math, Olympiad Bench & Mistral-7B, Llama3-8B, DeepSeekMath-7B, Qwen2-Math-7B & SFT on 1M synthetic Q/A; CoT; zero-shot pass@1 \\

Ye et al.~\citep{ye2025limo} & AIME24, MATH500, AMC23, OlympiadBench, CHMath, Gaokao, Kaoyan, GradeSchool, Minerva, GPQ & LIMO (800 curated), Qwen2.5-32B-Instruct, OpenAI-o1-preview, QwQ-32B-Preview & Zero-shot pass@1; in/out-of-domain; reasoning chain quality \\

Li et al.~\citep{li2025small} & MATH, GSM8K, AMC, AIME, OlympiadBench & Qwen/Llama (0.5B--70B) & Short/long CoT; small/large teacher; Mix Distillation \\

\midrule
\multicolumn{4}{l}{\textbf{Process-Aware Training Signals}} \\

Chen et al.~\citep{chen2024reverse} & Math: MATH, GSM8K; Commonsense: SQA, CSQA, ARC; Tabular: TabMWP; NLI: ANLI; Logic: Date & Gemini-1.5-Pro-001 (Teacher), Mistral-7B-Instruct-v0.3, Gemma-7B-Instruct (Students) & Zero-shot; SKD; Distill Step-by-Step; Q/A Augmentation \\

Lin et al.~\citep{lin2024critical} & GSM8K, MATH500 & Llama-3-8B/70B, DeepSeek-7B & cDPO vs SFT/DPO/TokenDPO/RPO (GSM8K 77.2, MATH 33.4) \\

Zeng et al.~\citep{zeng2024b} & GSM8K, MATH, APPS, ARC-C & Mistral-7B, Llama-3-8B & B-STAR vs SFT/Rest-EM/Iter.~RFT; highest P@1 \\

\midrule
\multicolumn{4}{l}{\textbf{Exploration Dynamics \& Test-Time Scaling}} \\

Levy et al.~\citep{levy2024same} & FlenQA (extractive QA with long contexts) & GPT-3.5, GPT-4, Gemini-Pro, Mistral Medium & Zero-shot reasoning; variable key-paragraph positions \\

Pan et al.~\citep{pan2025large} & Little Alchemy 2 & GPT-4o, o1, LLaMA3.1-8B, LLaMA3.1-70B, DeepSeek-R1 & Exploration over 500 trials per model \\

Wang et al.~\citep{wang2025thoughts} & AIME 2022--23, AIME2024, MATH500-Hard, GPQA Diamond & QwQ-32B-Preview, R1-Distill-Qwen-32B & Pass@k; weighted underthinking score; thought-switching analysis \\

\midrule
\multicolumn{4}{l}{\textbf{Diagnostics, Benchmarks \& Stability Metrics}} \\

Davidson et al.~\citep{davidson2024self} & Self-Generated Security Questions & LLaMA 3-8B, LLaMA 3-70B, Mistral 8x22B & Verdict models select best from $n$=2,3,5 \\

Hosseini et al.~\citep{hosseini2024not} & GSM8K (original, X-substituted, compositional) & GPT-4o/mini, LLAMA3, Phi 2/3, Gemini & 8-shot pass@1; pretrained vs instruction-tuned; compositional GSM \\

Liu et al.~\citep{liu2024your} & LiveMathBench, MATH500-L5, AIME2024-45, AIME2025 & Llama-3.1/3.3, Yi-1.5-34B, Gemma-2-27B, Qwen2.5 (7B--72B) & Greedy accuracy; G-Pass@16 and mG-Pass@16 for consistency \\

Zeng et al.~\citep{zeng2024mr} & LiveCodeBench, GPQA-Diamond, AIME 2024, MATH-500 & DeepSeek-R1, QwQ-32B, DeepSeek-R1-Distill-Qwen (32B/7B), AM-Distill-Qwen-32B & Multi-round thinking (1--4 rounds); pass@1 avg accuracy \\

\bottomrule
\end{tabularx}
\caption{Experimental setup overview of meta/self-evolving reasoning papers (Part~2).}
\label{tab:meta-reasoning-2}
\end{table}

\begin{table}[htbp]
\centering
\small
\begin{tabularx}{\textwidth}{ >{\raggedright\arraybackslash}p{0.2\textwidth} >{\raggedright\arraybackslash\hsize=0.8\hsize\linewidth=\hsize}X >{\raggedright\arraybackslash\hsize=0.8\hsize\linewidth=\hsize}X >{\raggedright\arraybackslash\hsize=1.4\hsize\linewidth=\hsize}X }

\toprule
\textbf{Author} & \textbf{Datasets} & \textbf{Models} & \textbf{Evaluation Setup} \\

\midrule
\multicolumn{4}{l}{\textbf{Memory \& Multimodal Self-Evolving}} \\

Kang et al.~\citep{kang2025lm2} & BABILong (extended bAbI) & LM2-1.7B, vanilla-LLaMA-1.7B, RMT-1.7B, LLaMA-3.2-1.2B, LLaMA-3.2-1.2B & Memory reasoning evaluation across 0K–128K context lengths measuring accuracy on single-step, multi-step, relation tracking, and query tasks \\

Liu et al.~\citep{liu2024diving} & MathV360K, MathVista (testmini) & MiniCPM-V-2.5 (8B), Phi-3.5-Vision (4B), InternVL-2 (2B) & Evaluated with exact-match accuracy on in-domain and OOD multimodal reasoning tasks \\

\midrule
\multicolumn{4}{l}{\textbf{Meta-Judging, Alignment \& Latent Control}} \\

Wu et al.~\citep{wu2024meta} & AlpacaEval 2, Arena-Hard, MT-Bench & Llama-3-8B-Instruct (seed) with Meta-Rewarding iterations & Actor evaluated via GPT-4 judge benchmarks; judge evaluated using Spearman correlation with human preferences \\

Galichin et al.~\citep{galichin2025have} & AIME 2024, MATH-500, GPQA-Diamond & DeepSeek-R1-LLaMA-8B & Evaluated using ReasonScore, sparsity (L0), reconstruction variance, and maj@4 accuracy\\

Wu et al.~\citep{wu2025effectively} & SEP Benchmark, XSTest, SORRY-Bench & R1-Qwen (7B/14B/32B), QwQ-32B, o3-mini, GPT-4o, GPT-4o-mini & Evaluated robustness and utility using LLM-as-a-judge\\

\bottomrule
\end{tabularx}

\caption{Experimental setup overview of meta/self-evolving reasoning papers (Part~3).}

\label{tab:meta-reasoning-3}
\end{table}

\subsection{Social and Cognitive Reasoning}
\label{reason:XIII}
\begin{figure}[htbp]
\centering
\definecolor{ink}{RGB}{38,41,54}
\definecolor{inksoft}{RGB}{92,97,115}
\definecolor{hair}{RGB}{200,205,216}
\definecolor{cBlue}{RGB}{60,87,150}
\definecolor{cAmber}{RGB}{168,120,43}
\definecolor{cGreen}{RGB}{92,136,85}
\definecolor{cViolet}{RGB}{103,87,140}
\definecolor{cRose}{RGB}{169,79,103}
\scalebox{0.8}{
\begin{tikzpicture}[
 node distance=0.7cm,
 box/.style={rectangle, draw=#1, line width=0.9pt, rounded corners=5pt, fill=#1!8,
   minimum height=0.85cm, align=center, font=\small, text width=2.8cm, text=ink},
 dec/.style={diamond, draw=cAmber, line width=0.9pt, fill=cAmber!10, aspect=1.4,
   align=center, font=\scriptsize, text=ink, inner sep=1pt},
 arr/.style={-{Stealth[length=2.4mm]}, draw=inksoft, line width=1pt},
 elab/.style={font=\scriptsize, text=inksoft, fill=white, inner sep=1.5pt},
]
\node[box=cBlue, text width=3.2cm] (input) {\textcolor{cBlue}{\textbf{Social Scenario}}\\``Is this action fair?''};
\node[dec, below=0.7cm of input] (split) {Mode?};
\node[box=cGreen, below left=1.1cm and 1.2cm of split]  (da)  {\textcolor{cGreen}{\textbf{Direct Answer}}\\{\scriptsize Low transparency}\\{\scriptsize Moderate bias}};
\node[box=cRose,  below right=1.1cm and 1.2cm of split] (cot) {\textcolor{cRose}{\textbf{Chain-of-Thought}}\\{\scriptsize High transparency}\\{\scriptsize \textcolor{cRose}{Inflated confidence}}\\{\scriptsize \textcolor{cRose}{Amplified bias}}};
\node[box=cViolet, below=3.3cm of split, text width=3.2cm] (out) {\textcolor{cViolet}{\textbf{Social Output}}};
\draw[arr] (input) -- (split);
\draw[arr] (split) -| node[elab, pos=0.25] {direct} (da);
\draw[arr] (split) -| node[elab, pos=0.25] {CoT}    (cot);
\draw[arr] (da)  |- (out);
\draw[arr, draw=cRose] (cot) |- (out);
\node[rectangle, draw=cRose, line width=0.8pt, dashed, rounded corners=4pt, fill=cRose!7,
  font=\scriptsize, text width=2.2cm, align=center, text=ink, inner sep=4pt, right=0.5cm of cot] (warn)
  {\textcolor{cRose}{\textbf{Paradox}}\\more steps can\\entrench bias};
\draw[-{Stealth[length=1.8mm]}, draw=cRose, line width=0.7pt, dashed] (warn) -- (cot);
\node[rectangle, draw=cBlue, line width=0.8pt, rounded corners=5pt, fill=cBlue!7,
  font=\scriptsize, text width=8.2cm, align=center, text=ink, inner sep=6pt, below=0.5cm of out] (mit)
  {\textcolor{cBlue}{\textbf{Mitigations}}~~ Bias-aware eval (BBQ) $\mid$ Cognitive priming $\mid$ Process judges (JudgeLRM) $\mid$ Multi-LLM fusion};
\draw[-{Stealth[length=2mm]}, draw=hair, line width=0.8pt, dashed] (out) -- (mit);
\end{tikzpicture}
}
\caption{The social reasoning paradox. Direct answering yields moderate bias with low transparency. Chain-of-thought reasoning increases transparency but can simultaneously inflate confidence and amplify social biases~\citep{wu2025does}. Mitigations include bias-aware benchmarks, cognitive priming, process-level judges, and multi-model fusion.}
\label{fig:social-reasoning-paradox}
\end{figure}
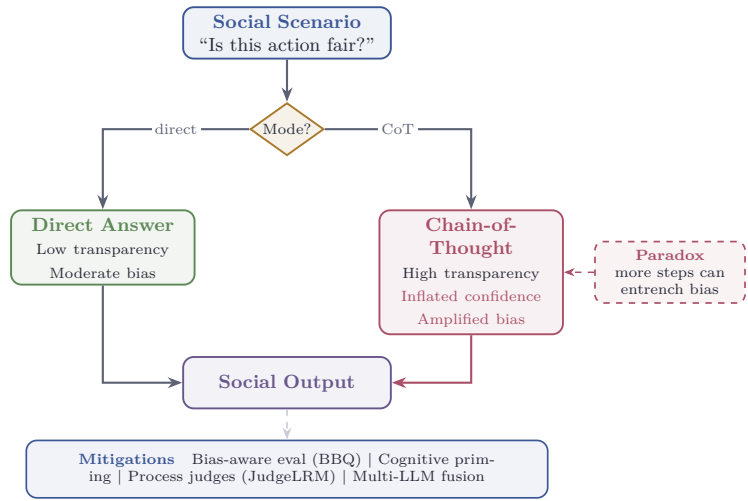

Social and cognitive reasoning requires models to handle intentions, beliefs, emotions, and social dynamics rather than only factual information. Unlike mathematical or logical reasoning that operates within formal systems, social cognition depends on context-dependent norms, cultural variation, theory of mind, and tolerance for the ambiguity that pervades human interaction. As LLMs are deployed in educational, therapeutic, collaborative, and decision-making settings, these capabilities become necessary for effective and safe operation.
The challenge extends to ensuring that models can participate in human social structures without introducing harmful biases, overconfidence, or misaligned judgments. This requires metacognitive capacity: understanding what people think, how and why they think it, and how social context shapes their reasoning. Figure~\ref{fig:social-reasoning-paradox} highlights a key tension in this space: chain-of-thought reasoning increases transparency but can simultaneously inflate confidence and amplify social biases, creating a paradox that demands bias-aware evaluation and process-level judges. Recent research has approached these problems through several interconnected areas outlined below. Tables~\ref{tab:social-cognitive-1} and~\ref{tab:social-cognitive-2} summarize the experimental setups.

\subsubsection{Theory of Mind and Social Cognitive Benchmarks}
Researchers are now systematically probing whether language models can reason about beliefs, desires, intentions, and emotions. BigToM~\citep{gandhi2023understanding} offers a procedurally generated benchmark for theory of mind evaluation at scale. Its results show that while GPT-4 exhibits human-like patterns on many tasks, performance is far from consistent, suggesting that current models lack stable representations of mental state reasoning.
Amirizaniani et al.~\citep{amirizaniani2024can} move beyond controlled tasks by using open-ended Change My View prompts, testing whether models can infer intentions and emotions in dynamic, conversational contexts. Despite explicit prompting strategies, they find persistent failures, reinforcing the idea that models do not robustly internalize social perspective-taking.
To address the gap between surface-level performance and deeper psychological grounding, propose a Sentient Agent judge that explicitly tracks emotion trajectories and inner thought processes. Their framework aligns with established psychological scales and surfaces empathy deficits that are otherwise hidden by task-level accuracy, pushing evaluation closer to real measures of social sensitivity.
Other efforts frame social cognition in more formal decision-making settings.~\citep{leng2023llm} introduce SUVA, which maps natural language utterances to game-theoretic scenarios of fairness and reciprocity. Their findings show that prosocial tendencies vary systematically with model size and training methods, highlighting that social reasoning is shaped by design choices rather than emerging automatically with scale.
~\citep{zhu2024can} benchmark across four cognitive tasks and nine datasets, showing that context comprehension, particularly under in-context learning and quantization, remains brittle. ~\citep{huang2024olympicarena} extend the evaluation frontier with OlympicArena, 11,163 real Olympiad-style problems across multilingual and multimodal contexts. By implementing process-level reasoning evaluation, they reveal that even GPT-4o struggles with stable higher-order reasoning, especially when tasks require maintaining coherent social or cognitive trajectories.

\subsubsection{Multimodal Social Understanding and Bias}
Social reasoning increasingly requires integration across multiple modalities, as human social interaction relies heavily on visual cues, contextual information, and cultural markers that extend beyond textual communication.
SocialGPT~\citep{li2024socialgpt} converts image-based representations of social relationships into textual narratives, allowing vision-language models to express social reasoning in an interpretable and analyzable form. Using vision foundation models in combination with LLMs and optimizing long prompt sequences through Generalized Self-Play Optimization (GSPO), SocialGPT shows that integrating visual information into textual reasoning pathways can improve both transparency and explanatory power in social cognition tasks.
As models expand into multimodal spaces, new vulnerabilities emerge. DALL-Eval~\citep{cho2023dall} systematically diagnoses visual reasoning capabilities and demographic biases in text-to-image systems, showing that even state-of-the-art models struggle with basic visual reasoning tasks such as counting and spatial relations while simultaneously encoding and amplifying social biases in gender and skin-tone. Multimodal reasoning carries the challenges of text-based reasoning and adds new dimensions of fairness and interpretability that require active management.

\subsubsection{Confidence Calibration and Bias in Social Reasoning}
\citep{fu2025multiple} show that chain-of-thought reasoning in multiple-choice scenarios systematically inflates model confidence regardless of correctness, indicating that self-reported confidence cannot be treated as a reliable proxy for reasoning quality in socially consequential contexts. ~\citep{goel2025great} highlight how similarity across model families leads to correlated errors, meaning that ensembles or AI-on-AI evaluation frameworks may reinforce rather than mitigate biases, an especially important consideration for automated systems intended to assess or moderate social reasoning.
~\citep{wu2025does} demonstrate that chain-of-thought reasoning itself can amplify social biases, even when logical validity is preserved. Using benchmarks such as BBQ, they show that explicit reasoning steps may inadvertently surface or entrench prejudicial associations, and that instruction tuning exerts uneven effects on bias expression depending on the type of social reasoning task.
These findings reveal a paradox: mechanisms like chain-of-thought, which enhance transparency, can simultaneously inflate misplaced confidence and amplify bias. Bias-aware evaluation frameworks, diverse oversight mechanisms, and calibration strategies that separate reasoning validity from confidence expression remain open needs.

\subsubsection{Cognitive Training and Evaluation Methods}
Recent research converges on two directions: embedding cognitive strategies directly into model behavior and designing evaluators that diagnose reasoning quality at a process level.
~\citep{gandhi2025cognitive} identify four key cognitive habits (verification, backtracking, subgoal decomposition, and backward chaining) that consistently drive self-improvement in reasoning. Their behavioral priming experiments show notable gains in LLaMA models relative to Qwen under identical conditions, suggesting that explicit cognitive behavior training can yield transferable reasoning benefits. ~\citep{chen2025judgelrm} develop JudgeLRM, reinforcement-trained reasoning judges that outperform supervised fine-tuning approaches by rewarding outcome-aligned but process-sensitive evaluations, setting a new standard for assessing deep reasoning capabilities.
Beyond behavior-level interventions, researchers are exploring lightweight mechanisms for dynamically shaping reasoning during generation. ~\citep{jin2025well} propose Adaptive Injection Decoding, which extends reasoning chains in real time by inserting short trigger phrases, eliminating the need for rigid prompt engineering while maintaining coherence. Similarly, ~\citep{aytes2025sketch} introduces Sketch-of-Thought, which routes among cognitive sketching strategies such as concept chaining, chunked symbolism, and expert lexicons, reducing token usage by up to 76\% without accuracy loss. MLKF (Multi-LLM Knowledge Fusion)~\citep{liu2024two} combines outputs from multiple LLMs to overcome shallow chain-of-thought reasoning and individual model blind spots, highlighting that collective reasoning strategies can augment both depth and breadth of cognitive inference.

\subsubsection{Integration and Future Directions}
Social and cognitive reasoning research has converged around three fundamental approaches: process-first evaluation through sophisticated benchmarks, theory of mind probes, and psychological grounding that reveals how models develop social understanding rather than just measuring outcomes; governance of confidence and bias through recognition that reasoning processes can miscalibrate confidence and amplify social biases, requiring diversified oversight and bias-aware evaluation frameworks; and cognitive control at test-time through sketching techniques, adaptive decoding methods, and multi-model fusion approaches that enhance reasoning efficiency while maintaining social cognitive capabilities.
Open challenges include developing standardized higher-order theory of mind benchmarks, implementing multimodal bias mitigation strategies that preserve social reasoning competence, and creating evaluation frameworks that combine psychological grounding with process-level auditing.

\begin{table}[htbp]
\centering
\small
\begin{tabularx}{\textwidth}{ >{\raggedright\arraybackslash}p{0.2\textwidth} >{\raggedright\arraybackslash\hsize=0.8\hsize\linewidth=\hsize}X >{\raggedright\arraybackslash\hsize=0.8\hsize\linewidth=\hsize}X >{\raggedright\arraybackslash\hsize=1.4\hsize\linewidth=\hsize}X }

\toprule
\textbf{Author} & \textbf{Datasets} & \textbf{Models} & \textbf{Evaluation Setup} \\
\midrule

\multicolumn{4}{l}{\textbf{Theory of Mind and Social Cognitive Benchmarks}} \\

Gandhi et al.~\citep{gandhi2023understanding} & BigToM & text-davinci-003, GPT-3.5-turbo, GPT-4-0314, Claude-v1.3, LLaMA-65B & Evaluated using 0-shot, 0-shot-CoT, 1-shot, and 1-shot-CoT prompts on belief and action inference tasks\\

Amirizaniani et al.~\citep{amirizaniani2024can} & Reddit r/ChangeMyView (845 posts with human and LLM responses)& Zephyr-7B, Llama2-Chat-13B, GPT-4 & Human-in-the-Loop evaluation by annotators \\

Zhang et al.~\citep{zhang2025sentient} & 100 supportive dialogue scenarios across 8 topics & GPT-4o, OpenAI-o1, DeepSeek-V3/R1, Claude-3.7, Gemini-2.5, Llama-3.3-70B, Qwen-2.5-72B & Evaluated using SAGE (Sentient-Agent-as-Judge) measuring Emotion scores correlated with BLRI relationship metrics \\

Leng et al.~\citep{leng2023llm} & SUVA-based simulated social interaction tasks & GPT-3.5, GPT-4, LLaMA-2 (13B, 70B), LLaMA-3 (8B, 70B), Mistral-7B, Mixtral-8x7B & Models run thousands of times with CoT and temperature 0.2 to analyze distributional preferences \\

Zhu et al.~\citep{zhu2024can} & 9 contextual understanding datasets & OPT (125M–2.7B), LLaMA (7B–30B), GPT-3.5, T5 (fine-tuned baseline) & In-Context Learning evaluation with few-shot demonstrations \\

Huang et al.~\citep{huang2024olympicarena} & OlympicArena & GPT-3.5, GPT-4, GPT-4o, Claude-3 Sonnet, Qwen-7B/32B & Zero-shot evaluation under three settings (multimodal, image-caption, text-only) \\

\midrule
\multicolumn{4}{l}{\textbf{Multimodal Social Understanding and Bias}} \\

Li et al.~\citep{li2024socialgpt} & PIPA (16 social relations) and PISC (6 social relations) image datasets & GPT-3.5, Vicuna-7B/13B, LLaMA2-7B/13B with VFMs (SAM, BLIP-2) and SocialGPT framework & Zero-shot social relation classification using generated social stories from images \\

Cho et al.~\citep{cho2023dall} & PAINTSKILLS (diagnostic dataset for object recognition, counting, and spatial relations) & DALL-ESmall, minDALL-E, Stable Diffusion v1.4, Karlo &Models fine-tuned and evaluated using DETR-based object detection to measure visual reasoning accuracy \\

\bottomrule
\end{tabularx}
\caption{Experimental setup overview of social and cognitive reasoning papers (Part~1).}
\label{tab:social-cognitive-1}

\end{table}

\begin{table}[htbp]
\centering
\small
\begin{tabularx}{\textwidth}{ >{\raggedright\arraybackslash}p{0.2\textwidth} >{\raggedright\arraybackslash\hsize=0.8\hsize\linewidth=\hsize}X >{\raggedright\arraybackslash\hsize=0.8\hsize\linewidth=\hsize}X >{\raggedright\arraybackslash\hsize=1.4\hsize\linewidth=\hsize}X }

\toprule
\textbf{Author} & \textbf{Datasets} & \textbf{Models} & \textbf{Evaluation Setup} \\
\midrule

\multicolumn{4}{l}{\textbf{Confidence Calibration and Bias in Social Reasoning}} \\

Fu et al.~\citep{fu2025multiple} & MMLU (57 subjects, ~15K MCQ questions) & LLaMA-3.1-8B, LLaMA-3.2-11B, Mistral-7B, Gemma-2-9B, Yi-1.5-9B, GPT-4o-mini, GPT-4o & Comparison of direct answering vs Chain-of-Thought prompting\\

Goel et al.~\citep{goel2025great} & MMLU-Pro (8,707 filtered questions), BBH, and 15 NLP classification datasets & Qwen2.5 (7B–72B), LLaMA-3.1/3.2, Gemma-2, Mistral, Phi-2, and others (~130 models total) & LLM-as-a-judge evaluation where models assess free-text answers without ground truth\\

Wu et al.~\citep{wu2025does} & BBQ dataset & Llama-3.1-8B-Instruct, Qwen2.5-32B, Marco-o1, DeepSeek-R1-Distill-Llama-8B, DeepSeek-R1-Distill-Qwen-32B & Zero-shot evaluation with model-specific generation settings on NVIDIA A100 GPUs \\

\midrule
\multicolumn{4}{l}{\textbf{Cognitive Training and Evaluation Methods}} \\

Chen et al.~\citep{chen2025judgelrm} & JudgeLM and PandaLM & Qwen2.5/Qwen3 bases & Pairwise LLM-judge evaluation for open-ended QA \\

Jin et al.~\citep{jin2025well} & MultiArith, GSM8K, StrategyQA & LLaMA-3.1-8B, Mistral-7B-v0.3, Gemma-7B & Zero-shot and Zero-shot-CoT prompting evaluation\\

Aytes et al.~\citep{aytes2025sketch} & Mathematical, Commonsense, Logical & Qwen-2.5 (7B, 14B, 32B), LLaMA-3.1-8B, LLaMA-3.2-11B & Few-shot prompting (temperature 0.5); 3 runs averaged; accuracy(exact match / GPT-4o judge) \\

Liu et al.~\citep{liu2024two} & Laptop14, Restaurant14 (ABSA); CrossNER-Literature & Vicuna-13B, WizardLM-13B (backbone) & Zero-shot evaluation; temperature 0.1, top-p 0.95; metrics: Accuracy / Macro-F1 (ABSA), Micro-F1 (NER)\\

\bottomrule
\end{tabularx}

\caption{Experimental setup overview of social and cognitive reasoning papers (Part~2).}

\label{tab:social-cognitive-2}

\end{table}

\section{Open Challenges and Future Opportunities in LLM Reasoning}

\subsection{Fundamental Limitations Across Reasoning Domains }
LLM reasoning across domains, despite rapid advances, continues to reveal structural limitations. What often appears as sophisticated reasoning is better understood as structured pattern matching. Chain-of-thought prompting, for example, can improve accuracy on symbolic tasks largely by scaffolding answers into multi-step traces, yet this reflects the model’s ability to reproduce the form of reasoning rather than its underlying logic. In mathematics, process-level supervision and latent-control methods push models toward more reliable intermediate steps, but even here, improvements rely on external scaffolds rather than the model’s internalization of genuine schemas.

This fragility is even more acute when tasks extend beyond narrow text-only settings. In multimodal reasoning, models tend to struggle when abstraction, spatial manipulation, or multi-step visual inference is needed. They tend to fall back on linguistic shortcuts and lack perceptual understanding. Commonsense reasoning benchmarks also overestimate progress as models succeed by exploiting artifacts or shallow correlations. Recent multilingual reasoning advances show that reasoning processes can be transferred across languages through alignment and modular routing, but consistency and faithfulness are still unreliable. This is particularly the case for low-resource or typologically distant languages, where the lack of coherence comes from representational mismatches.

Even within language-only domains, structural weaknesses emerge when reasoning requires multi-hop inference or integration across extended contexts. The models handle the first step of reasoning competently. Coherence, however, often breaks down across chains, exposing a reliance on shallow heuristics rather than the flexible synthesis of scattered evidence. Long-context extensions, retrieval augmentation, and memory-inspired designs provide partial solutions, yet models typically underuse available information, and scaling these methods remains computationally costly. Tool-augmented and agentic approaches show promise in modular planning, retrieval, and verification, but integrating these components while balancing efficiency, safety, and reasoning quality remains an open problem.

Today’s LLMs excel at producing locally plausible reasoning steps but remain brittle when tasks demand genuine abstraction, robustness across modalities, or sustained integration over long horizons. Whether the challenge lies in distinguishing imitation from understanding, preserving stability across languages and perceptual inputs, or coherently linking multi-step chains, the underlying limitation is the absence of mechanisms for principled generalization and structural consistency. Overcoming these weaknesses will require moving beyond incremental scaling toward architectures and training paradigms that treat reasoning not as isolated outputs but as processes that must remain reliable across domains and contexts.

\subsection{Critical Research Questions}
\subsubsection{Scale, Emergence, and Architectural Innovation}

\noindent\textit{RQ: What is the relationship between model scale and emergent reasoning abilities?}

\begin{itemize}
\item \textbf{Scale without structure plateaus.} Larger models built on generic corpora often display shallow pattern matching with limited principled inference. Mid-sized models trained with domain-specialized pretraining, carefully filtered corpora, and reinforcement-driven optimization frequently rival or exceed far larger counterparts. Emergence is a product of aligning scale with structure: symbolic corpora, initialization strategies that favor inferential behavior, and objectives that reward process fidelity.

\item \textbf{Reasoning-first foundations outperform generic scale.} Code- and math-centric LLMs demonstrate that curated datasets, long-context architectures, and modular designs can transform efficiency gains into reasoning depth. Mixture-of-Experts frameworks further show that scale can be made selective and efficient, with specialized reasoning modules activated only when needed.

\item \textbf{RL and self-play can replace supervised traces.} Reinforcement learning and self-play reveal that structured reasoning behaviors can emerge even without supervised examples, provided the reward landscape encourages verification, backtracking, and exploration.
\end{itemize}

\subsubsection{Prompting Strategies and Reasoning Elicitation}

\noindent\textit{RQ: How do prompting strategies impact reasoning performance across different domains?}

\begin{itemize}
\item \textbf{Grounded scaffolding over longer chains.} The effectiveness of prompting depends on anchoring reasoning to intermediate artifacts (evolving tables, knowledge graphs, or highlighted input spans) rather than producing longer text. This makes each step transparent and verifiable, improving reliability in domains where grounding and factual consistency matter most. Prompt-free elicitation shows that stepwise reasoning can surface naturally with the right decoding and self-consistency mechanisms, reducing dependence on handcrafted prompts.

\item \textbf{Self-regulating reasoning processes.} Approaches that introduce inner dialogue, selective self-correction, or recursive anticipation show that models benefit from mechanisms that decide when to expand reasoning, when to intervene, and when to stop. This helps avoid both shallow pattern following and unproductive overthinking, making reasoning more efficient across diverse task types.

\item \textbf{Inference-time search and revision.} Treating reasoning as a search process augmented by revision, backtracking, recombination, and verification improves robustness without requiring additional labels.
\end{itemize}

\subsubsection{Tool Integration and Modular Reasoning}

\noindent\textit{RQ: How can we effectively integrate symbolic, neural, and tool-augmented reasoning?}

\begin{itemize}
\item \textbf{From ad-hoc invocation to structured orchestration.} Separating high-level planning from tactical execution allows models to decompose problems, explore alternative action paths, and invoke tools with far greater reliability and efficiency. Modules can be searched, sequenced, and optimized as part of a broader reasoning pipeline.

\item \textbf{Multi-agent collaboration and distillation.} Role specialization, debate, and structured verification create collective intelligence that surpasses what a single model can achieve. Recent distillation methods show these gains can be compressed back into individual models. Unified retrieval-generation architectures blur the boundary between knowledge access and text generation, reducing latency and error propagation by treating retrieval as an integral reasoning subtask.

\item \textbf{Procedural scaffolds as externalized reasoning.} Iterative retrieval frameworks and programmatic scaffolds reframe reasoning as a controlled sequence of state transitions, whether through multi-hop query reformulation, symbolic program execution, or reusable algorithmic abstractions. These scaffolds act as externalized memory and verification layers, grounding neural models in interpretable, verifiable structures that improve generalization across mathematics, coding, and knowledge-intensive QA.
\end{itemize}

\subsection{Evaluation and Safety Challenges}
\begin{figure}[]
 \centering
 \includegraphics[width=\textwidth]{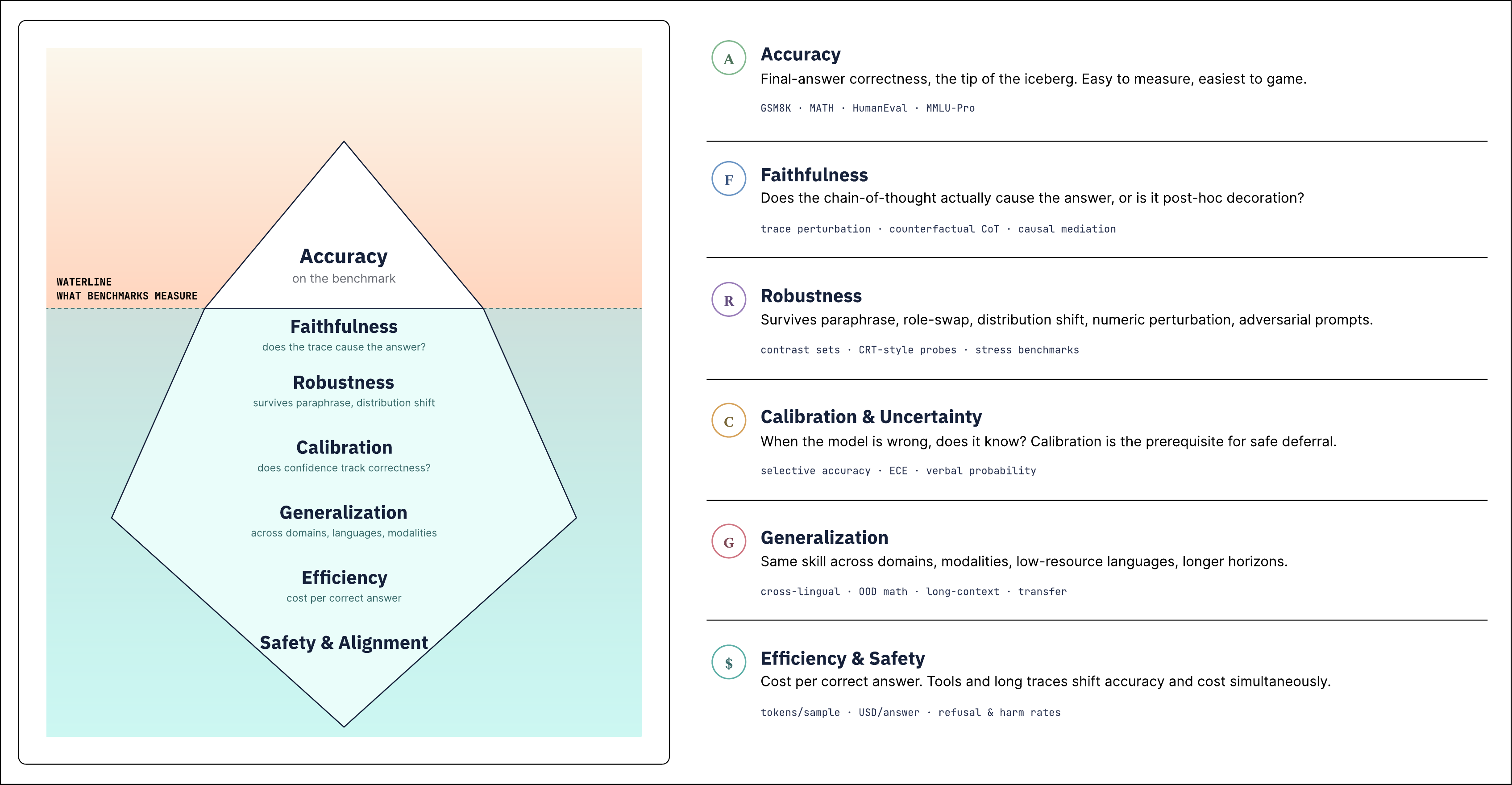}
 \caption{Accuracy is the tip; faithfulness, robustness, calibration, generalization, efficiency, and safety sit below the waterline. Addresses your paper's recurring point that benchmarks overstate progress.
 }
 \label{fig:evaluation_pipeline}
\end{figure}
\subsubsection{Reasoning Process Transparency and Evaluation}
A central challenge for mathematical and general reasoning models is ensuring that their processes are transparent and their abilities evaluated in ways that go beyond surface-level accuracy. Standard benchmarks often conflate pattern exploitation with genuine reasoning, making it difficult to assess whether models truly “understand” or simply memorize correlations. To address this, evaluation has begun to shift toward frameworks that probe intermediate reasoning, robustness to perturbations, and task-specific fidelity. Robustness-focused studies reveal how fragile current systems remain that even small rephrasings, symbolic substitutions, or structural alterations can cause dramatic performance drops, suggesting that many models lack deeper logical generalization. At the opposite extreme, Olympiad-style datasets establish upper-bound tests of capability, demonstrating that frontier models still struggle with high-level abstraction and multi-hop reasoning. Complementing these benchmark-driven approaches, intrinsic analyses such as embedding-trajectory volatility provide diagnostic signals when models encounter out-of-distribution inputs, reframing evaluation as a tool for detecting when reasoning confidence is misplaced.

Beyond general-purpose testbeds, domain-specific frameworks have emerged to capture the epistemic norms of particular fields. In mathematics, specialized datasets now test symbolic rigor, proof verification, and generalization to Olympiad-level difficulty. In programming, execution-centered benchmarks such as LiveCodeBench and SWE-bench prioritize compilation fidelity, bug localization, and repair robustness, offering a more realistic measure of software intelligence than simple completion accuracy. In multimodal reasoning, frameworks evaluate consistency between textual explanations and visual or perceptual evidence, demanding alignment across modalities rather than isolated correctness. These targeted evaluations mitigate overfitting and ensure that progress reflects substantive capabilities valued in practice.
Recent work also emphasizes metacognitive and reflective evaluation, where models categorize problems into skill clusters, critique their own solutions, and refine strategies across iterations. This self-awareness offers a route toward learning-to-learn rather than one-off problem solving. In parallel, interpretability efforts show that externalizing intermediate reasoning into visual artifacts, tables, or structured highlights boosts accuracy on spatial and symbolic tasks and provides human-readable traces of logic. Transparency carries risks, though: chain-of-thought prompting has been shown to inflate unwarranted confidence and amplify social biases. Calibration strategies that separate reasoning quality from confidence expression remain necessary.

\subsubsection{Social, Ethical, and Safety Considerations}
Progress in social intelligence, safety, and cross-linguistic alignment reveals both significant advances and persistent vulnerabilities. Evaluation frameworks need to extend beyond accuracy to include psychological grounding, adversarial resilience, and cultural inclusivity. Three dimensions stand out:

\begin{itemize}
\item \textbf{Social cognition and theory of mind.} A major line of inquiry examines whether models genuinely capture aspects of human social cognition such as beliefs, intentions, and emotions. Benchmark studies show that frontier models can reproduce many signals of theory of mind, yet their reasoning remains brittle, often failing to generalize across dynamic conversational contexts or higher-order inference tasks. Process-oriented evaluations, such as tracking emotional trajectories or mapping linguistic inputs to fairness scenarios, highlight that social reasoning depends heavily on training design choices rather than emerging as a stable capacity. The cumulative evidence suggests that models mimic social perspective without deeply internalizing it, making psychologically grounded benchmarks indispensable for measuring real progress.

\item \textbf{Agentic safety and adversarial resilience.} Ensuring the safety and security of agentic systems has become increasingly urgent. Reasoning-capable safety guards demonstrate the potential to surpass static filters, but adversarial research shows that multi-turn, multi-agent jailbreaks can still bypass safeguards with high success rates. Security audits further reveal system-level risks that expose sensitive behaviors. Addressing these threats requires multi-layered defenses: resilient guard models, robust adversarial training datasets, and infrastructure-level protections. Improving reliability in tool use through preference-optimized training also reduces unintentional harms, making agents more dependable in real-world deployments.

\item \textbf{Multilingual fairness and reasoning consistency.} Inconsistent reasoning paths across languages risk inequitable outcomes. Recent methods directly optimize reasoning consistency by aligning intermediate steps rather than just final answers. Instruction tuning on multilingual reasoning data and program-based reasoning approaches provide stronger cross-lingual transfer, especially in low-resource settings, suggesting that structured reasoning formats such as code or tables offer a more stable substrate for alignment. Fairness and safety in multilingual contexts therefore require coherence in the reasoning process itself, beyond translation quality alone.
\end{itemize}

\subsection{Learning Paradigms and Future Architectures}

\subsubsection{Beyond Supervised Learning: RL and Self-Improvement}
Recent advances demonstrate that progress in reasoning depends less on simply scaling supervised learning and more on integrating reinforcement, search, preference optimization, and process-level feedback into the training loop. Instead of treating reasoning as a one-shot mapping from input to output, new paradigms teach models to explore trajectories, evaluate intermediate steps, and refine strategies over time. Search-based methods show that internalizing the structure of planning can yield reusable heuristics, while stepwise preference optimization highlights the value of fine-grained credit assignment within reasoning chains. Process reward models, generative rubrics, and reflective augmentation extend this further by turning intermediate reasoning into explicit learning signals, improving interpretability and robustness. Parallel efforts in efficiency and compute governance reframe inference itself as an optimization problem, with token budgets, speculative decoding, and uncertainty-triggered revisits ensuring that reasoning remains tractable at scale. Complementary work on synthetic data generation, capacity-matched distillation, and curriculum-style training highlights the role of carefully structured data in stabilizing reasoning for both large and small models. Together, these developments suggest that the path forward is not merely “more data” or “bigger models,” but architectures that learn to reason through iterative self-improvement, guided exploration, and adaptive allocation of cognitive resources.
 
\subsubsection{Unified Reasoning Frameworks and Future Directions}
Across domains, a shared trajectory is emerging toward unified reasoning frameworks that flexibly combine diverse strategies while managing efficiency, safety, and interpretability. Rather than fragmenting symbolic CoT, search, temporal reasoning, retrieval, or multimodal grounding into isolated pipelines, modular controllers are being designed to orchestrate when to search, retrieve, visualize, or reflect, with structural preservation in graphs, layouts, or temporal schemas anchoring reasoning to stable representations. Advances in code and algorithmic tasks emphasize process fidelity through executable scaffolds and formal verification, while tool-augmented and agentic systems illustrate how procedural planning and compute governance extend reliability under real-world constraints. Multilingual research underscores the need for reasoning-path alignment across languages, ensuring fairness and coherence beyond outcome accuracy, and social–cognitive work demonstrates the necessity of psychologically grounded evaluation to reveal perspective-taking and bias calibration deficits. Mathematical and multi-hop reasoning further highlight the value of stepwise supervision and adaptive search for compositional and symbolic abstraction. 

These strands point toward hybrid architectures where internal deliberation is tightly interwoven with external knowledge access, process-level feedback, and robust evaluation metrics. The open challenge is to standardize how controllers dynamically balance accuracy, interpretability, compute, and security while preserving cross-lingual and cross-modal fidelity. Progress will require architectures that not only unify reasoning across domains but also embed safeguards into the reasoning loop, yielding systems that are trustworthy, equitable, and adaptable across the diverse contexts where reasoning capabilities truly matter.

\section{Conclusion}
This survey covers reasoning in LLMs across chain-of-thought prompting, reinforcement learning, process supervision, retrieval-augmented systems, and more. These methods have produced real gains, but the reasoning they enable remains statistical, fragile under distributional shifts, and shallow in abstraction.
A recurring theme is the tension between pattern matching and genuine understanding. Despite generating outputs that mimic reasoning, models often lack principled generalization, struggle with coherence in extended reasoning chains, and remain brittle when confronted with novel variants of familiar tasks. This is compounded by challenges of integration and coherence in reasoning processes that span multiple steps, modalities, or tools often falter in consistency, leading to contradictions and failure in long-horizon planning. Evaluation further lags behind capability: current metrics overwhelmingly emphasize correctness of final answers, masking whether a model has reasoned or merely guessed.

The deficits across domains are deeply interconnected. Weaknesses in commonsense undermine visual reasoning; multilingual brittleness exposes gaps in structural alignment; poor temporal modeling reflects limits in internal simulation and memory. These point to a core absence: current LLMs lack a unified, explicit, and interpretable model of reasoning that can be verified, corrected, or governed.
Looking ahead, the path forward lies in building modular, tool-augmented, and cognitively aligned architectures. Symbolic-neural hybrids, agentic systems with self-reflection, and cross-modal representation alignment promise more robust and generalizable reasoning. Process-oriented learning through step-level supervision, reward models that value reasoning quality, and training signals that enforce decomposition and verification offers a way to move beyond surface heuristics. Hybrid architectures that integrate scaffolding from code reasoning, grounded interaction from tool use, and explicit memory for temporal and long-context reasoning are promising directions. Unified evaluation frameworks that measure reasoning depth, coherence, and generalization across domains, rather than task-specific accuracy alone, are equally needed.

These technical directions raise open questions: how should reasoning in machines be defined and measured? Can models learn to simulate, verify, and explain their thought processes, rather than only predicting answers? And how can advances in reasoning maintain transparency, reliability, and fairness across languages, cultures, and modalities?

\bibliography{tmlr}
\bibliographystyle{tmlr}


\end{document}